\PassOptionsToPackage{hyphens}{url}
\PassOptionsToPackage{table,xcdraw,dvipsnames}{xcolor}
\documentclass{article}
\usepackage{arxiv}

\usepackage[utf8]{inputenc}

\usepackage[natbib,style=numeric,doi=false,isbn=false,url=false,block=ragged]{biblatex}
\usepackage[final, kerning=true, factor=1100, stretch=10, shrink=10]{microtype}

\usepackage{amssymb}
\usepackage{stmaryrd}

\usepackage{mathtools}
\usepackage{longtable}
\usepackage{tabu}
\usepackage{booktabs}
\usepackage{multirow}
\usepackage{etoolbox}
\usepackage{siunitx}
\usepackage{xspace}
\usepackage{subcaption}
\usepackage[inline]{enumitem}
\usepackage{placeins}
\usepackage{afterpage}
\usepackage{tikz}
\definecolor{upbblue}{RGB}{0,32,91}
\definecolor{upbgray}{RGB}{85,85,85}
\usepackage[colorlinks = true,
            linkcolor = black,
            urlcolor  = upbblue,
            citecolor = upbblue,
            anchorcolor = black]{hyperref}	%
\usepackage{textcase}
\usepackage{hypernat}
\usepackage{adjustbox}
\usepackage{algorithm}
\usepackage[noend]{algpseudocode}

\definecolor{lime}{HTML}{A6CE39}
\DeclareRobustCommand{\orcidicon}{
	\begin{tikzpicture}
	\draw[lime, fill=lime] (0,0) 
	circle [radius=0.16] 
	node[white] {{\fontfamily{qag}\selectfont \tiny ID}};
	\draw[white, fill=white] (-0.0625,0.095) 
	circle [radius=0.007];
	\end{tikzpicture}
	\hspace{-2mm}
}
\foreach \x in {A, ..., Z}{\expandafter\xdef\csname orcid\x\endcsname{\noexpand\href{https://orcid.org/\csname orcidauthor\x\endcsname}
			{\noexpand\orcidicon}}
}

\usepackage{xwatermark}
\newwatermark[firstpage,color=gray!60,angle=270,scale=0.4, xpos=3.8in,ypos=0]{\href{https://doi.org/10.1016/j.ijar.2021.10.002}{\color{gray}{Publication DOI}}}

\usepackage[amsmath]{ntheorem}
\newcommand*{\qed}{\hbox{}\hfill$\Box$}
\usepackage[capitalise,noabbrev]{cleveref}
\crefname{appendix}{Appendix}{Appendices}
\theoremstyle{plain}
\newtheorem{theorem}{Theorem}[section]

\newtheorem{proposition}[theorem]{Proposition}

\newtheorem{claim}{Claim}[theorem]
\newtheorem{corollary}[theorem]{Corollary}

\theoremstyle{plain}
\theorembodyfont{\normalfont}

\newtheorem{example}[theorem]{Example}
\theoremsymbol{\ensuremath{\blacksquare}}
\theoremseparator{.}
\newtheorem*{proof}{Proof}%

\usepackage{tikz}

\usepackage[most]{tcolorbox}
\usepackage{xparse}

\usepackage[framemethod=TikZ]{mdframed}
\definecolor{examplegray}{rgb}{.907, .945, .943}  %
\mdfdefinestyle{MyFrame}{%
  linecolor=White,
  outerlinewidth=.3pt,
  roundcorner=5pt,
  innertopmargin=\baselineskip,
  innerbottommargin=\baselineskip,
  innerrightmargin=5pt,
  innerleftmargin=5pt,
  backgroundcolor=examplegray}

\newcommand*{\defeq}{\mathrel{\vcenter{\baselineskip0.5ex \lineskiplimit0pt
      \hbox{\footnotesize.}\hbox{\footnotesize.}}}%
  =}
\newcommand{\sharpP}{\#P\xspace}

\newcommand{\fromto}{\longrightarrow}
\renewcommand{\to}{\longrightarrow}
\renewcommand{\vec}[1]{\boldsymbol{#1}}
\newcommand{\on}{\operatorname}
\DeclareMathOperator{\Prob}{\mathit{P}}
\DeclareMathOperator{\prob}{\mathit{p}}
\renewcommand{\restriction}{\mathord{\upharpoonright}}

\newcommand{\IR}{\mathbb{R}}
\newcommand{\R}{\mathbb{R}}

\newcommand{\IC}{\mathcal{C}}
\newcommand{\IN}{\mathbb{N}}
\newcommand{\N}{\mathbb{N}}

\newcommand*{\sothat}{:}
\newcommand{\thatis}{i.\,e.,\xspace}
\newcommand*{\from}{\colon}
\newcommand{\data}{\mathcal{D}}
\newcommand{\argmin}{\operatorname*{\arg\,\min}}
\newcommand{\argmax}{\operatorname*{\arg\,\max}}

\newcommand{\cB}{\mathcal{B}}
\newcommand{\cC}{\mathcal{C}}
\newcommand{\cX}{\mathcal{X}}
\newcommand{\cD}{\mathcal{D}}

\newcommand{\cQ}{\mathcal{Q}}

\newcommand{\DomainOfUk}[1]{D_{#1}}

\newcommand{\cZ}{\mathcal{Z}}
\newcommand{\cL}{\mathcal{L}}

\newcommand{\fatenet}{\textsc{FATE-Net}\xspace}
\newcommand{\fetanet}{\textsc{FETA-Net}\xspace}
\newcommand{\fetalinear}{\textsc{FETA-Linear}\xspace}

\newcommand{\sda}{\textsc{SDA}\xspace}
\newcommand{\ranknet}{\textsc{RankNet}\xspace}
\newcommand{\ranksvm}{\textsc{RankSVM}\xspace}

\newcommand{\glm}{\textsc{GenLinearModel}\xspace}
\newcommand{\allpositive}{\textsc{AllPositive}\xspace}

\newcommand{\pairwisesvm}{\textsc{PairwiseSVM}\xspace}

\newcommand{\lrscheduler}{\textsc{LRScheduler}\xspace}

\newcommand{\metric}{\on{m}}

\newcommand{\qi}[1]{Q_{#1}}
\newcommand{\ci}[1]{C_{#1}}
\newcommand{\textlower}[1]{\MakeTextLowercase{#1}}
\newcommand{\card}[1]{\lvert #1 \rvert}
\newcommand{\cset}{\{\vec{x}_1, \ldots , \vec{x}_n\}}

\newcommand{\floor}[1]{\bigl \lfloor #1 \bigr\rfloor }
\newcommand{\set}[1]{\left\{#1\right\}}

\newcommand{\tqi}[1]{\tilde{Q}_{#1}}
\newcommand{\lvq}[1]{\boldsymbol{l}_{#1}}
\newcommand{\mq}[1]{MQ\num{#1}}
\newcommand{\mql}[1]{MQ\num{#1}-list}

\newcommand{\relmat}{R_{rel}}
\newcommand{\tagset}{T}
\newcommand{\mset}{M}
\newcommand{\mmap}{\mathcal{M}_f}
\newcommand{\nmovies}{N_m}
\newcommand{\ntags}{N_t}
\newcommand{\cosine}{\on{sim}_{\text{ws}}}

\newcommand{\tp}{\text{tag-pop}}
\newcommand{\df}{\text{tag-spec}}
\newcommand{\ui}{[0,1]}

\newcommand{\topk}[1]{\textsc{Top-$#1$}\xspace}
\newcommand{\acc}{\textsc{Accuracy}\xspace}

\DeclarePairedDelimiter{\indic}{\llbracket}{\rrbracket}
\DeclarePairedDelimiterX{\norm}[1]{\lVert}{\rVert}{#1}
\newcommand{\given}{\mid}

\newcommand{\makename}[3][s]{%
  \expandafter\newcommand\csname #2\endcsname{#3\xspace}%
  \expandafter\newcommand\csname #2s\endcsname{#3#1\xspace}%
}
\makename{qset}{task set}
\makename{Qset}{Task set}
\makename{cproblem}{choice task}
\makename{context}{context}
\makename{cotext}{cotext}
\makename{dc}{singleton choice}
\makename{cattopk}{Top-$k$ Categorical Accuracy}
\makename{catacc}{Categorical Accuracy}

\makename{msm}{Tag Genome Similar Movie}
\makename{mdm}{Tag Genome Dissimilar Movie}
\makename{bcf}{Best Critique-Fit Movie}
\makename{wcf}{Impostor Critique-Fit Movie}
\makename{wcs}{\emph{weighted cosine similarity}}
\makename{critfit}{\emph{critique-fit similarity}}
\makename{critdist}{\emph{critique-distance}}

\usepackage{acronym}
\newacro{FETA}{First Evaluate Then Aggregate}
\newacro{FATE}{First Aggregate Then Evaluate}
\newacro{SDA}{set-de\-pen\-dent aggregation}
\newacro{DTA}{decision-theoretic approach}
\newacro{CNN}{convolutional neural network}
\newacro{SCM}{singleton choice model}
\newacro{MOEA}{multi-objective evolutionary algorithm}
\newacro{MNIST}{Modified National Institute of Standards and Technology}
\newacro{ReLU}{rectified linear units}
\newacro{SELU}{self-normalizing linear units}
\newacro{BN}{batch normalization}
\newacro{SGD}{stochastic gradient descent}
\newacro{LETOR}{LEarning TO Rank}
\newacro{PL}{Preference learning}
\newacro{ERR}{Expected rank regression}
\newacro{IIA}{independence of irrelevant alternatives}
\newacro{DIA}{dependence on irrelevant alternatives}
\newacro{ML}{mixed logit}
\newacro{NL}{nested logit}
\newacro{GNL}{generalized nested logit}
\newacro{MNL}{multinomial logit}
\newacro{GEV}{generalized extreme value}
\newacro{RUM}{random utility model}

\newcommand{\Csingleton}[2]{C_{\mathrm{singleton}}( #1, #2 )}
\newcommand{\Csubset}[3]{C_{\mathrm{subset}}^{ #2 }(#1, #3 )}

\title{Learning Context-Dependent Choice Functions}

\usepackage{authblk}

\author[1\thanks{\tt{kiudee@mail.upb.de}}]{Karlson Pfannschmidt\orcidA{}}
\author[1\thanks{\tt{prithag@mail.upb.de}}]{Pritha Gupta\orcidB{}}
\author[1\thanks{\tt{bjoernha@mail.upb.de}}]{Björn Haddenhorst\orcidC{}}
\author[2\thanks{\tt{eyke@ifi.lmu.de}}]{Eyke Hüllermeier\orcidD{}}

\affil[1]{Paderborn University, Warburger Straße 100, Paderborn, Germany}
\affil[2]{LMU Munich, Akademiestr. 7, Munich, Germany}

\begin{document}

\maketitle

  \begin{abstract}%
    Choice functions accept a set of alternatives as input and produce a
     preferred subset of these alternatives as output.
    We study the problem of learning such functions under conditions of
    \emph{context-dependence} of preferences, which means that the preference in
    favor of a certain choice alternative may depend on what other options are also
    available.
    In spite of its practical relevance, this kind of context-dependence has
    received little attention in preference learning so far.
    We propose a suitable model based on context-dependent (latent) utility
    functions, thereby reducing the problem to the task of learning such utility
    functions.
    Practically, this comes with a number of challenges.
    For example, the set of alternatives provided as input to a choice
    function can be of any size, and the output of the function should not depend
    on the order in which the alternatives are presented.
    To meet these requirements, we propose two general approaches based on two
    representations of context-dependent utility functions, as well as
    instantiations in the form of appropriate end-to-end trainable neural network
    architectures.
    Moreover, to demonstrate the performance of both networks, we present extensive
    empirical evaluations on both synthetic and real-world datasets.
  \end{abstract}


\section{Introduction} %
\label{sec:intro}

The notion of \emph{preference} plays a central role in various
scientific disciplines, such as economics, psychology, and more recently also
computer science and artificial intelligence \citep{Domshlak}.
In these fields, mathematical formalisms have been developed for modelling and
reasoning about preferences, and for analyzing data that originates from
observed or revealed preferences.
In this regard, \emph{choice} observations are of specific interest,
in which a subset of ``good'' alternatives is selected from a set of available
candidates.
In particular, starting with the seminal
work by \citet{arrow1951}, \emph{choice functions} have been
analyzed as a key concept of a formal theory of choice and preference.
The study of pairwise preferences even goes back to work by
\citet{fechner1860elemente}, who considered the varying perception
of different stimuli.

In machine learning, preferences are at the core of
\emph{preference learning},
which has received increasing attention in recent years
\parencite{PL-book}.
Roughly speaking, the goal in preference learning is to learn (predictive)
preference models from preference data.
Somewhat surprisingly, and in spite of a close connection between ranking and
choice, the problem of learning \emph{subset} choice functions has received very little
attention so far, with only a few notable exceptions
\citep{benson2018,tomlinson2020}.
In this paper, we therefore address the problem of learning choice functions,
which express preferences in terms of subsets (or equivalently, bipartitions)
of $Q$.
From a machine learning point of view, the problem of learning
choice functions comes with a number of challenges.
For example, while algorithms for supervised learning normally assume inputs
in the form of feature vectors of fixed length, the inputs in our setting
are neither vectors nor of fixed size.
Instead, a choice function is supposed to accept inputs in the form of sets
$Q$ of any size, and to return a subset (choice) of the elements as output.
In case a set $Q$ is represented by an ordered list of its elements,
a choice function thus has to be invariant with respect to permutations of its input.

\begin{figure}[tb]
  \centering
  \begin{subfigure}[c]{0.23\linewidth}
    \includegraphics[width=\linewidth]{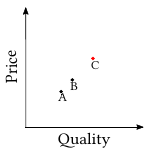}
    \subcaption{Compromise}
    \label{fig:compromise}
  \end{subfigure}
  \begin{subfigure}[c]{0.23\linewidth}
    \includegraphics[width=\linewidth]{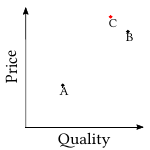}
    \subcaption{Attraction}
    \label{fig:attraction}
  \end{subfigure}
  \begin{subfigure}[c]{0.23\linewidth}
    \includegraphics[width=\linewidth]{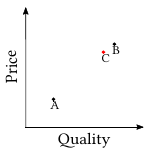}
    \subcaption{Similarity}
    \label{fig:similarity}
  \end{subfigure}
  \caption{Context effects identified in the literature
    \citep{rooderkerk2011incorporating}.}
  \label{fig:context}
\end{figure}

Not less interestingly, and in fact the key motivation of this paper,
choice functions could be \emph{context-dependent}, in the sense that
the preference in favor of an alternative may depend on what other options are
available.
Context-dependence of this kind has been observed, for example, in marketing
studies \citep{debreu1960review, bettman1998constructive}, and has been
investigated systematically in fields like economics and psychology.
More specifically, three major context effects have been identified in the
literature,
the compromise effect \citep{simonson1989choice},
the attraction effect \citep{huber1983market}, and the similarity effect
\citep{tversky1972elimination}:
\begin{itemize}
  \item The compromise effect states that the relative utility of an object
        increases by
        adding an extreme option that makes it a compromise in the set of
        alternatives
        \citep{rooderkerk2011incorporating}.
        For instance, consider the set of objects $\{A, B\}$ in
        \cref{fig:compromise}.
        The ordering of these objects depends on how much the consumer is
        weighing the quality of the product in relation to its price.
        If price is the main constraint, then the preference order will be
        $A \succ B$.
        But as soon as another extreme option $C$ becomes
        available, object $B$ 
        may be
        considered more favorable,
        because it represents a compromise between the three alternatives.
        Thus,
        the preference relation between $A$ and
        $B$ 
        might get inverted and turned into
        $B \succ A$.

  \item
        \cref{fig:attraction} illustrates the attraction effect.
        Here, if we add another object $C$ to the
        set of objects $\{A, B\}$, where $C$ is
        slightly dominated by $B$,
        the relative utility share for object $B$ increases with
        respect to $A$.
        The major psychological reason is that consumers have a strong
        preference for dominating products \citep{huber1983market}.
        Thus, the preference relation between $A$ and
        $B$ may again be influenced.

  \item
        The similarity or substitution effect is another phenomenon, according
        to which
        the presence of similar objects tends to reduce the overall probability
        of an object to be chosen, as it will divide the loyalty of potential
        consumers
        \citep{huber1983market}.
        In \cref{fig:similarity}, $B$ and
        $C$ are two similar objects.
        Consumers who prefer high quality will be divided amongst the two objects,
        resulting in a decrease of the relative utility share of object
        $B$.
        Again, this may lead to turning a preference $B \succ A$ into
        $A \succ
          B$, at least on an aggregate (population) level, if
        preferences are
        defined on the basis of choice probabilities.
\end{itemize}

Context-dependence as explained above has received only limited consideration in the
machine learning literature until recently
\citep{Chen16b,pfannschmidt2018,benson2018,seshadri19a,rosenfeld2020,Bower2020,Kleinberg2017}.

Additionally, the context effects discussed so far focus on effects that have
been observed for humans, but ignore that the space of (subset) choice functions and thus the number of possible applications is much larger.
Many algorithmic problems can be framed as a choice problem, e.\,g., in the
Knapsack problem one is tasked in choosing a set of maximal utility while obeying capacity constraints.
Computing the medoid of a set of points (i.\,e., the point with minimal distance to each other point) is a singleton choice problem.
It is clear that these problems cannot be solved by considering each choice
alternative individually, but the complete choice context needs to be
incorporated.
In practice, there are many abstract choice problems similar to these, e.\,g.,
portfolio selection \citep{markowitz1952}, algorithm selection \citep{rice1976, bischl2016} and
team selection \citep{vojnovic2016on}
just to name a few.
All these problems have in common, that the context-dependence naturally
arises because the output depends jointly on all objects in the set and
not because a decision maker behaves rationally or irrationally.

Motivated by its practical relevance, we formalize the problem of learning
context-dependent choice functions.
To this end, we provide a formal definition of such functions and propose a
data-generating process consisting of two stages: First, choice alternatives
are scored in terms of latent utility degrees, and then, a choice
set is determined on the basis of these scores (\cref{sec:probpref}).
Based on this model, we propose two representations of the latent
(context-dependent) utility, called \ac{FETA} and \ac{FATE}, which have appealing
properties from a learning point of view (\cref{sec:learning}), as
well as realizations of these models in terms of neural network architectures
(\cref{sec:endtoend}).
Thanks to these architectures, called \fetanet and \fatenet
\citep{pfannschmidt2018}, we are able to learn subset choices
on sets of objects in an end-to-end trainable manner.
To demonstrate the performance of both networks, we present extensive empirical
evaluations on both synthetic and real-world choice datasets (\cref{sec:eval}).
Additional information and supplementary material is provided in an appendix, to which we will refer occasionally.

\section{Related Literature} %
\label{sec:related}

The problem of how to model preferences in general has been extensively studied from
different viewpoints in the
past.
From an axiomatic/normative perspective, one posits which properties have to
hold for preferences to be considered ``rational,'' and studies consequences of
these properties.
Luce's choice axiom was introduced in \citeyear{luce1959} by
\citeauthor{luce1959} and requires that the preference between two items
does not depend on the presence or absence of any other choice alternative,
a property commonly referred to as \acf{IIA}.
The set of objects from which a particular preference is observed is also
called the \emph{context} \citep{mellers1983, chakravarti1983,rooderkerk2011incorporating}, and thus preferences obeying \ac{IIA} are also called context-independent \citep{kelman1996}.
In the same year, \citet[pp.\,56f]{debreu1959} proved the ordinal
representation
theorem, which shows that preferences can be represented by a continuous
utility function, if certain conditions including transitivity are assumed to
hold.
A related line of research was concerned with the concept of
\emph{revealed
  preferences}, for which most axioms can be reduced to some
notion of
transitivity \citep{samuelson1938,houthakker1950,sen1971}.

On the other side of the spectrum, observational studies in economics and
psychology were more concerned with how humans actually behave, 
and studied how the observed behavior deviates from \ac{IIA}
\citep{tversky1969,tversky1972elimination,huber1982adding,huber1983market,
  simonson1989choice,payne1992,simonson1992,shafir1993,tversky1993,doyle1999,
  payne1999,sedikides1999}.
It consistently was observed that choice behavior depended on the specific collection of alternatives available, the \emph{context} of the choice.
\citet{rooderkerk2011incorporating} and \citet{rieskamp2006} provide an
extensive overview of the different context effects which were identified
over the years and which we already showcased in the introduction.
This motivated researchers to come up with methods able to model these
violations.
Classical \acp{RUM}, like the \ac{MNL} model, are not able to take these
effects into account.
Therefore, extensions of \acp{RUM} were proposed, which are able to
capture the compromise and attraction effect \citep{tversky1993, kivetz2004,orhun2009}, the
similarity effect \citep{tversky1972elimination, kamakura1984}
or all of the above \citep{rooderkerk2011incorporating}.
One important line of research focuses on the assumption that the
decision maker chooses
based on multiple utility functions (so called ``multiple selfs'', or “multi-self” for short), which are suitably aggregated.
This setting has been studied in economics
\citep{may1954,kalai2002,kivetz2004,fudenberg2006,manzini2007,green2007}
and
psychology \citep{tversky1972elimination,shafir1993,tversky1993}.
Continuing this line of research, \citet{ambrus2014} show that by
utilizing a
collection of context-independent utility functions, combined with a suitable
aggregation, one is able to model arbitrary choice functions.
That is, choice behavior across multiple sets can be modelled even though it
might violate context-independence.

While traditional research on preferences, as discussed above, is mostly of a
normative, prescriptive or descriptive nature, the advent of machine learning
triggered a shift towards ``predictive'' models.
\citet{rosenfeld2020} build on ideas of the multi-self literature and
propose
to learn set-dependent weights and embeddings, which are then linearly
combined to arrive at an aggregated score for each object.
\citet{benson2018} consider the problem of learning preferences in
the form of subsets of objects.
To this end, they extend the classical multinomial logit model to account for
violations of context-independence.
Higher-order interactions between objects are added specifically for those
subsets that cause a violation.
The set of objects for which choices or choice sets are observed is assumed to
be fixed.
Therefore, the approach cannot be used for arbitrary task sets, where it
can happen that an object is only observed once.
Our approach to decompose a context-dependent utility function into an
aggregation across smaller sub-contexts has been a recent, promising direction
in studying choices
\citep{pfannschmidt2018,seshadri19a}, and will be the focus of this paper.

Decomposition approaches have also been employed in the related field of
``learning to rank''.
\citet{AiBGC18} employ a context-independent model to pre-sort the objects,
while a recurrent neural network is used in a subsequent step to fine-tune the
ranking.
The \ac{FATE} approach, introduced in the context of choice by \citet{pfannschmidt2018}, obviates the need to pre-sort the objects, by directly embedding each object to produce a
representation for each set of objects (aggregation), which is then used as the
context to produce the final ranking (evaluation).
The authors also introduce an algorithm where this order is swapped, called
\ac{FETA}, in which each object is scored in the context of another object first,
and only then the scores are aggregated to produce a final ranking.
\citet{AiWBGBN19} later consider a similar decomposition, where higher order
interactions are approximated by employing sampling.

\section{A Probabilistic Model of Choice} %
\label{sec:probpref}
We start by establishing the necessary notation (refer to \cref{asec:notation} for an overview). Throughout this paper, $\indic{A}$ is  defined to be $1$ if $A$ is a true statement, and $0$ otherwise. 
We will denote by $\cX \subset \IR^d$ a set of reference objects
serving as choice alternatives, which, for simplicity, we assume to be finite
(albeit of arbitrary size), if not explicitly stated otherwise.
An object or item $\vec{x} \in \cX$ is represented by a vector of
features
$\vec{x} = (x_1, \dots, x_d) \in \cX$.
A non-empty subset $\cQ$ of $2^{\cX}\setminus \{\emptyset\}$ is called a \emph{choice task space} if $\emptyset \not\in \cQ \not= \emptyset$ and any $Q\in \cQ$ is called a \emph{choice task}. A \emph{choice} for $Q\in \cQ$ is a non-empty subset of $Q$ and the set $\cC \coloneqq \bigcup_{Q\in \cQ} 2^{Q} \setminus \{\emptyset\}$ of choices for any $Q\in \cQ$ is called the \emph{choice space}.

We say that a function $c\from \cQ \to \cC$ is a \emph{(subset) choice function (for $\cQ$)} if $c(Q) \subseteq Q$ is fulfilled for any $Q\in \cQ$, and in case $|c(Q)|=1$ holds for any $Q\in \cQ$, $c$ is called a \emph{singleton choice function (for $\cQ$)}.
A typical example for a real-world singleton choice function is when a user enters a query
in a search engine and receives a list of results ($Q$) of which they pick one and click on.
Subset choice functions usually occur, when a diverse set of objects is sought,
e.\,g., a search engine decides on a set of the most relevant, but diverse, results to
display to the user.

As common in machine learning, the input-output dependency of interest, in our
case between tasks and choices, is not assumed to be deterministic.
Instead, we assume a probabilistic dependence, which is captured by a
(conditional) probability distribution $\prob( \cdot \given Q)$ on
the non-empty subsets of $Q$ for every $Q\in \cQ$. Here, $p(C\mid Q)$ is interpreted as the probability to observe the  choice $C$ given the task $Q$. For the sake of convenience, we suppose w.l.o.g.  $\prob( \cdot \given Q)$ to be extended to $\cC$ via $\prob( C \given Q) \coloneqq 0$ for any $C\in \cC \setminus 2^{Q}$. Moreover, we write for short $\prob(\vec{x} \given Q)$ for $p(\{\vec{x}\} \given Q)$.
In case $\prob(Q)$ is the latent probability that $Q\in \cQ$ is given as task, the whole data-generating process is modelled by the joint distribution
\begin{equation}\label{eq:joint}
  \prob(Q,C) \coloneqq
  \prob(Q) \cdot \prob(C \given Q)
\end{equation}
on $\cQ \times \cC$.

We call the choice probabilities
\emph{context-independent} if 
\[
  \frac{\prob( C \mid Q)}{\prob( C' \mid Q)} =
  \frac{\prob( C \mid Q')}{\prob( C' \mid Q')}
\]
is fulfilled for every $Q,Q' \in \cQ$ and any $C,C' \in \cC$ with $C,C'\subseteq Q\cap Q'$. Conversely, we say that a system of choice distributions is
\emph{context-dependent}, if
this equality is violated on at least one pair of $Q, Q' \in \cQ$. This definition extends in a straight-forward and consistent way the notion of \acf{IIA} introduced by \citet{arrow1951}, which was originally only defined for the case of singleton choice, in which $\cC$ consists of elements of size one only.
We choose to use the more general term of context-(in)dependence, for the simple reason that the
notion of “irrelevant” alternatives is rather tailored to the analysis of human choices but less meaningful in our more general setting of arbitrary choice functions. 

As an example, consider the knapsack problem, where the goal is to select a set of objects which maximize
a certain utility, while obeying capacity constraints.
It is clear that the decision on which object to include in the choice set needs to
incorporate the complete choice task context, and that
one is not able to ascertain the relative choice probability of two alternatives
while ignoring all others.
As already explained in the introduction, context-independence is often violated in practice.
This motivates the development of context-dependent learning methods.

\paragraph{Utility-Based Choices} %
\label{sub:two_stage}

\begin{figure}[tb]
  \centering
  \includegraphics{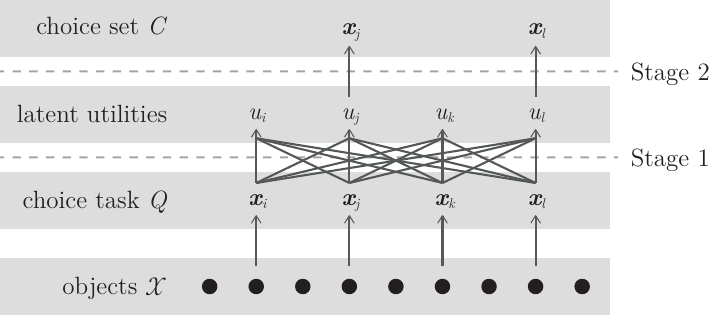}
  \caption{Overview of the data-generating process:
    First, a task $Q$ is produced (with probability
    $p(Q)$) by sampling from $\cX$.
    The objects in $Q$ are assigned latent utility
    degrees, and the choice set is finally constructed on the basis of these scores.
  }
  \label{fig:setting}
\end{figure}

We propose to model choices as the result of a two-stage process
(cf. \cref{fig:setting} for an overview), grounding them on the
notion of \emph{utility}:
In the first stage, each object in a given task $Q\in \cQ$ is assigned a
real-valued utility score.
Then in the second stage, choices are generated based on these scores.

Utility theory has a long history in economics
\citep{neumann1944,debreu1954,may1954}. Originally introduced as a way to measure the satisfaction achieved by a certain alternative \citep{bentham1789},
it is nowadays common in decision theory to consider
utility more as an abstract value that ought to be maximized
by any rational decision maker \citep{neumann1944, russell2010}.
This is formalized by means of a \emph{generalized utility function} (for $\cQ$)
\begin{equation}\label{eq:latut}
    U \from \{(\vec{x},Q) \sothat \vec{x} \in Q \in \cQ \} \to \IR, \quad (\vec{x},Q) \mapsto U(\vec{x},Q),
\end{equation}
which allows for modelling the utility of an object as a function of
both, properties of the object itself as well as properties of other choice
alternatives in $Q$, which constitute the context in
which $\vec{x}$ is considered: $U (\vec{x} , Q )$
expresses a degree of utility of $\vec{x}$ in the context
$Q$, \thatis given the availability of other choice alternatives
$\vec{x'} \in Q \setminus \{
  \vec{x} \}$.
The score $U(\vec{x},Q)$ is supposed to capture an abstract notion of utility, which in turn
reflects the propensity of $\vec{x}$ to be chosen in any task $Q$.

We call a utility function \emph{context-independent} in case $U(\vec{x},Q) = U(\vec{x},Q')$ holds for any $Q,Q'\in \cQ$ with $\vec{x} \in Q\cap Q'$ and \emph{context-dependent} otherwise. Via abbreviating $U(\vec{x}) \coloneqq U(\vec{x},Q)$ for some arbitrary $Q\in \cQ$ with $\vec{x} \in Q$, any context-independent utility function may be thought of as a function $U: \cX \fromto \IR$.

Moving on to the second stage, based on a utility function $U$, one may define in a deterministic manner for $Q\in \cQ$ the corresponding \emph{singleton choice}  as
\begin{equation}\label{eq:detsingleton}
    \Csingleton{U}{Q} \coloneqq \argmax_{\vec{x} \in Q} U(\vec{x},Q)
\end{equation}
and for $t\in \IR$ the \emph{subset choice (with threshold $t$)} as
\begin{equation}\label{eq:detsubset}
    \Csubset{U}{t}{Q} \coloneqq \{\vec{x} \in Q \sothat U(\vec{x},Q) \geq t\}.
\end{equation}
Clearly, $\Csingleton{U}{\cdot}$ and $\Csubset{U}{t}{\cdot}$ are in fact choice functions and in case $\vec{x} \mapsto U(\vec{x},Q)$ is injective (\thatis there are no ties),
for any $Q\in \cQ$, the former one is a singleton choice function.
There is an interesting connection to social choice theory, where
a social choice rule is employed to select an outcome out of a set of possible
outcomes in order to maximize some notion of utility for a population of
individuals with possibly varying utility functions.
The injectivity of such a social choice rule is called \emph{resoluteness} and it 
is an important property considered in social choice theory,
where it also plays a role in several impossibility results
\citep{kelly1990,ozkes2021}.
The singleton choice is a special case of the more general top-$k$ choice,
where the goal is to select the $k$ best objects.
It differs from subset choice in so far that the size of the choice sets is
always fixed, whereas in subset choice it can vary.
The top-$k$ choice setting has strong connections to the ranking setting,
which we will discuss below.

Further note that using thresholding to convert a set of scores into a partition
is a standard approach in multi-label classification \citep{koyejo2015} and
multi-criteria sorting \citep{alvarez2021}.

In the probabilistic setting, the utility function $U$ may serve to model probabilistic choices $p(\cdot \mid Q)$, $Q\in \cQ$, on $\cC$ by using the utility scores as the corresponding  parameters of the distributions. Certainly, there are various ways in which this idea could be realized:

\begin{description}
  \item[Singleton choice] In the case of \emph{singleton choice},
  a natural assumption is the 
        \acf{MNL} model, in which for any $Q\in \cQ$ and $\vec{x} \in Q$,
        \begin{equation}\label{eq:mnl}
          \prob_{\text{MNL}}( \vec{x}  \mid Q )
          \coloneqq \frac{\exp\left( U(\vec{x},Q) \right)}{\sum_{\vec{x'} \in Q} \exp( U(\vec{x'},Q) )}.
        \end{equation}
        and $p(C\mid Q) = 0$ for any $C\in 2^{Q}$ of size $\geq 2$ \citep{berkson1944,gurland1960,cox1966,mantel1966,theil1969}.
        Note here that these choice probabilities are context-independent, if $U$ is
        context-independent.
        An important special case is the Bradley-Terry-Luce model \citep{bradley1952}, which only considers pairwise comparisons (\thatis $|Q|=2$ for all $Q\in\cQ$).

  \item[Subset choice] For the choice of arbitrary subsets (not limited to singleton sets), a simple model is obtained by treating the inclusion or exclusion of each
        object $\vec{x}$ in a task $Q$ as independent given the utilities.
        This results in the distributions $\prob(\cdot \mid  Q)$ given by
        \begin{equation}\label{eq:likebin}
          \prob(C  \mid Q)  \coloneqq \gamma(U,Q)\prod_{\vec{x} \in Q} \frac{\exp(\indic{\vec{x} \in C}U(\vec{x},Q))}{1+\exp(U(\vec{x},Q))} \enspace 
        \end{equation}
        for any non-empty $C\in 2^{Q}$ and $Q\in \cQ$, where $\gamma(U,Q)$ is a constant such that $\sum_{C\in 2^{Q}\setminus \{\emptyset\}} \prob(C\mid Q) = 1$ holds. If $U$ is context-independent, the quantity 
        \begin{align*}
            \frac{\prob(C\mid Q)}{\prob(C'\mid Q)} = \prod_{\vec{x} \in Q} \frac{\exp(\indic{\vec{x} \in C}U(\vec{x}))}{\exp(\indic{\vec{x} \in C'}U(\vec{x}))} = \prod_{\vec{x} \in C\cup C'}
            \frac{\exp(\indic{\vec{x} \in C}U(\vec{x}))}{\exp(\indic{\vec{x} \in C'}U(\vec{x}))}
        \end{align*}
        does not depend on $Q$, and thus the choice probabilities $\prob(C \mid Q)$ are context-independent as well.

  \item[Choices based on rankings]
        Yet another type of model is obtained by assuming that, based on the latent
        utilities $U(\vec{x},Q)$, $\vec{x} \in Q$, a ranking $\pi$ on $Q$ is sampled first and then turned into a choice set via a (possibly probabilistic) procedure $g\colon \pi \mapsto g(\pi) \in 2^{Q}$ afterwards. The probability $\prob(C\mid Q)$ is then simply the probability that this procedure results in the output $C$, \thatis  
        \begin{equation}\label{eq:othertype}
          \prob(C \mid Q ) \coloneqq	\sum\nolimits_{\pi}
          \prob(\pi) \, \prob(g(\pi) = C) \enspace ,
        \end{equation}
        where the sum is taken over all possible rankings $\pi$ over $Q$.
        An approach of that kind might be appealing, because probability
        distributions on rankings have been studied quite thoroughly in the literature.
        Important families of ranking distributions include distance-based ranking models
        \citep{fligner1986}, of which the Mallows model \citep{mallows1957} is a popular
        instance, and multistage ranking models \citep{fligner1988}, most prominently represented by the 
        Plackett-Luce distribution \citep{plackett1975}.
        An important special case for $g$ is \emph{top-$k$ choice}, where the first
        $k$ objects are chosen deterministically (\thatis
        $g(\pi) = \{\pi^{-1}(1),\dots,\pi^{-1}(k)\} \subseteq Q$ holds with probability $1$ for any ranking $\pi \from Q \to \mathbb{N}$).
        This can be generalized, for example, by assuming that the size
        $k$ is not fixed but random.
        An even more general model has recently been proposed by
        \citet{fahandar2017}, where choices are not necessarily restricted to
        top-$k$ sets.
\end{description}
In this paper, we are mainly interested in tackling the problem of learning
context-dependent choice functions from training data.
The performance of a particular hypothesis, \thatis a choice function $c \from \cQ \to \cC$,
is measured by an appropriate loss function (see \cref{sec:learning}).
In \cref{sub:loss_functions} we go into more detail on how to derive suitable
loss functions from \eqref{eq:mnl} and \eqref{eq:likebin}.
After having introduced suitable models for utility-based choices, we now turn
to the problem of representing context-dependent choice functions.

\section{Learning Context-Dependent Choice Functions}
\label{sec:learning}
Our main interest in this paper is to tackle choice from
a machine learning perspective.
More specifically, we seek to induce a predictive choice function $c \from \cQ \to \cC$
from training
data $\cD = \{ (Q_i , C_i) \}_{i=1}^N \subset \cQ \times \cC$ in the form of exemplary tasks
$Q_i$
together with observed choices $C_i \in 2^{Q_{i}}$.
The performance of such a function is measured in terms of its expected loss (risk)
\[
  R(c) \coloneqq \int_{\cQ \times \cC} L(C, c(Q)) \mathop{d\hspace{-2pt}\prob(Q,C)} \, ,
\]
where $L: \, \cC \times \cC \fromto \mathbb{R}$ is a loss function (cf. \cref{sub:loss_functions} for an overview of the loss functions we consider), and $\prob$ the probability measure
associated with the distribution \eqref{eq:joint}, \thatis the underlying data-generating process
modelling the probability of observing tasks $Q$
together with choices~$C$.
The Bayes predictor~$c^*$ assigns each task
$Q$ the respective loss minimizer 
\[
  c^*\from Q \mapsto \argmin_{\hat{C} \in \cC} \int_{\cC}
  L(C, \hat{C}) \mathop{d\hspace{-2pt}\prob(C \mid Q)} \, .
\]
Since $\prob(Q, C)$ is usually unknown, one therefore opts to
minimize the \emph{empirical risk}
\begin{equation}
  R_{\text{emp}}(c) \coloneqq
  \frac{1}{N}\sum_{i=1}^N L(C_i, c(Q_i))
\end{equation}
on the given data $\mathcal{D}$ instead.

Assuming the data to be generated according to one of \eqref{eq:detsingleton}--\eqref{eq:othertype} 
(known to the learner) and by means of an (unknown) latent utility function \eqref{eq:latut}, this loss minimization problem essentially comes down to learning the generalized
utility function \eqref{eq:latut}.
This function, while allowing one to model context-dependence, causes several
practical problems, mainly because its second argument,
$Q$, is a \emph{set} of \emph{variable} size.

Many machine learning models such as neural networks or support vector machines require data to be given in the form of a feature vector $\vec{x} \in \R^{m}$. Hence, in order to apply such a model for learning a utility function
$U \from \{(\vec{x},Q) \sothat \vec{x} \in Q \in \cQ \} \to \IR$,
we have to fix an injective feature transformation $\Psi \from \cQ \to \R^{m}$. 

We choose to represent $Q = \{ \vec{y}_{1},\dots,\vec{y}_{k}\} \in \cQ \subset \R^{d}$ 
by the vector $(\vec{y}_{1},\dots,\vec{y}_{k}) \in \R^{kd}$. Of course, this does only define a valid  transformation $\Psi$ in case $|Q|$ is the same for each $Q\in \cQ$. Assuming this to be the case, we may consider a utility function
$U \from \{(\vec{x},Q) \sothat \vec{x} \in Q \in \cQ \} \to \IR$
as a function $\R^{(k+1)d} \fromto \R$. Noticing that ${Q = \{\vec{x}_{\sigma(i)} \sothat i\in [k]\}}$ holds for any bijection $\sigma \from [k] \to [k]$, this function should necessarily be \emph{permutation-invariant} or \emph{symmetric} in the sense that 
\begin{equation}
    U(\vec{x},(\vec{x}_{1},\dots,\vec{x}_{k})) = U(\vec{x},(\vec{x}_{\sigma(1)},\dots,\vec{x}_{\sigma(k)}))
\end{equation}
for each permutation $\sigma$ on $[k]$ \citep{stanley2001}.

The utility choice models proposed below will enforce this property and are also capable of dealing with tasks of different sizes.
More specifically, we present two general decompositions, which are able to
approximate a generalized latent utility function \eqref{eq:latut}.
\cref{sub:FETA} describes \acs{FETA}, which
decomposes \eqref{eq:latut} into
first- and second-order (or, more generally, higher order) utility
functions and aggregates the corresponding scores into an overall
utility score.
The \acs{FATE} approach (\cref{sub:FATE}), on the other hand, first computes an embedding
of the complete object context $Q$ in a space of fixed
dimensionality, and evaluates the utility of each object in that space.
The former could be advantageous for datasets, of which the choice task
contexts can be expressed through \emph{local} interactions,
while the latter is useful, if the set of objects as
a whole can be summarized by suitable \emph{global} properties
(e.\,g., choosing that element of a set, which is closest to the centroid of all elements in this set).

\subsection{\acl{FETA}} %
\label{sub:FETA}
Recall that the overall objective is to model the context-dependent utility function
\eqref{eq:latut}, \thatis the utility of each object should not only depend
on object attributes, but also on the choice task $Q$.
One way of handling the problem of rating objects in contexts of variable size
is to decompose a context into sub-contexts of a fixed size $k$
\citep{pfannschmidt2018,seshadri19a}.
More specifically, the idea is to learn 
\emph{sub-utility functions} $U_{0},\dots,U_{K}$
of the form $U_{0}:\cX \fromto \IR$ and
\[
U_{k} \from \DomainOfUk{k}\fromto \IR \, , \quad  \DomainOfUk{k} \coloneqq \{(\vec{x},A) \sothat \vec{x} \in \cX \text{ and } A\subseteq \cX \setminus \{\vec{x}\} \text{ with } |A| = k\}
\]
for $1\leq k\leq K \leq |Q|$, and represent the original function \eqref{eq:latut} as an aggregation
\begin{align}\label{eq:agg}
  U(\vec{x} , Q)   \coloneqq   U_{0}(\vec{x}) + \sum_{k=1}^K  \overline{U}_k(\vec{x}, Q),
\end{align}
where $\overline{U}_{k}(\vec{x},Q)$ is the average over the values $U_{k}(\vec{x},Q')$ for subsets $Q'$ of $Q\setminus \{\vec{x}\}$ consisting of $k$ distinct elements, \thatis formally 
\begin{align*}
    \overline{U}_{k}(\vec{x},Q) =
    \frac{1}{\binom{|Q|}{k}-\binom{|Q|-1}{k-1}}
    \hspace{23pt}
    \sum_{\mathclap{Q' \subseteq Q\setminus \{\vec{x}\}\colon |Q'|=k}}\hspace{10pt} U_{k}(\vec{x},Q').
\end{align*}
Note, that the sum is taken w.r.t. to all $k$-sized subsets $Q'$ of $Q\setminus \{\vec{x}\}$, potentially including some in $2^\cX \setminus \cQ$.
Here, $U_{k}(\vec{x},Q)$ may be thought of as a measure to which extent an item $\vec{x}$ is preferred to the elements of $Q$, and $\overline{U}_{k}(\vec{x},Q)$ as an indicator of how much $\vec{x}$ is  on average  preferred to $k$ distinct elements from $Q \setminus \{\vec{x}\}$.
We refer to this approach as \acf{FETA}, because an
alternative is first evaluated in each sub-context, and these evaluations are
then aggregated. Accordingly, 
we call $U$ defined in  \eqref{eq:agg} the \emph{FETA utility function with sub-utility functions $U_{0},\dots,U_{K}$} and denote it by $U_{\text{FETA}}^{U_{0},\dots,U_{K}}$.

\citet{batsell1985} propose a related expansion in the context of market share
modelling.
\citet{seshadri19a} call it an instantiation of the
\emph{universal logit model}, since it can be seen as a generalization of
the multinomial logit model~\eqref{eq:mnl}, when conditioning on
the task $Q$.

Roughly speaking, the motivation behind the above decomposition is that
dependencies and interaction effects between objects should only occur up
to a certain order $K+1$, or at least can be limited to
this order without losing too much information.
To see what we mean by “order” in this context, observe that
the first order model ($K=0$) reduces to $U_0(\vec{x})$
and thus only models the inherent utility of each object.
A second order model ($K=1$) then introduces pairwise terms.
This is an assumption that is commonly made in the literature on aggregation
functions \citep{grab_af}.
The reason why the utilities are averaged for a fixed $k$, but summed
across different $k$, is to give each order equal weight.
This prevents the utility from being dominated by higher-order interactions.
Furthermore, it allows the sub-utility functions to output scores in
roughly the same scale, which is advantageous when the model is applied
to choice tasks $Q$ of varying size.

Given the models of context-dependent choices as outlined above, the
learning problem essentially comes down to learning the
utility function \eqref{eq:agg} of order $K+1$.
From this function, one can then derive the utility function
\eqref{eq:latut}, which in turn allows for deriving predictions of
choices via the choice functions discussed before.

In this paper, we realize \eqref{eq:agg} for the special case
$K=1$, which can be seen as a second-order approximation of a
context-dependent utility function.
Thus, we propose the representation of a choice function $c$
based on a latent sub-utility function $U_0 \sothat \mathcal{X} \fromto \IR$ and a
pairwise function
$U_1 \sothat \DomainOfUk{1}
  \fromto \IR$.
In this way, the FETA utility function with sub-utility functions $U_{0},U_{1}$ may be written as 
 \begin{equation}
  \label{eq:fetapair}
  U(\vec{x}, Q)
  = U_{0}(\vec{x}) + \frac{1}{|Q|-1} \sum\nolimits_{\vec{y} \in Q \setminus \{\vec{x}\}} U_{1}(\vec{x},\{\vec{y}\}) \, .
\end{equation}
The value $U_{0}(\vec{x})$ can be seen as a kind of inherent, context-independent utility
of $\vec{x}$, whereas the scores $U_1(\vec{x} , \{\vec{y}\})$, $\vec{y} \in Q \setminus \{\vec{x}\}$, 
serve as ``corrections'' of this utility in the context of the task $Q$.
\begin{figure}[htb]  %
  \begin{mdframed}[style=MyFrame]
    \begin{example}[\ac{FETA}: Context-dependence]\label{ex:feta}
      As a simple illustration, suppose $\cX$ to consist of $4$ elements $\vec{a},\vec{b},\vec{c},\vec{d} \in \IR^{d}$, let $\cQ = 2^{\cX} \setminus \{\emptyset\}$ and  $U$ be the FETA utility function with sub-utility functions  $U_{0},U_{1}$ defined as follows:
      \begin{center}
      \scalebox{0.9}{
      \begin{tabular}{l|S|SSSS}
       & {$U_{0}( \cdot )$} & {$U_1(\cdot,\vec{a})$} &  {$U_{1}(\cdot,\vec{b})$} &  {$U_{1}(\cdot,\vec{c})$} &  {$U_{1}(\cdot,\vec{d})$} \\\midrule
        {$\vec{a}$} & {$-0.8$} & {---} & {$1.2$} & {$0.8$} & {$0.0$} \\
        {$\vec{b}$} & {$-0.7$} & {$0.0$} & {---} & {$1.2$} & {$1.4$} \\
        {$\vec{c}$} & {$-0.7$} & {$0.6$} & {$0.0$} & {---} & {$0.2$} \\
        {$\vec{d}$} & {$-0.8$} & {$1.0$} & {$0.0$} & {$0.8$} & {---} \\ 
        \end{tabular}
      }
      \end{center}
      Then, the utilities for the tasks $\{\vec{a},\vec{b},\vec{c}\}$ and $\{\vec{a},\vec{b},\vec{d}\}$ are given as 
       \begin{center}
      \scalebox{0.9}{
      \begin{tabular}{l|S|S}
       & {$U(\cdot,\{\vec{a},\vec{b},\vec{c}\})$} & {$U(\cdot , \{\vec{a},\vec{b},\vec{d}\})$} \\\midrule
        {$\vec{a}$} & {$0.2$} & {$-0.2$} \\
        {$\vec{b}$} & {$-0.1$} & {$0.0$} \\
        {$\vec{c}$} & {$-0.4$} & {---} \\
        {$\vec{d}$} & {---} & {$-0.3$} \\ 
        \end{tabular}
      }
      \end{center}
      For the task $\{\vec{a},\vec{b},\vec{c}\}$ the item $\vec{a}$ has a higher utility score than $\vec{b}$, whereas $\vec{b}$ is preferred over $\vec{a}$ for the task $\{\vec{a},\vec{b},\vec{d}\}$, \thatis the preference  between $\vec{a}$ and $\vec{b}$ changes depending on whether the third item in the task set is $\vec{c}$ or $\vec{d}$.
    \end{example}
  \end{mdframed}
\end{figure}
\citet{seshadri19a} propose a similar approximation, but instead of averaging the
task context, the authors simply sum up all utilities and impose sum-to-zero constraints to guarantee identifiability.

As for the FETA model $U_{\mathrm{FETA}}^{(U_{0},U_{1})}$,
we will now see that it is identifiable up to the choice of $U_{0}$.
\begin{proposition}\label{Prop_FETA_idf}
	Suppose $|\mathcal{X}| \geq 4$ and $\mathcal{Q}$ to be  such that for any distinct $\vec{x},\vec{y},\vec{z} \in \mathcal{X}$ there is some $Q\in \mathcal{Q}$ with $\{\vec{x},\vec{y}\} \subseteq Q \not\ni \vec{z}$. Let  $U_{0},\tilde{U}_{0}\colon \mathcal{X} \to \R$ and $U_{1},\tilde{U}_{1}\colon \DomainOfUk{1} \to \R$ be arbitrary. Then, we have $U_{\mathrm{FETA}}^{(U_{0},U_{1})} = U_{\mathrm{FETA}}^{(\tilde{U}_{0},\tilde{U}_{1})}$ if and only if
	\begin{equation*}
	    \forall \vec{x} \in \mathcal{X}, \forall \vec{y}\in \mathcal{X} \setminus \{\vec{x}\}\colon
	    \tilde{U}_{1}(\vec{x},\{\vec{y}\}) = U_{1}(\vec{x},\{\vec{y}\}) - \tilde{U}_{0}(\vec{x}) + U_{0}(\vec{x}).
	\end{equation*}
\end{proposition}
    \begin{proof}
	$\Leftarrow$ is clear.
	For proving the remaining implication $\Rightarrow$, suppose that
	$U_{\mathrm{FETA}}^{(U_{0},U_{1})} = U_{\mathrm{FETA}}^{(\tilde{U}_{0},\tilde{U}_{1})}$.
	\begin{claim}\label{Prop_FETA_idf_c1}
	For any distinct $\vec{x},\vec{y},\vec{z} \in \mathcal{X}$ we have 
	\begin{equation*}
		U_{1}(\vec{x},\{\vec{y}\}) - U_{1}(\vec{x},\{\vec{z}\}) = \tilde{U}_{1}(\vec{x},\{\vec{y}\}) - \tilde{U}_{1}(\vec{x},\{\vec{z}\}).
	\end{equation*}
    \end{claim}
	\begin{proof}
	For arbitrary $Q,Q'\subseteq \mathcal{X}$ with $|Q|=|Q'|$, $\{\vec{x},\vec{y}\} \subseteq Q \not\ni \vec{z}$ and $\{\vec{x},\vec{z}\} \subseteq Q' \not\ni \vec{y}$ we have
	\begin{align*}
		U_{\mathrm{FETA}}^{(U_{0},U_{1})}(\vec{x},Q) - U_{\mathrm{FETA}}^{(U_{0},U_{1})}(\vec{x},Q') = \frac{1}{|Q|-1}\left( U_{1}(\vec{x},\{\vec{y}\}) - U_{1}(\vec{x},\{\vec{z}\})\right).
	\end{align*}
	Since this holds for arbitrary $(U_{0},U_{1})$ (and thus also for $(\tilde{U}_{0},\tilde{U}_{1})$), \cref{Prop_FETA_idf_c1} follows. \qed
    \end{proof}
	
	Now, let $\vec{x}_{0}\in \mathcal{X}$ be fixed for the moment and define $b\colon \mathcal{X} \to \R$ via 
	\begin{equation*}
		b(\vec{x}) \coloneqq \begin{cases} \tilde{U}_{0}(\vec{x}) - U_{0}(\vec{x}),\quad &\text{if } \vec{x} = \vec{x}_{0},\\ 
		U_{1}(\vec{x},\{\vec{x}_{0}\}) - \tilde{U}_{1}(\vec{x},\{\vec{x}_{0}\}),\quad &\text{if } \vec{x} \not= \vec{x}_{0}.
		\end{cases}
	\end{equation*}
	According to \cref{Prop_FETA_idf_c1} we have for any distinct $\vec{x},\vec{y} \in \mathcal{X} \setminus \{\vec{x}_{0}\}$ the identity  
	\begin{equation*}
		\tilde{U}_{1}(\vec{x},\{\vec{y}\}) = U_{1}(\vec{x},\{\vec{y}\}) - \left( U_{1}(\vec{x},\{\vec{x}_{0}\}) - \tilde{U}_{1}(\vec{x},\{\vec{x}_{0}\}) \right) = U_{1}(\vec{x},\{\vec{y}\}) - b(\vec{x}).
	\end{equation*}
	Moreover, the definition of $b$ assures that $\tilde{U}_{1}(\vec{x},\{\vec{x}_{0}\}) = U_{1}(\vec{x},\{\vec{x}_{0}\}) - b(\vec{x})$ holds for any $\vec{x} \not= \vec{x}_{0}$, i.\,e., $b$ already fulfills 
	\begin{equation}
		\label{eq_proof_1}
		\forall \vec{x}\in \mathcal{X} \setminus \{\vec{x}_{0}\}\colon \forall \vec{y}\in \mathcal{X} \setminus \{\vec{x}\}\colon \tilde{U}_{1}(\vec{x},\{\vec{y}\}) = U_{1}(\vec{x},\{\vec{y}\}) - b(\vec{x}).
	\end{equation}
	For $\vec{x} \in \mathcal{X} \setminus \{\vec{x}_{0}\}$ we may choose a query set $Q \subseteq \mathcal{X} \setminus \{\vec{x}_{0}\}$ and then \eqref{eq_proof_1} assures us 
	\begin{align*}
		\tilde{U}_{0}(\vec{x}) - U_{0}(\vec{x})
		&= U_{\mathrm{FETA}}^{(\tilde{U}_{0},\tilde{U}_{1})} (\vec{x},Q) - U_{\mathrm{FETA}}^{(U_{0},U_{1})}(\vec{x},Q)
		+ \frac{1}{|Q|-1} \sum\nolimits_{\vec{y} \in Q\setminus \{\vec{x}\}} \left( U_{1}(\vec{x},\{\vec{y}\}) - \tilde{U}_{1}(\vec{x},\{\vec{y}\})\right) \\
		&= \frac{1}{|Q|-1} \sum\nolimits_{\vec{y} \in Q\setminus \{\vec{x}\}} \left( U_{1}(\vec{x},\{\vec{y}\}) - \tilde{U}_{1}(\vec{x},\{\vec{y}\})\right) \\
		&= \frac{1}{|Q|-1} \sum\nolimits_{\vec{y} \in Q\setminus \{\vec{x}\}} b(\vec{x})\\
		&= b(\vec{x}).
	\end{align*}
	Since $\tilde{U}_{0}(\vec{x}_{0}) = U_{0}(\vec{x}_{0}) + b(\vec{x}_{0})$ holds by definition of $b$, we thus have shown
	\begin{equation}\label{eq_proof_3}
		\forall \vec{x} \in \mathcal{X}\colon b(\vec{x}) =  \tilde{U}_{0}(\vec{x}) -  U_{0}(\vec{x}).
	\end{equation}
	With regard to \eqref{eq_proof_1} it remains to show
	\begin{equation}\label{eq_proof_2}
		\forall \vec{y} \in \mathcal{X} \setminus \{\vec{x}_{0}\}\colon \tilde{U}_{1}(\vec{x}_{0},\{\vec{y}\}) = U_{1}(\vec{x}_{0},\vec{y}) - b(\vec{x}_{0}).
	\end{equation}
	 For this, note that the same argumentation as before with $\vec{x}_{0}$ replaced by some arbitrary $\vec{x}_{1} \in \mathcal{X} \setminus \{\vec{x}_{0}\}$ shows us that $b$ also fulfills \eqref{eq_proof_1} with $\vec{x}_{0}$ replaced by $\vec{x}_{1}$. In particular, \eqref{eq_proof_2} holds. Combining \eqref{eq_proof_1}, \eqref{eq_proof_3} and \eqref{eq_proof_2} completes the proof.
	 \\\qed
	\end{proof}

\begin{corollary}
	Suppose $\mathcal{X}$ and $\mathcal{Q}$ are as in \cref{Prop_FETA_idf} and let  $U_{0}\colon \mathcal{X} \to \R$ be fixed. Then, the mapping $U_{1} \mapsto U_{\mathrm{FETA}}^{(U_{0},U_{1})}$ is injective.
\end{corollary}

Another interesting theoretical question concerns the expressiveness of the
\ac{FETA} decomposition: Which predictors $c\from \cQ \to \cC$
can be represented by  \ac{FETA}?
The following result shows that the decomposition into pairwise utilities
\eqref{eq:fetapair} is indeed a restriction, in the sense that it does
not
allow for representing the entire class of predictors
in case $|\cX| \geq 7$.

\begin{proposition}
    If $|\cX| \geq 7$, not every singleton choice function on $\cX$ can be expressed via the second order FETA model. More precisely: For distinct $\vec{a},\vec{b},\vec{c},\vec{x},\vec{y},\vec{x'},\vec{y'} \in \cX$ there do not exist sub-utility functions  $U_{0} \from \cX  \to \R$, $U_{1} \from \DomainOfUk{1}\to \R$ and $t\in \R$ such that the choice function $c \from \cQ \to \cC$ defined either via $c(\cdot) \coloneqq \Csingleton{U_\mathrm{FETA}^{U_{0},U_{1}}}{\cdot} $ or via $c(\cdot) \coloneqq \Csubset{U_\mathrm{FETA}^{U_{0},U_{1}}}{t}{\cdot}$  fulfills 
    \begin{align}
        c(\{\vec{a},\vec{b},\vec{c},\vec{x}\}) &= \{\vec{a}\}, & c(\{\vec{a},\vec{b},\vec{c},\vec{y}\}) &= \{\vec{a}\},  &  c(\{\vec{a},\vec{b},\vec{x},\vec{y}\}) &= \{\vec{b}\}, \label{eq_FETA_counterex_1} \\
        c(\{\vec{a},\vec{b},\vec{c},\vec{x'}\}) &= \{\vec{b}\}, & c(\{\vec{a},\vec{b},\vec{c},\vec{y'}\}) &= \{\vec{b}\}, & c(\{\vec{a},\vec{b},\vec{x'},\vec{y'}\}) &= \{\vec{a}\}. \label{eq_FETA_counterex_2}
    \end{align}
\begin{proof}
    We prove the statement indirectly. To this end, fix distinct $\vec{a}$, $\vec{b}$, $\vec{c}$, $\vec{x}$, $\vec{y}$, $\vec{x'}$, $\vec{y'} \in \cX$ and assume there were some $U_{0},U_{1}$ and $t\in \R$ such that $c$ defined either via $c(\cdot) \coloneqq \Csingleton{U_\mathrm{FETA}^{U_{0},U_{1}}}{\cdot}$  or via $c(\cdot) \coloneqq \Csubset{U_\mathrm{FETA}^{U_{0},U_{1}}}{t}{\cdot}$ fulfills both \eqref{eq_FETA_counterex_1} and \eqref{eq_FETA_counterex_2}.
     With the convenient abbreviations $u_{\vec{r}} \coloneqq U_{0}(\vec{r})$ and $u_{\vec{r},\vec{s}} \coloneqq U_{1}(\vec{r},\{\vec{s}\})$, the following constraints for \eqref{eq:fetapair} immediately follow from \eqref{eq_FETA_counterex_1}:
    \begin{align*}
        u_{\vec{a}} + \frac{1}{3} \left( u_{\vec{a},\vec{b}} + u_{\vec{a},\vec{c}} + u_{\vec{a},\vec{x}}\right) &> u_{\vec{b}} + \frac{1}{3} \left( u_{\vec{b},\vec{a}} + u_{\vec{b},\vec{c}} + u_{\vec{b},\vec{x}}\right),\\
        u_{\vec{a}} + \frac{1}{3} \left( u_{\vec{a},\vec{b}} + u_{\vec{a},\vec{c}} + u_{\vec{a},\vec{y}}\right) &> u_{\vec{b}} + \frac{1}{3} \left( u_{\vec{b},\vec{a}} + u_{\vec{b},\vec{c}} + u_{\vec{b},\vec{y}}\right),\\
        u_{\vec{b}} + \frac{1}{3} \left( u_{\vec{b},\vec{a}} + u_{\vec{b},\vec{x}} + u_{\vec{b},\vec{y}}\right) &> u_{\vec{a}} + \frac{1}{3} \left( u_{\vec{a},\vec{b}} + u_{\vec{a},\vec{x}} + u_{\vec{a},\vec{y}}\right).
    \end{align*}
    Summing up the first two inequalities and then applying the third one yields 
    \begin{align*}
        &2u_{\vec{a}} + \frac{1}{3} \left( 2u_{\vec{a},\vec{b}} + 2u_{\vec{a},\vec{c}} + u_{\vec{a},\vec{x}} + u_{\vec{a},\vec{y}} \right)
        \\
        &> u_{\vec{b}} + \frac{1}{3}\left( u_{\vec{b},\vec{a}} +u_{\vec{b},\vec{x}} + u_{\vec{b},\vec{y}} \right) + u_{\vec{b}} + \frac{1}{3}\left( u_{\vec{b},\vec{a}} + 2u_{\vec{b},\vec{c}} \right) \\
        &> u_{\vec{a}} + \frac{1}{3}\left( u_{\vec{a},\vec{b}} + u_{\vec{a},\vec{x}} + u_{\vec{a},\vec{y}} \right) + u_{\vec{b}} + \frac{1}{3} \left( u_{\vec{b},\vec{a}} + 2u_{\vec{b},\vec{c}} \right),
    \end{align*}
    from which we obtain via subtracting common terms 
    \begin{equation}
        u_{\vec{a}} + \frac{1}{3}\left( u_{\vec{a},\vec{b}} + 2u_{\vec{a},\vec{c}} \right) > u_{\vec{b}} + \frac{1}{3} \left( u_{\vec{b},\vec{a}} + 2u_{\vec{b},\vec{c}} \right). \label{eq_FETA_counterex_3}
    \end{equation}
    Exactly the same argumentation (with the roles of $\vec{a}$ and $\vec{b}$ interchanged and $\vec{x}$ resp. $\vec{y}$ replaced by $\vec{x'}$ resp. $\vec{y'}$) lets us infer from \eqref{eq_FETA_counterex_2} 
    \begin{equation*}
         u_{\vec{b}} + \frac{1}{3} \left( u_{\vec{b},\vec{a}} + 2u_{\vec{b},\vec{c}} \right) > u_{\vec{a}} + \frac{1}{3}\left( u_{\vec{a},\vec{b}} + 2u_{\vec{a},\vec{c}} \right),
    \end{equation*}
    which contradicts \eqref{eq_FETA_counterex_3}. This completes the proof. \qed
\end{proof}
\end{proposition}

Note that a limited expressivity should not necessarily be seen as a negative property. In particular, from a machine learning perspective, an overly excessive expressivity (or \emph{capacity} of the underlying hypothesis space) is connected with the practical problem of poor generalization due to overfitting, \thatis being overly expressive may prevent the learner from identifying the right model.  
In any case, we expect \ac{FETA} to work well for all choice
functions that (approximately) decompose into a pairwise relation between
objects.
Naturally, this leads to the question whether it is possible to incorporate
more of the set-based context without ultimately increasing computational
complexity.
This question motivated our next decomposition.

\subsection{\acl{FATE}} %
\label{sub:FATE}

To deal with the problem of task contexts of variable size, our previous approach
was to decompose the context into sub-contexts of a fixed size, evaluate an
object $\vec{x}$ in each of the sub-contexts, and then
aggregate these evaluations into an overall assessment.
An alternative to this \ac{FETA} strategy, and in a
sense contrariwise approach, consists of first aggregating the task into
a representation of fixed size, and then evaluating the object
$\vec{x}$ in the presence of this task representative.

More specifically, the \ac{FATE} approach requires a mapping $\phi$ from $\cX$ to some $m$-dimensional embedding space  $\cZ \subseteq \IR^{m}$ as well as a context-dependent \emph{sub-utility function} $U' \from \cX \times \cZ \to \IR$. To evaluate an object $\vec{x}$ in  a choice task $Q \in \cQ$, the \ac{FATE} strategy first computes  $\frac{1}{|Q|} \sum_{\vec{y} \in Q} \phi(\vec{y})$ as representative for the task and then evaluates it via $U'$ as 
\begin{equation}\label{eq:NEW_FATE_def}
    U(\vec{x},Q) \coloneqq U'\left( \vec{x}, \frac{1}{|Q|} \sum\nolimits_{\vec{y} \in Q} \phi(\vec{y}) \right). 
\end{equation}
We call this $U$ the \emph{\ac{FATE} utility function with sub-utility function  $U'$ and transformation $\phi$} and denote it by $U_{\text{FATE}}^{U',\phi}$.

\begin{figure}[htb]  %
  \begin{mdframed}[style=MyFrame]
    \begin{example}[\ac{FATE}: Context-dependence]
    
      Similar as in \cref{ex:feta}, suppose $\cX$ to consist of four elements $\vec{a},\vec{b},\vec{c},\vec{d} \in \IR^{d}$, let $\cZ\coloneqq \IR$ and $\phi \from \cX \to \cZ$ and $U' \from \cZ \to \IR$ be such that   
      \begin{center}
      \scalebox{0.8}{
      \begin{tabular}{l|SSS}
      {$\mathcal{X}$} & {$\phi(\cdot)$} & {$U'(\cdot, 2)$} & {$U'(\cdot, 3)$}\\\midrule
      {$\vec{a}$} & 1 &  0.5 & -0.1 \\
      {$\vec{b}$} & 2 & -0.1 &  0.5 \\
      {$\vec{c}$} & 3 & -0.2 & -0.2 \\
      {$\vec{d}$} & 6 & -0.3 & -0.3
      \end{tabular}
      }
      \end{center}
       and $U'(\cdot,z)$ be arbitrary for any $z\in \R \setminus\{2,3\}$.
        For $Q_{1} \coloneqq \{\vec{a},\vec{b},\vec{c}\} $ and $Q_{2} \coloneqq \{\vec{a},\vec{b},\vec{d}\}$ the quantity  $\frac{1}{|Q_{i}|}\sum_{\vec{y} \in Q_i} \phi(\vec{y})$ is $2$ if $i=1$ and $3$ if $i=2$. Consequently, we have 
        $U(\vec{a},Q_{1}) = U'(\vec{a},2) = 0.5 > -0.1 = U'(\vec{b},2) = U(\vec{b},Q_{1})$ and at the same time $U(\vec{a},Q_{2}) = U'(\vec{a},3) < U'(\vec{b},3) = U(\vec{b},Q_{2})$, \thatis the preference between $\vec{a}$ and $\vec{b}$ changes depending on whether the third item in the set is $\vec{c}$ or $\vec{d}$.
    \end{example}
  \end{mdframed}
\end{figure}

This approach is related to recent advances on dealing with
set-valued inputs in neural networks \citep{zaheer2017,ravan2017,batt2018},
where a permutation-equivariant network directly maps from sets of objects to scores.
\citet{rosenfeld2020} propose to learn set-dependent aggregation functions 
with an inductive bias towards principles from behavioral choice theory.
They note that general models like Deep Sets \citep{zaheer2017},
which try to approximate set functions
using a permutation-invariant neural network,
are overly general, because they have a high \emph{violation capacity},
i.\,e., the flexibility of the model to change its choices, when
objects are removed from the choice task.
The \ac{FATE} approach on the other hand first condenses the task context into a representative and only then scores each object.
The resulting model has an inductive bias that favors functions for which the object
utility depends on such a set-global reference object.
This could be advantageous for datasets where the set of objects as a whole can
be summarized by suitable \emph{global} properties 
(e.\,g., choosing that element from a set, which is closest to the centroid of all elements in the set),
such that the task to score the objects with this context becomes easy.
\ac{FETA} on the other hand, incorporates task-information through \emph{local}
interactions.

Without further assumptions on $\phi$ and $U'$, this model is able to express any possible choice function $c$ on $\cQ$, as we show in the following. The proof of the upcoming result is similar to the proof of Theorem 2 by \citet{zaheer2017}.
\begin{proposition}
   Suppose $\mathcal{X}$ to be countable and $\mathcal{Q} \subseteq \{Q \subseteq \mathcal{X} \sothat |Q| < \infty\}$. There exists a parametrization $\phi \from \mathcal{X} \to \IR$ with the following property:
	\begin{itemize}
		\item[(i)]
		For any singleton choice function $c$ on $\mathcal{Q}$, there is a utility function $U_{c}' \from \mathcal{X} \times \IR \to \IR$ such that $\Csingleton{ U_{\mathrm{FATE}}^{U'_{c},\phi}}{Q} = c(Q)$ holds for any $Q\in \mathcal{Q}$.
		\item[(ii)]
		For any subset choice function $c$ on $\mathcal{Q}$ there exists a utility function $U'_{c} \from \mathcal{X} \times \IR \to \IR$ with $\Csubset{U_{\mathrm{FATE}}^{U'_{c},\phi}}{1/2}{Q} = c(Q)$ for any $Q\in \mathcal{Q}$.
	\end{itemize}
\begin{proof}
    Since $\mathcal{X}$ is countable, there exists an injective function $\delta \from \mathcal{X} \to  \mathbb{N}$. For $\vec{x}\in \mathcal{X}$ define
	\begin{equation*}
		\phi(\vec{x}) \coloneqq \ln\left( p_{\delta(\vec{x})} \right),
	\end{equation*}
	wherein $p_{i} \in \mathbb{N}$ denotes the $i$-th prime number for any $i\in \mathbb{N}$. Before proving (i) and (ii), we show that the mapping 
	\begin{equation*}
	    \Phi \from \cQ \to \IR, \quad Q \mapsto \frac{1}{|Q|}\sum\nolimits_{\vec{x}\in Q} \phi(\vec{x})
	\end{equation*}
	is injective. 
	For this, let $Q,Q'\in \cQ$ with $\Phi(Q) = \Phi(Q')$. Then,  
	\begin{equation*}
		\frac{|Q'|}{|Q|} = \frac{\ln\left( \prod_{\vec{x}\in Q'} p_{\delta(\vec{x})} \right)}{\ln \left( \prod_{\vec{x}\in Q} p_{\delta(\vec{x})} \right)} = \log_{b}(a)
	\end{equation*}
	holds for the integers $a \coloneqq \prod\nolimits_{\vec{x}\in Q'} p_{\delta(\vec{x})}$ and $b \coloneqq \prod\nolimits_{\vec{x}\in Q} p_{\delta(\vec{x})}$, \thatis $a^{|Q|} = b^{|Q'|}$. As $a$ and $b$ are both products of distinct primes, the uniqueness of the prime factorization lets us infer $a=b$ and thus also $Q=Q'$.
	
	We proceed with proving (i) and (ii) simultaneously. For this, suppose any choice function $c$ on $\cQ$ to be fixed. Since $\Phi$ from above is injective, there exists a mapping $\Psi \from \IR \to \cQ$ such that $\Psi(\frac{1}{|Q|}\sum_{\vec{x}\in \mathcal{Q}} \phi(\vec{x})) = Q$ holds for any $Q\in \mathcal{Q}$.
	Note, that $\Psi\restriction_{\Phi(\mathcal{Q})}$ is the inverse function of $\Phi$.
	Thus, the claim follows with the choice
	\begin{align*}
		U'_{c}(\vec{x},\vec{z}) \coloneqq \indic{\vec{x}\in c(\Psi(\vec{z}))}.
	\end{align*}
	\qed
\end{proof}
\end{proposition}

Although this expressivity is desirable in general, it comes at a cost.
The FATE model $U_{\mathrm{FATE}}^{(\phi,U')}$ as such is \emph{not identifiable}:
For example, suppose $U'\colon\mathcal{X} \times \mathcal{Z} \to \R$ is of the form $U'(\vec{x},\vec{z}) \coloneqq f(\vec{x}) + \norm{z}_{2}$ for some function $f\colon\mathcal{X} \to \R$, where $\norm{\cdot}_{2}$ denotes the standard euclidean norm in $\R^{d} \supseteq \mathcal{Z}$. For arbitrary $\phi_{1}\colon \mathcal{X} \to \mathcal{Z}$, we obtain with $\phi_{2} \coloneqq -\phi_{1}$ that
\begin{align*}
	U_{\mathrm{FATE}}^{(\phi_{1},U')}(\vec{x},Q) - U_{\mathrm{FATE}}^{(\phi_{2},U')}(\vec{x},Q)
	= \norm*{\frac{1}{|Q|} \sum\nolimits_{\vec{y} \in Q} \phi_{1}(\vec{x})}_{2} - \norm*{-\frac{1}{|Q|} \sum\nolimits_{\vec{y} \in Q} \phi_{1}(\vec{x})}_{2}  = 0
\end{align*}
for any $Q\in \mathcal{Q}, \vec{x} \in Q \subseteq \mathcal{X}$, i.\,e., $U_{\mathrm{FATE}}^{(\phi_{1},U')} =  U_{\mathrm{FATE}}^{(\phi_{2},U')}$ holds.

\subsection{Linear Sub-Utility Functions}

A related question concerns the expressivity of the \ac{FATE} and \ac{FETA} approaches,
when the underlying sub-utility functions and transformations are linear functions.
In case $\phi$ and $U'$ are chosen as linear functions in the sense that $\phi(\vec{x}) = \vec{A}\vec{x}$ and $U'(\vec{x},\vec{z}) = \vec{c}^{t}\vec{x} + \vec{d}^{t} \vec{z}$ for any $(\vec{x},\vec{z}) \in \cX \times \cZ$ and some $\vec{A} \in \IR^{d\times m}$, $\vec{c} \in \IR^{d}$ and $\vec{d} \in \IR^{m}$,  \eqref{eq:NEW_FATE_def} takes the form  
\begin{equation*}
    U_{\text{FATE}}^{U',\phi}(\vec{x},Q) = \vec{c}^{t}\vec{x} + \vec{d}^{t}\left( \frac{1}{|Q|} \sum\nolimits_{\vec{y}\in Q} \vec{A} \vec{y} \right).
\end{equation*}
As the second summand therein does not depend on $\vec{x}$, for any $Q\in \cQ$, the singleton choice $\Csingleton{U_{\text{FATE}}^{U',\phi}}{Q}$ is the same as that corresponding to the linear utility function $\vec{x} \mapsto \vec{c}^{t} \vec{x}$ and thus independent of the context $Q$. Consequently, at least one of $U'$ and $\phi$ has to be non-linear in order to model context-dependent choices.

In contrast to this, for the case of \ac{FETA}, linearity of the sub-utility functions does not imply context-independence of the model: If $U_{0}$ and $U_{1}$ are linear in the sense that $U_{0}(\vec{x}) = \vec{b}^{t}\vec{x}$ and $U_{1}(\vec{x},\{\vec{y}\}) = \vec{c}^{t}\vec{x} + \vec{d}^{t} \vec{y}$ for any distinct $\vec{x},\vec{y} \in \cX$ and some weight vectors $\vec{b}, \vec{c}, \vec{d} \in \IR^{d}$, the FETA utility function with sub-utility functions $U_0, U_1$ is given as  
\begin{align*}
    U(\vec{x},Q) &= \vec{b}^{t} \vec{x} + \frac{1}{|Q|-1} \sum\nolimits_{\vec{y} \in Q \setminus \{\vec{x}\} } \left( \vec{c}^{t}\vec{x} + \vec{d}^{t} \vec{y} \right) \\
    &=  \left( \vec{b}+\vec{c}\right)^{t} \vec{x} + \frac{1}{|Q|-1} \sum\nolimits_{\vec{y} \in Q\setminus \{\vec{x}\}} \vec{d}^{t}\vec{y}
\end{align*}
for any $\vec{x} \in Q \in \cQ$. As the second summand therein depends not only on $Q$ but also on $\vec{x}$, $U$ can in general \textbf{not} be represented as a linear function.

\section{Implementation Using Neural Networks} %
\label{sec:endtoend}
Having defined the decomposition strategies \acs{FETA} and
\acs{FATE} in the preceding
section, we are still missing an algorithm, which can actually learn the
utility functions involved.
In this section, we propose realizations of the \ac{FETA} and \ac{FATE} approaches in
terms of neural network architectures \fetanet and \fatenet, respectively.
Our design goals for both neural networks are twofold.
First, they should be end-to-end trainable using (stochastic) gradient descent,
such that they can be used as part of a larger neural network architecture.
To this end, we ensure that the outputs of the networks are differentiable
almost everywhere with respect to the weights.
Similarly, the loss functions employed in conjunction with a regularization term
for the weights should also be differentiable almost everywhere
and convex with respect to the utilities.
Second, the architectures should be able to generalize beyond the task
sizes encountered in the training data,
since in practice it is unreasonable to expect all choice tasks to be of the
same size.

\subsection{\fetanet Architecture} %
\label{sub:fetanet}
We will now describe our first neural network architecture \fetanet and its
training.
Recall from \cref{sub:FETA} that we seek to predict
utility scores
$U(\vec{x}_i, Q)$ of the form \eqref{eq:fetapair}
for every object $\vec{x}_i \in Q$.
What we need to learn, therefore, is the functions $U_0$ and
$U_1$.
\begin{figure}[t]
  \centering
  \includegraphics{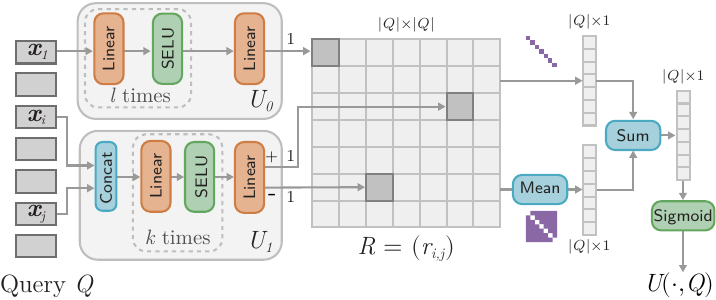}
  \caption{The \fetanet architecture implementing the \ac{FETA}
    approach.
  Layers with trainable weights are shown in orange, while operations
  without trainable weights are drawn in blue and non-linearities are
  depicted in green.}
  \label{fig:fetanet}
\end{figure}%
In \fetanet, we do so by means of a deep neural network
architecture (shown in \cref{fig:fetanet}).
The network is trained in a set of data $\cD = \{(Q_i, C_i)\}_{i=1}^N$,  where
each $Q_i$ is a choice task and
$C_{i} \in 2^{Q_{i}} \setminus \{\emptyset\}$ the choice set observed for that task.

The main component is the neural network tasked with learning the pairwise
utility function $U_1$ (depicted in blue).
It receives the feature vectors of two objects $\vec{x}_i$ and
$\vec{x}_j$
and outputs a score for $\vec{x}_i$ in the presence of object
$\vec{x}_j$.
To build up the complete matrix $R = (r_{i,j})$ would require
iterating over all pairs of objects in $Q$.
This is why we choose to adopt the CmpNN approach by \citet{Rigutini2011}
for the pairwise scoring function, \thatis instead of one output
neuron we utilize two $U_1^{+}$ and $U_1^{-}$.
Weight sharing ensures that
$U_1^{+}(\vec{x}_i, \vec{x}_j) =
 U_1^{-}(\vec{x}_j, \vec{x}_i)$ and
 $U_1^{-}(\vec{x}_i, \vec{x}_j) =
 U_1^{+}(\vec{x}_j, \vec{x}_i)$ holds.
For the diagonal, we evaluate a separate network $U_0(\vec{x}_i)$,
which learns a latent utility component for each object
(corresponding to the case $k=0$ in \eqref{eq:agg}).
With that it suffices to iterate over all combinations of objects once,
and to construct the matrix $R$ as follows:
\begin{equation}
  r_{i,j} = \begin{cases}
    U_1^{+}(\vec{x}_i, \vec{x}_j) & \text{if } i < j \\
    U_1^{-}(\vec{x}_i, \vec{x}_j) & \text{if } i > j \\
    U_0(\vec{x}_i) & \text{otherwise}
  \end{cases}
\end{equation}
Then, each row of the relation $R$ is averaged  to
obtain a score
$U(\vec{x}_i, Q) = r_{i, i} + \frac{1}{|Q| - 1}\sum_{1 \leq j\neq i\leq |Q|} r_{i,j}$
for each object $\vec{x}_i \in Q$.
Therefore, the network $U_1$ is a mapping $\IR^d \times \IR^d \fromto \IR^2$
and $U_0$ a mapping $\IR^d \fromto \IR$
which can be instantiated by any neural network architectures suitable for the
given objects.
For our experiments later on, we shall use deep, densely connected networks.
We treat the number of layers and units as hyperparameters and optimize them
jointly with all the other hyperparameters.

\begin{algorithm}[tb]
  \caption{\fetanet training algorithm}
  \label{alg:feta}

  \begin{algorithmic}[1]
    \Require
    \Statex Dataset $\cD = \{(Q_i, C_i)\}_{i=1}^N$ with $Q_i \subset \IR^d$
    \Statex Pairwise network $U_1 \colon \IR^d \times \IR^d \fromto \IR^2$, parametrized by $\theta_1$
    \Statex Diagonal network $U_0 \colon \IR^d \fromto \IR$, parametrized by $\theta_0$
    \Statex Batch size $b \in \IN$, Number of epochs
    $E \in \IN$
    \Statex Step size schedule $\vec{\eta} = (\eta_1, \eta_2, \dots)$ with
    $\eta_i \in \IR_{>0}\ \forall i\in \IN$
    \Statex Loss function $L\colon \cC \times \bigcup_{k\in\IN}\IR^k \fromto \IR$

    \Procedure{Train-\fetanet}{$\cD, U_0, U_1, b, E, \vec{\eta}, L$}
    \State Initialize random weight vectors $\theta_0, \theta_1$
    \For{Epoch $ep \in [E]$}
    \State $\cD \gets \Call{Shuffle}{\cD}$
    \State $T \gets \left\lceil\frac{N}{b}\right\rceil$
    \State Construct mini-batches $\cB_1, \dots, \cB_T$
    \For{Iteration $t \in [T]$}
    \State $\ell_t \gets 0$
    \ForAll{$(Q, C) \in \cB_t$}
    \For{$1 \leq i \leq j \leq |Q|$}
    \If{$i < j$}
    \State $r^{\text{tmp}} \gets
      U_1(\vec{x}_i, \vec{x}_j)$
    \State $r_{i,j} \gets
      r_0^{\text{tmp}},\quad r_{j,i}
      \gets r_1^{\text{tmp}}$
    \Else
    \State $r_{i, i} \gets
      U_0(\vec{x}_i)$
    \EndIf
    \EndFor
    \State $\vec{u} \gets \left(r_{i, i} + \frac{1}{|Q|-1}\sum_{1\leq j\neq i \leq |Q|} r_{i, j}\right)_{i=1}^{|Q|}$
    \State $\ell_t \gets \ell_t  + L(C, \vec{u})$
    \EndFor
    \State $\theta_0 \gets \theta_0 - \frac{\eta_{ep\cdot T+t}}{|\cB_t|}
      \nabla_{\theta_0} \ell_t$
    \State $\theta_1 \gets \theta_1 - \frac{\eta_{ep\cdot T+t}}{|\cB_t|}
      \nabla_{\theta_1} \ell_t$
    \EndFor
    \EndFor
    \EndProcedure
  \end{algorithmic}
\end{algorithm}
The complete training algorithm for \fetanet is shown in
\cref{alg:feta}, which is an instantiation of stochastic
gradient descent.
We will denote the weight vectors of the networks $U_0$
and $U_1$ by
$\theta_0$ and $\theta_1$, respectively.
In the beginning, these weight vectors are suitably initialized in order to
avoid exploding/vanishing gradients \citep{glorot2010,he2015}.
In each epoch, the algorithm shuffles the given dataset and constructs
mini-batches $\cB_1, \dots, \cB_T$ with $\cB_i \subset \cD$ for
all
$i \in [T]$.
In lines 10 to 18, the pairwise relation is constructed as described above.
The utilities $\vec{u} = (u_1, \dots u_{|Q|})$ for the objects inside the
task $Q$ are computed in line 19 by summing the
pairwise relation $r_{i,j}$
across the columns of the matrix.
Finally, the loss is computed in line 20 and added to the cumulative loss
for the batch.
The weight vectors $\theta_0$ and $\theta_1$
are updated using
backpropagation in lines 22--23.

It is easy to see, that the training runtime complexity per epoch (including backpropagation) of \fetanet is $\mathcal{O}\left(N d q^2\right)$, where $N$ denotes
the number of instances, $d$ is the number of features
per object, and
$q \coloneqq \max_{(Q, Y) \in \cD} |Q|$ is an upper bound on the number of
objects in each choice task.
For a new task $Q$, the prediction time is in
$\mathcal{O}\left(d |Q|^2\right)$.

\subsection{\fatenet Architecture} %
\label{sub:fatenet}
\begin{figure}[tb]
  \centering
  \includegraphics{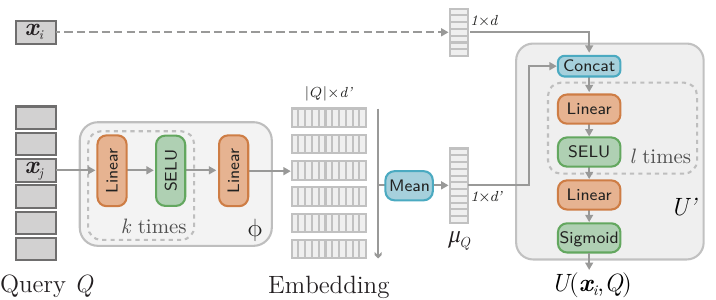}
  \caption{The \fatenet architecture implementing the \ac{FATE}
    approach.
    Here we show the score head for object $\vec{x}_i$.
    Layers with trainable weights are shown in orange, while operations
    without trainable weights are drawn in blue and non-linearities are
    depicted in green.
  }
  \label{fig:newnet}
\end{figure}
The second architecture we propose is called \fatenet, and the structure for
predicting the score for one object is depicted in
\cref{fig:newnet}.
Inputs are the $n$ objects of the task
$Q = \{\vec{x}_1, \dots, \vec{x}_n\}$ (shown in green).
Each object is independently passed through a deep, densely connected
embedding layer (shown in blue).
The embedding layer approximates the function $\phi$ in
\eqref{eq:NEW_FATE_def} and is a map $\IR^d \to \IR^{d'}$.
Note that we employ weight sharing, \thatis the same embedding is used for
each object.
Then, the representative $\mu_Q$ for the task $Q$ is
computed by averaging the
representations of each object.
To calculate the score $U(\vec{x}_i, \mu_{Q})$ for an object
$\vec{x}_i$, the feature vector
is concatenated with $\mu_{Q}$ to form the input to the final
joint neural network
layers (here depicted in orange).
Again, weight sharing is used to learn only one scoring network.
For both neural networks, we treat the number of layers, units and embedding
dimensions as hyperparameters, which are to be optimized.

\begin{algorithm}[tb]
  \caption{\fatenet training algorithm}
  \label{alg:fate}

  \begin{algorithmic}[1]
    \Require
    \Statex Dataset $\cD = \{(Q_i, C_i)\}_{i=1}^N$ with $Q_i \subset \IR^d$
    \Statex Embedding network $\phi \colon \IR^d \fromto \IR^{d'}$, parametrized by $\theta_{\phi}$
    \Statex Utility network $U \colon \IR^d \times \IR^{d'} \fromto \IR$, parametrized by $\theta_{U}$
    \Statex Batch size $b \in \IN$, Number of epochs
    $E \in \IN$
    \Statex Step size schedule $\vec{\eta} = (\eta_1, \eta_2, \dots)$ with
    $\eta_i \in \IR_{>0}\ \forall i\in \IN$
    \Statex Loss function $L\colon \cC \times \bigcup_{k\in\IN}\IR^k \fromto \IR$
    \Statex
    \Procedure{Train-\fatenet}{$\cD, U_0, U_1, b, E, \vec{\eta}, L$}
    \State Initialize random weight vectors $\theta_{\phi}, \theta_U$
    \label{alg:fate:line:init}
    \For{Epoch $ep \in [E]$}
    \State $\cD \gets \Call{Shuffle}{\cD}$
    \State $T \gets \left\lceil\frac{N}{b}\right\rceil$
    \State Construct mini-batches $\cB_1, \dots, \cB_T$
    \label{alg:fate:line:mb}
    \For{Iteration $t \in [T]$}
    \State $\ell_t \gets 0$
    \ForAll{$(Q, C) \in \cB_t$}
    \State $\mu_Q \gets \frac{1}{|Q|} \sum_{\vec{x}\in Q}
      \phi(\vec{x})$
    \label{alg:fate:line:mu}
    \State $\vec{u} \gets
      \bigl(U(\vec{x},
      \mu_Q)\bigr)_{\vec{x} \in Q}$
    \label{alg:fate:line:ut}
    \State $\ell_t \gets \ell_t  + L(C, \vec{u})$
    \label{alg:fate:line:loss}
    \EndFor
    \State $\theta_U \gets \theta_U - \frac{\eta_{ep\cdot T+t}}
      {|\cB_t|}
      \nabla_{\theta_U} \ell_t$
    \label{alg:fate:line:gradU}
    \State $\theta_{\phi} \gets \theta_{\phi} -
      \frac{\eta_{ep\cdot T+t}}{|\cB_t|} \nabla_{\theta_{\phi}} \ell_t$
    \label{alg:fate:line:gradp}
    \EndFor
    \EndFor
    \EndProcedure
  \end{algorithmic}
\end{algorithm}
The detailed training algorithm is shown in \cref{alg:fate}.
As mentioned before for \fetanet, it is an instantiation of stochastic
gradient descent.
We will denote the weight vectors of the networks $U$
and $\phi$ by
$\theta_U$ and $\theta_{\phi}$, respectively.
The initialization of the weight vectors and the construction of the
mini-batches (lines \ref{alg:fate:line:init}--\ref{alg:fate:line:mb})
is
again the same as for \fetanet.
In line~\ref{alg:fate:line:mu}, the representative object
$\mu_Q$ is constructed by first
mapping each object to the embedding space using $\phi_{\theta_\phi}$,
and then computing the centroid of the embedded points.
The embedding network can be any network that receives an object and
returns a $d'$-dimensional real-valued vector, and should
be adapted to the
data at hand.
The utility scores $\vec{u}$ are then computed by evaluating
each object
$\vec{x} \in Q$ in conjunction with the representative point
$\mu_Q$
(see line~\ref{alg:fate:line:ut}).
The cumulative loss for the mini batch is updated in
line~\ref{alg:fate:line:loss}.
The weight vectors $\theta_{\phi}$ and $\theta_U$
are updated by calculating
the gradient of the loss using backpropagation and scaling it by an
appropriate learning rate
(lines~\ref{alg:fate:line:gradp}--\ref{alg:fate:line:gradU}).

The training runtime complexity per epoch of \fatenet (including backpropagation) is
$\mathcal{O}\left(N d q^2\right)$,
where $N$ denotes the number of choice tasks,
$d$ is the number of features
per object, and $q$ is an upper bound on the number of
objects in each task.
For a new choice task $Q$, the prediction can be done
in only
$\mathcal{O}(d |Q|)$ time (\thatis \emph{linear} in the number of objects).
This is due to the fact that $\mu_Q$ only needs to be computed once.

\section{Empirical Evaluation} %
\label{sec:eval}
The main goal of our empirical evaluation is to find out for which
kind of problems \fatenet and \fetanet work well.
Moreover, we wish to compare
these approaches with existing methods for ranking and choice.
In particular,	the following questions will be addressed:
\begin{itemize}
  \item Are the decompositions \ac{FATE} and
        \ac{FETA} suitable for learning context-dependent choice functions?
  \item How important is (i) the complexity/expressiveness of the underlying model
        class and (ii) its ability to model context-dependent choice functions, and how do these two
        factors interact?
        For example, are deep neural networks (i.\,e.\ \fatenet and \fetanet) really needed, or would a simpler (e.\,g.\ linear)
        model also suffice?
        Can the additional complexity/expressiveness compensate for the inability
        to model context-dependent choice functions?
  \item To what extent is our approach able to generalize over the task size?
        For example, is it possible to produce accurate predictions on tasks of a
        specific size, even if that size has never occurred in the training data?
\end{itemize}
For the first two questions, we evaluate the approaches on a variety of general
choice and \dc problems.
We also introduce the variant \fetalinear, which learns the
\ac{FETA} decomposition using only linear functions, to ascertain
whether it is able to account for some of the context-effects present in the data.

In addition, we evaluate the performance of different logit models used in
economics:
\acf{MNL} \citep{mcfadden1974mnl}, \acf{NL} \citep{Williams1977nlm},
\acf{GNL} \citep{wen2001generalized} and \acf{ML} \citep{train2009}.
The first logit model is the \ac{MNL} model (referred as \glm for subset choice task), which assumes that the choice between two objects does not depend on other objects in the
set~\citep{luce1959}. 
The \ac{NL} and \ac{GNL} belong to the \ac{GEV}
class of models that learn correlations amongst the objects in the given set,
which implicitly accounts for some of the context effects, but mainly the similarity effect~\citep{ben1985, tversky1972elimination}.
\ac{GEV} models allocate the objects in the given task
$Q$ into different sets called nests and learn
correlations between the objects inside each nest
\citep{wen2001generalized,train2009}.
These nests are disjoint in case of \ac{NL}~\citep{Williams1977nlm}.
\ac{GNL} is the most general model of this class, which allows the fractional
allocation of each object in $Q$ to each nest and it
learns the  correlation between them~\citep{wen2001generalized}.
\ac{ML} estimates the choice probability as a mixture of multiple logits~\citep{mcfadden2000mixed, yu2012four}.

Another model which was proposed for solving the task of singleton choice is the
\pairwisesvm, which makes use of induced pairwise preferences to fit a linear model~\citep{evgeniou2005generalized, maldonado2015advanced}.

As a recent context-dependent baseline model, we implement the \ac{SDA} approach by \citet{rosenfeld2020}.
We also implement the \ranknet model as an additional context-independent baseline, which learns a non-linear utility for each object by converting them to pairwise preferences~\citep{Tesauro1989,Burges2005}.
Due to a lack of algorithms specifically designed for the subset choice problem,
we employ the same thresholding of the utilities described in \eqref{eq:detsubset}
we use for our approaches. 
The threshold is tuned on a small validation set for all approaches,
using the $F_1$-score as target loss (see \cref{sub:experimentaldetails} for details).

All in all, we compare to both deep neural networks and linear models, so that we
have baselines of varying representative power, which helps to contextualize the
performance of our approaches on each dataset.
Finally, to answer the third question, we train the different models on a fixed
task size and predict on queries of deviating size.

\subsection{Setup} %
\label{sub:setup}

All experiments are implemented in Python, and the code and the dataset
generators are publicly available%
\footnote{\url{https://github.com/kiudee/cs-ranking}}.
To properly compare all models in a fair and unbiased way, we make
sure to optimize the hyperparameters of each model by employing
Bayesian optimization in a nested validation loop
(we use the Gaussian process based implementation in scikit-optimize~\citep{skopt})
The final out-of-sample estimates are then computed using another outer
cross-validation loop with the best hyperparameters found in each fold.
The loss functions and the datasets considered throughout our empirical evaluation
are introduced in the following two subsections, respectively (see~\cref{sub:experimentaldetails} for more details).

The experiments were run on a compute cluster with a mix of NVIDIA GTX 1080 Ti and RTX 2080 Ti GPUs (on average 15-20) and Intel Xeon E5-2670 processors.
One job consisting of one outer split with complete hyperparameter optimization on the validation set took on average \num{8} hours.
The training of \fatenet and \fetanet on average (across datasets) required \num{11} hours.
Combined, all experiments took roughly \num{11400} GPU hours and \num{6000} CPU hours.%

\subsection{Loss Functions} %
\label{sub:loss_functions}
As explained in \cref{sec:learning}, our goal during learning is to
minimize a suitable target loss $L\colon \cC \times \cC \fromto \IR$.
This is usually the loss one is interested in minimizing, e.\,g.,
the $F_1$-measure in our case.
Since these losses are usually not differentiable, they cannot readily
be used in a gradient descent algorithm.
Therefore, during training we opt to minimize surrogate losses which
are differentiable almost everywhere instead.
In this section, we will first introduce the target
losses we consider (cf. \cref{ssub:target_loss_functions}).
We then derive surrogate losses based on the probabilistic choice models
introduced in \cref{sec:probpref} and based on practical
considerations (cf. \cref{ssub:surrogate_loss_functions}).

\subsubsection{Target Loss Functions} %
\label{ssub:target_loss_functions}
The canonical loss function, which we focus on in the singleton choice setting, is the
\emph{categorical 0/1-loss}
\begin{equation}\label{eq:zerooneloss}
  L_{0/1}(C, C') \coloneqq 
  \indic{C \neq C'},
\end{equation}
\thatis in case the ground-truth choice $C$ is $\{\vec{x}\}$, each false prediction $C'\not=\{\vec{x}\}$ is penalized with a loss of $1$.
In addition, we will call the quantity
$1 - L_{0/1}(C, C')$
the \emph{categorical accuracy}.
Moving from singleton to subset choice, where $C$ and
$C'$
can now be choice sets of arbitrary size, the same loss function
\eqref{eq:zerooneloss} can still be used.
To signify that it is used in subset choice, we will call it the \emph{subset
$0/1$-loss}. Targeting the subset $0/1$-loss is problematic, especially
whenever a task $Q$ contains
many objects, since already one incorrectly predicted object results in
the whole prediction being declared incorrect.
One could instead opt to consider the average of the item-wise
$0/1$-loss, which is called the Hamming loss in the
setting of multi-label classification \citep{waegeman2012}.
However, this loss exhibits some properties that could be questioned in
the context of choice.
In particular, the non-prediction of a selected item (false negative) is
penalized in the same way as the prediction of a non-selected item (false
positive), although positives and negatives might be highly imbalanced.

A more suitable measure, which is widely used in classification, is the
\emph{$F_1$-measure} defined as  
\begin{equation}\label{eq:f1x}
  F_1(C, C') \coloneqq
   \frac{2 \sum_{\vec{x} \in \cX} \indic{\vec{x} \in C \cap C'}}{\sum_{\vec{x} \in \cX} \indic{\vec{x} \in C} + \sum_{\vec{x} \in \cX} \indic{\vec{x} \in C'}}
\end{equation}
for any $C,C' \in \cC$. This measure takes values in $[0,1]$ and large values indicate conformity between $C$ and $C'$, whence an appropriate loss can be defined as \footnote{Later on, we will nevertheless report the $F_1$-measure
  itself, which is common practice in machine learning.
}
\[
  L_{F_1}(C,C') \defeq 1 - F_1(C,C') .
\]
In spite of the existence of other measures that specifically aim at correctly
predicting positives,
such as the informedness \citep{powers2003,powers2011evaluation}, we
will mostly focus on $L_{F_1}$ as the target loss, because it
is well known and commonly used as a performance metric.
That means that we will use it as the validation loss for the
Bayesian hyperparameter optimization we run for every learner.
Additional evaluation measures we report are described in \cref{asub:evaluation_metrics}.

\subsubsection{Surrogate Losses} %
\label{ssub:surrogate_loss_functions}

The probabilistic setting for choice that we introduced in \cref{sec:probpref} suggests a natural approach to learning and
prediction:
\begin{itemize}
  \item First, a learner is trained using the log-likelihood of the probabilistic model
        as a loss function.
        This loss function is not only differentiable, but also calibrated in the sense
        of being minimized by the true (conditional) probabilities.
        In other words, a learner trained with this loss is supposed to predict
        (unbiased) probabilities on the choice space $\cC$
        (conditioned on the query).
  \item Thus,  given a query for which a prediction is sought, a probability
        distribution on the choice space $\cC$ can be obtained
        as a prediction, which in turn allows for minimizing any target loss in
        expectation.
\end{itemize}
More specifically, let $U(\cdot,Q)$ denote the latent utility scores  $U(\vec{x},Q)$, ${\vec{x} \in Q}$, predicted by a learner on a query $Q\in \cQ$. In a singleton choice scenario, where the data is supposed to be generated according to choice probabilities
${\prob^{\tilde{U}}_{\text{MNL}}(\vec{x}\mid Q)} = \prob_{\text{MNL}}(\vec{x}\mid Q)$
of the form \eqref{eq:mnl} for some unknown ground-truth~$\tilde{U}$, one may define the corresponding categorical cross-entropy loss gained when observing~${C=\{\vec{x}\} \in \cC}$ 
\begin{align}
  L_{\text{CE}} \big(\{\vec{x}\}, U(\cdot,Q) \big) &\coloneqq -\log\big( \prob^{U}_{\text{MNL}}(\vec{x}  \mid Q) \big) \notag \\
  &\hspace{2pt}= \log \left( \sum\nolimits_{\vec{y} \in Q} \exp(U(\vec{y},Q)) \right) -  U(\vec{x},Q)\ .\label{eq:loglosssingletonchoice}
\end{align}
This expression is minimized in case $\vec{x} = \argmax_{\vec{y} \in Q} U(\vec{y},Q)$.

If dealing with subset choice data that is presumably sampled according to the choice probability distribution $\prob^{U}(C\mid Q) = \prob(C\mid Q)$ from \eqref{eq:likebin}, it is natural to measure  prediction $C\in 2^Q \setminus \{\emptyset\}$ by means of the corresponding binary cross-entropy loss
\begin{align} 
  L_{\text{BE}}(C, U(\cdot,Q)) &\coloneqq - \log\big( \prob^{U}(C  \mid Q) \big) \notag \\
  &\hspace{2pt}=  \sum\nolimits_{\vec{y} \in Q} \log \big(1+\exp(U(\vec{y}, Q)) \big) - \indic{\vec{y} \in C} U(\vec{y}, Q). \label{eq:loglosschoice}
\end{align}

In spite of the theoretical justification of the logistic losses discussed
above, we found that ``hinge-variants'' of the respective 0/1-losses may
sometimes lead to more stable results.
More specifically, for the singleton choice setting
\emph{categorical hinge loss} defined via 
\begin{equation}\label{eq:categoricalhinge}%
  L_{\text{CH}} (\{\vec{x}\},
  U(\cdot,Q)) \coloneqq \max \Bigl(1
  + \max\nolimits_{\vec{y} \in Q \setminus \{\vec{x}\}} U(\vec{y},Q) - U(\vec{x},Q), 0 \Bigr) \, ,
\end{equation}
for any $\vec{x} \in Q \in \cQ$,
is inspired by the hinge loss used in multi-class
classification~\citep{dogan2016unified, moore2011reg} and can be used instead of
\eqref{eq:loglosssingletonchoice}.

Finally, for training \fatenet and \fetanet in the experiments below, we use the
\emph{binary cross-entropy} loss for the subset choice setting and the 
\emph{categorical hinge} loss for the \dc setting,
since these turned out to work well in preliminary experiments.
In addition, an $L_2$-regularization term for the magnitude of the weights
is added and optimized as part of the loss during training.

\paragraph{Convexity of the Surrogate Losses}
An important consideration for the surrogate losses to be used during training is
whether they are convex with respect to the utility scores $U(\vec{x}, Q)$.
All three losses introduced above are indeed convex.
To see this for $L_{\text{CE}}$, notice that \eqref{eq:loglosssingletonchoice} can
equivalently be written as
$ \log \bigl( \sum\nolimits_{\vec{y} \in Q} \exp(U(\vec{y},Q) -  U(\vec{x},Q)) \bigr) $.
The inner difference of utilities is linear and therefore convex.
The outer function is also known as \emph{LogSumExp} and is defined via
$\operatorname{LSE}(\vec{x}) \coloneqq \log(\sum_{j\in [m]} \exp(x_j))$.
It is convex and since it is also strictly decreasing in each argument,
the composition \eqref{eq:loglosssingletonchoice} is convex as well.

As for the binary cross-entropy $L_{\text{BE}}$, note that the inner function
$s\colon \IR \to \IR$, $s(x) \coloneqq \log(1 + \exp(x))$ 
of \eqref{eq:loglosschoice} is smooth with 
strictly positive first and second derivatives
and hence convex and non-decreasing.
Similarly, $\tilde{s}(x) \coloneqq s(x) - x$ 
is convex and strictly decreasing on $\IR$.
Hence, we can conclude that \eqref{eq:loglosschoice} is convex.

Finally, the categorical hinge \eqref{eq:categoricalhinge} contains the function
$h\colon \IR^m \to \IR$, $\vec{x} \mapsto \log\bigl( \sum_{j\in [m]} \exp(x_j - x_i)\bigr)$,
which is convex as the logarithm of a maximum of convex functions.
Since $s\colon \IR \to \IR$, $x \mapsto \max(1+x, 0)$ is convex and non-decreasing,
$s \circ h$ and therefore \eqref{eq:categoricalhinge} is convex as well.

The FETA model further decomposes $U(\vec{x}, Q)$ into an aggregation of sub-utility functions
$U_0$ and $U_1$.
It is therefore interesting to ask whether the surrogate losses are
also convex with respect to the sub-utility values $U_{0}(\vec{x})$, $U_{1}(\vec{x},\{\vec{y}\})$.
We can answer this question in the affirmative, since the
FETA utility values are positively weighted sums of these sub-utility scores.

However, the overall learning problem depends on the parameter $\theta$ of the realization of $U_{\text{FETA}}$ and $U_{\text{FATE}}$ and the corresponding loss function can possibly still be non-convex w.r.t. $\theta$ (as this is the case with the neural networks employed here).
That means in practice we lose the guarantee of stochastic gradient descent to find a global optimum, but with careful tuning of the optimization process one can still expect to find reasonable solutions.

\subsection{Datasets} %
\label{sub:datasets}
We now introduce the learning problems used for the empirical comparison as follows:
\begin{table}[tb]  %
  \centering
  \caption{Overview of the choice datasets used in the
    experiments.
    Bracket notation is used to denote the range of values.
  }
  \label{tab:datasets:overview}
  \sisetup{
    table-figures-decimal = 0,
    table-figures-exponent = 0,
    table-number-alignment = center
  }
    \begin{tabular}{
        ll%
        S[table-figures-integer=6]
        S[table-figures-integer=7]
        S[table-figures-integer=3]
        S[table-figures-integer=2]}
      \toprule
      Problem & Dataset          & {\#\,Train}     & {\#\,Test}    & {\#\,Features} & {$\card{Q}$}   \\
      \midrule
      \multirow{10}{*}{Singleton Choice}
              & Medoid           & \num{10000}     & \num{100000}  & \num{5}        & \num{10}       \\
              & Hypervolume      & \num{10000}     & \num{100000}  & \num{2}        & \num{10}       \\
              & MNIST-Mode       & \num{10000}     & \num{100000}  & \num{128}      & \num{10}       \\
              & MNIST-Unique     & \num{10000}     & \num{100000}  & \num{128}      & \num{10}       \\
              & \mdm             & \num{10000}     & \num{100000}  & \num{1128}     & \num{10}       \\
              & \msm             & \num{10000}     & \num{100000}  & \num{1128}     & \num{10}       \\
              & LETOR-\mql{2007} & ${[1353,1356]}$ & ${[336,339]}$ & \num{46}       & ${[257,1346]}$ \\
              & LETOR-\mql{2008} & ${[627,628]}$   & ${[156,157]}$ & \num{46}       & ${[204,1831]}$ \\
              & Expedia          & \num{78041}     & \num{312229}  & \num{17}       & ${[5,38]}$     \\
              & Sushi            & \num{7000}      & \num{3000}    & \num{7}        & \num{10}       \\

      \midrule
      \multirow{6}{*}{Subset Choice}
              & Pareto-front-\num{2}D & \num{10000}     & \num{100000}  & \num{2}        & \num{30}       \\
              & Pareto-front-\num{5}D & \num{10000}     & \num{100000}  & \num{5}        & \num{30}       \\
              & MNIST-Mode       & \num{10000}     & \num{100000}  & \num{128}      & \num{10}       \\
              & MNIST-Unique     & \num{10000}     & \num{100000}  & \num{128}      & \num{10}       \\
              & LETOR-\mq{2007}  & ${[1160,1172]}$ & ${[283,295]}$ & \num{46}       & ${[6,147]}$    \\
              & LETOR-\mq{2008}  & ${[442,459]}$   & ${[105,122]}$ & \num{46}       & ${[5,121]}$    \\
              & Expedia          & \num{79855}     & \num{319489}  & \num{17}       & ${[5,38]}$     \\
      \bottomrule
    \end{tabular}
\end{table}

\noindent
\begin{enumerate}[label=(\alph*)]
  \item The Medoid problem, where the task is to predict the medoid of
        a set of points in a Euclidean space.
  \item The Pareto-front problem, in which the learner has to predict the set
        of points which are Pareto-optimal.
  \item The Hypervolume singleton choice problem, where the task is to select
        the point of the Pareto-front which contributes the most to the
        hypervolume.
  \item Different choice problems defined on the well-known MNIST dataset.
  \item Similarity/dissimilarity-based movie selection using the MovieLens Tag
        Genome dataset \citep{vig2012tag}.
  \item The \ac{LETOR} \mq{2007} and
        \mq{2008} datasets \citep{letor2013} consisting
        of query-document pairs, with the goal to select the relevant
        documents.
  \item The Expedia hotel dataset featuring search results and relevance
        labels for each hotel with the goal to select booked/considered hotels \citep{Expedia2016}.
  \item The Sushi dataset, where the task is to choose the most preferred
        sushi from a set of $10$ options provided to a user.
\end{enumerate}
See \cref{tab:datasets:overview} for an overview of the datasets and their
properties.
In the following sections, we will describe the different datasets, their
motivation, and if applicable, how they are generated.

\subsubsection{The Medoid Problem} %
\label{sub:medoid}

The motivation for this problem is the general idea of learning to choose a
most representative element from a set.
More concretely,
the medoid of a set is the object with the smallest
cumulative dissimilarity to all other objects of the set%
\footnote{As opposed to the centroid, which is usually not part of the original set.}.
It is commonly used as a representative element, especially for structured
objects such as graphs, $2$-D
trajectories, images, etc.\ \parencite{van2003new, zhang2005}.

Formally, we are interested in learning the choice function $c_{\operatorname{medoid}} \from \cQ \to \cC$ given as 
\begin{equation*}
    c_{\operatorname{medoid}}(Q) \coloneqq \argmin\nolimits_{\vec{x} \in Q} \frac{1}{|Q|} \sum\nolimits_{\vec{y} \in Q} \norm{\vec{x}-\vec{y}},
\end{equation*}
where we write here and throughout the remainder of this paper $\norm{\cdot}$ for the standard euclidean norm defined as $\norm{\vec{z}} = \sqrt{\vec{z}^{t}\vec{z}}$.
The \dc produced by this procedure incorporates all
pairwise distances among the objects, which makes it a good
context-dependent learning problem to investigate.
In particular, $c_{\operatorname{medoid}}$ is sensitive to
changes of the elements in the task. 
With $U_{0}(\vec{x}) \coloneqq 0$ and $U_{1}(\vec{x},\{\vec{y}\}) \coloneqq -\norm{\vec{x} - \vec{y}}$ we clearly have 
\begin{align*}
    c_{\operatorname{medoid}}(Q) &= \argmin_{\vec{x} \in Q} \frac{1}{|Q|-1} \sum_{\vec{y} \in Q} \norm{\vec{x}-\vec{y}}\\
    &= \argmax_{\vec{x} \in Q} U_{0}(\vec{x}) + \frac{1}{|Q|-1} \sum_{\vec{y} \in Q\setminus \{\vec{x}\}} U_{1}(\vec{x},\{\vec{y}\})
\end{align*} and thus  $U_{\text{FETA}}^{U_{0},U_{1}}$  is able to  exactly model $c_{\operatorname{medoid}}$.

In contrast to this, for the \ac{FATE} approach, it is not immediately obvious
if and how it is capable of modelling $c_{\operatorname{medoid}}$ exactly.
However, the choices 
$\mathcal{Z} \coloneqq \mathcal{X}$, $\phi \coloneqq \mathrm{id}_{\mathcal{X}}$ and $U'(\vec{x},\vec{z}) \coloneqq -\norm{\vec{x}-\vec{z}}$ yield
\begin{equation*}
	U_{\mathrm{FATE}}^{U',\phi}(\vec{x},Q) = -\norm{\vec{x}-\operatorname{centroid}(Q)}
\end{equation*}
with $\operatorname{centroid}(Q) \coloneqq \frac{1}{|Q|} \sum_{\vec{y}\in Q} \vec{y}$ being the centroid of $Q$.
Thus, the item $\vec{x} \in Q$, which is closest to $\operatorname{centroid}(Q)$,
\thatis $\argmax_{\vec{x} \in Q} U_{\mathrm{FATE}}^{U',\phi}(\vec{x},Q)$,
is likely to coincide with the medoid of $Q$.
As we construct our synthetic medoid dataset by sampling $Q$ according to the uniform distribution $\nu$ on $\{A\subseteq [0,1]^{d} \sothat |A| = r\}$ for some predefined $r\in \N$,  there is with $U_{\mathrm{FATE}}^{U',\phi}$ a \ac{FATE}-instance, which is expected to have (for the case of singleton choice) an accuracy of at least
\begin{equation*}
	\Prob_{Q\sim \nu} \left(c_{\operatorname{medoid}}(Q) = \argmin\nolimits_{\vec{x}\in Q} \norm{\vec{x}-\operatorname{centroid}(Q)} \right)
\end{equation*}
on the synthetic medoid dataset.
An empirical evaluation revealed that this value is \SI{89.56}{\percent} for $r=10$ and $d=5$.
For the details on this dataset, confer \cref{sub:dgp:medoid}.

\subsubsection{The Pareto-Front Problem} %
\label{sub:pareto}
The computation of a Pareto-optimal set of points is an important problem
in optimization and various fields of application~\citep{geilen2007algebra}.
We say $\vec{x}\in \mathcal{X} \subseteq \IR^{d}$ is \emph{dominated by $\vec{y} \in \IR^{d}$} (short: $\vec{y}\succ \vec{x}$) if $x_{i} \leq y_{i}$ holds for any $1\leq i\leq d$ and $x_j < y_j$ for at least one $1 \leq j \leq d$.
For any set $Q \in \cQ$ we define the \emph{Pareto-set} or \emph{Pareto-front} of $Q$ as
\begin{equation*}
	c_{\mathrm{Pareto}}(Q) \coloneqq \{ \vec{x} \in Q \sothat \vec{x} \text{ is not dominated by any element } \vec{y}\in Q\setminus \{\vec{x}\}\}.
\end{equation*}	
We wish to investigate the possibility to learn the mapping from sets of points
to their respective Pareto-sets.
It is clear that the size of the Pareto-sets is not constant, which makes
it a good candidate for a general subset choice problem.
With the choices $U_{0}(\vec{x}) \coloneqq 0$ and $U_{1}(\vec{x},\{\vec{y}\}) \coloneqq -\indic{\vec{y} \succ \vec{x}}$ we have 
\begin{align*}
		U_{\text{FETA}}^{U_{0},U_{1}}(\vec{x},Q) = - \sum\nolimits_{\vec{y} \in Q} \indic{\vec{y}\succ \vec{x}} \in \begin{cases} (-\infty,-1], \quad &\text{if } \vec{x} \not\in c_{\mathrm{Pareto}}(Q), \\ \{0\},\quad &\text{otherwise.}\end{cases}. 
\end{align*}
Hence, $c_{\mathrm{Pareto}}(Q) = \argmax_{\vec{x}\in Q} U_{\mathrm{FETA}}^{(U_{1})}(\vec{x},Q)$ holds trivially for each $Q \in \cQ$, \thatis the Pareto problem is exactly solvable via the FETA approach.
We created our corresponding synthetic dataset
by generating a set of points uniformly at
random in $\IR^2$ and $\IR^5$ to construct a \cproblem
$Q$, and the ground-truth is
the Pareto-set of $Q$ containing only the non-dominated
objects.
In order to perform the experiments, we generate sets of
\num{30} random
points in $\IR^2$ and $\IR^5$, and determine the choices as described in
detail
in~\cref{sub:dgp:pareto}.

\
\color{black}

\subsubsection{Hypervolume}
\label{subsec_Hypervol}
A related but much harder problem is the computation of hypervolume
contributions of objects on a Pareto front.
The hypervolume $\lambda_{\mathrm{HypVol}}(Q)$ of a subset $Q\subseteq \IR^{d}$ describes the volume of the union of the subspaces dominated by each individual point $\vec{x} = (x_{1},\dots,x_{d})$ in the Pareto set of $Q$ and can formally be defined as 
\begin{align*}
    \lambda_{\mathrm{HypVol}}(Q) &\coloneqq \lambda\left( \bigcup\nolimits_{\vec{x} \in c_{\mathrm{Pareto}}(Q)} [0,x_{1}] \times \dots \times [0,x_{d}]\right) \\
    &= 
    \lambda\left( \bigcup\nolimits_{\vec{x} \in Q} [0,x_{1}] \times \dots \times [0,x_{d}]\right)
\end{align*}  
where $\lambda$ denotes the Lebesgue measure of $\IR^{d}$. In the context of \acp{MOEA}, one usually computes the
contributions $\lambda_{\mathrm{HypVol}}(Q) - \lambda_{\mathrm{HypVol}}(Q\setminus \{\vec{x}\})$ of
each point $\vec{x} \in Q$ to the overall hypervolume $\lambda_{\mathrm{HypVol}}(Q)$,
\thatis the reduction in hypervolume caused by removing one object from
the set. We consider the problem of learning the corresponding Hypervolume choice function $c_{\mathrm{HypVol}} \from \cQ \to \cC$, which picks that element $\vec{x} \in Q$ with the smallest contribution to the overall hypervolume, \thatis 
\begin{align*}
    c_{\mathrm{HypVol}}(Q) &\coloneqq \argmax_{\vec{x} \in Q} \lambda_{\mathrm{HypVol}}(Q) - \lambda_{\mathrm{HypVol}}(Q\setminus \{\vec{x}\}) \\
    &\hspace{2pt}= 
    \argmin_{\vec{x} \in Q} \lambda_{\mathrm{HypVol}}(Q\setminus \{\vec{x}\}).
\end{align*}
As shown by \citet[Theorem 1]{bringmann2010}, it is \sharpP-hard to calculate $c_{\mathrm{HypVol}}(Q)$.
Here, we generate sets of \num{10} random points in
$\IR^2$ and determine
the singleton choice.

\subsubsection{MNIST Number Problems} %
\label{sub:mnist}
The original goal of the \ac{MNIST} dataset was to facilitate the
comparison between different handwritten digits
classifiers~\citep{mnisthandwrittendigit}.
It consists of \num{70000} $28 \times 28$
grayscale images.
We use the dataset to create challenging choice problems, both singleton and
general subset choice.
To level the playing field between all the approaches, we first train a
\ac{CNN} on \num{10000} instances and use
it to extract high level
features for the remaining \num{60000} images
(see~\cref{sub:dgp:mnist} for more details).
To convert this dataset to a choice problem, we randomly sample sets of
\num{10} numbers and choose based on the following
procedures:
\begin{enumerate}
  \item \textbf{Mode}: For the Mode dataset, we choose the numbers that
        occur most often in the \cproblem $Q$.
        For example, given a set of numbers $\{1,$
        $1$, $2$,
        $4$, $4$,
        $5$, $5$,
        $6$, $6$,
        $6\}$,
        we choose all instances with value equal to the mode value
        $6$.
        For the \dc task, we only output one of the numbers (the representation of which has the least
        angle to a predefined vector).
  \item \textbf{Unique}: Here, we choose all numbers that occur
        only once in the set of sampled label values.
        For example, given a set of numbers $\{1$,
        $1$, $2$,
        $3$, $4$,
        $4$,
        $5$, $5$,
        $6$, $6\}$, we
        choose the numbers $\{2, 3\}$.
        For the \dc problem, we ensure that exactly one of the digits is unique.
\end{enumerate}

\subsubsection{MovieLens Tag Genome} %
\label{sub:taggenome}
The MovieLens Tag Genome dataset consists of a large collection of
movies and community curated tags~\citep{vig2012tag}.
For each movie, the relevance of every tag is provided on a continuous
scale in $[0,1]$.
Thus, the complete relevance vector of a movie can be regarded as that
movies' ``genome.''

We consider the problem of choosing the most similar/dissimilar movie from
a set of movies, where one movie is regarded as the reference to
which the others are compared.
We define this reference movie to be the medoid of the movies in a given
set.
To compute similarities in tag relevance space, we use the weighted cosine
similarity as proposed by
\citet{vig2011navigating}.

\subsubsection{LETOR} %
\label{sub:letor}
\ac{LETOR} is a collection of benchmark datasets for
different learning-to-rank problems \citep{letor2013}.
The Gov2 web page collection, consisting of roughly 25\,M pages, is the corpus
and the query sets of the Million Query track of the TREC
\num{2007} and \num{2008}
\citep{Trec2007,Trec2008} are used to create \num{8}
datasets.
Each query-document pair is defined by a vector consisting of
\num{46} features.
We use the supervised ranking datasets \mq{2007} and
\mq{2008} to create the choice dataset.
We treat all documents with a relevance score of 1 and 2 as the chosen
objects.
Since all queries include multiple documents with relevance scores
\num{1} and \num{2}, we cannot
extract \dcs from this dataset.
The listwise ranking datasets \mql{2007} and
\mql{2008} contain real-valued scores of the documents in the
underlying permutations, and hence facilitate the \dc for each query
(details of the exact procedure can be found in
\cref{asub:letor}).

\subsubsection{Expedia} %
\label{sub:expedia}
The Expedia dataset was released on the Kaggle website as a competition in 2016
\citep{Expedia2016}.
It consists of \num{399344} lists of hotels, each resulting
from a search query of a user.
For each hotel, there are \num{45} features and a relevance
score, indicating how relevant the hotel is to the provided query.
A score of \num{0} means that it was not relevant, a
score of \num{1} indicates that the user clicked on it,
and a \num{2} implies that the hotel was booked.
It is straightforward to construct choice datasets: for singleton choice the
goal is simply to predict the booked hotel, whereas for subset choice we
required the learners to output the complete set of hotels that were at least
clicked on (see \cref{asub:expedia} for more details).

\subsubsection{SUSHI} %
\label{sub:sushi}
SUSHI\footnote{This dataset can be downloaded from \url{http://www.kamishima.net/sushi/}}
is a dataset created by
\citet{kamishima03} specifically for the task of \emph{object ranking}.
The authors considered \num{100} sushis and asked users to rank them according to
their preference.
The dataset consists of two sets of \num{5000} rankings.
Each ranking consists of \num{10} sushis, which were ranked by users in a survey.
For the first set, the authors asked the users to rank the top-10 most popular
sushis.
In the second set, users were shown random sets of \num{10} sushis instead.
Each sushi is described by \num{7} object features.
Additional user features are available, but not used in our experiments.
For our experiments, we merge both datasets into a single one containing \num{10000}
instances. We use it as a \dc dataset by choosing the most preferred sushi as
the \dc for the given \qset $Q$ (details of the exact procedure can be
found in \cref{asub:sushi}).

\subsection{Results and Discussion} %
\label{sub:results}
In this section, we provide the results obtained by evaluating different
subset choice and \dc models on the datasets.
To be concise, we only show plots for the target losses here and list the
complete set of results in
\cref{tab:choicemodels,tab:singletonchoice,tab:singletonchoice2}
in \cref{asub:results}.
It is illuminating to compare the performance of \fatenet and
\fetanet to
the context-independent neural network \ranknet.
This provides a rough indicator for how important being able to model
context-dependence is.

\begin{figure}[t]  %
  \centering
  \includegraphics{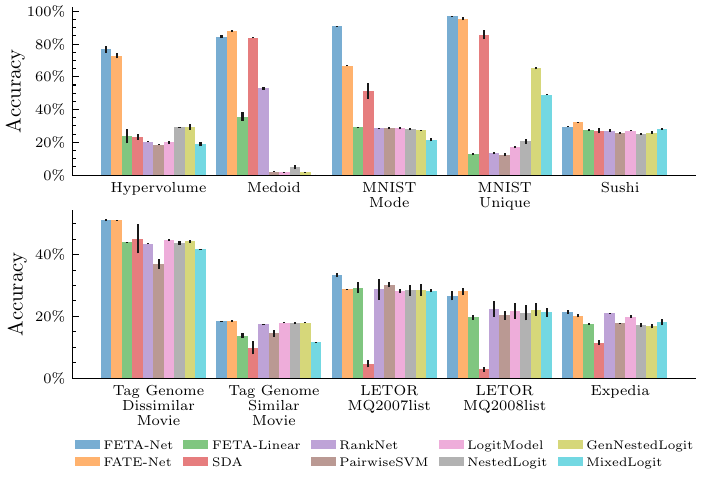}
  \caption{Categorical accuracies and standard deviations (vertical bars) of the \dc
    models on different \dc tasks (measured across \num{5} outer
    cross-validation folds).}
  \label{fig:singletonchoice}
\end{figure}

\subsubsection{Singleton Choice}

We will start by discussing the results for the singleton choice models
(cf. \Cref{fig:singletonchoice}), where the bars depict the mean value of
the \textlower{\catacc} \eqref{eq:accuracy} across the cross-validation folds, with black
lines depicting the standard deviation.

The first observation is that \fatenet and \fetanet significantly outperform
all other baselines on the tasks for which it was clear that the underlying
choice function is context-dependent (\thatis Hypervolume, Medoid and the
MNIST datasets).
The \ac{SDA} network, which is also a context-dependent model, achieves
competitive results on the Medoid and the MNIST datasets.
The linear \ac{FETA} variant \fetalinear non-linear neural network \ranknet
perform comparably to the other baseline approaches.
This suggests that a combination of non-linearity and
the ability to model context-dependence is really necessary to improve on these tasks.
One notable exception is the Medoid dataset, for which \ranknet and
\fetalinear manage to outperform the other baselines by a large margin.

For the MNIST-Unique problem, \fatenet and \fetanet achieve an accuracy of
more than \SI{90}{\percent} and \ac{SDA} is competitive with over \SI{80}{\percent}.
Additionally, the \ac{GNL} and \ac{ML} models are also able to perform better
than the other baselines.
It is easy to see that the dataset exhibits the similarity
context effect proposed by \citet{huber1983market},
\thatis adding multiple instances of the same digit to the choice task reduces the choice probability of all equal digits to 0.
As is apparent, the \ac{GNL} and \ac{ML} model are able to account for it
and score better than chance.

\begin{figure}[tp]
    \centering
    \includegraphics{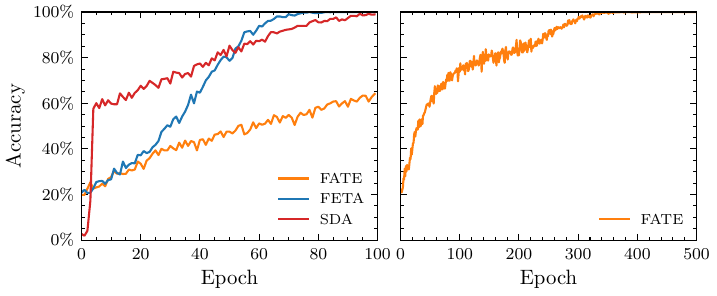}
    \caption{Result of the infinite data experiment for \fatenet, \fetanet and \ac{SDA}
    on the synthetic unique problem.
    For the left plot the neural networks were calibrated to have roughly equal numbers
    of parameters.
    The right plot shows the repetition of the experiment where \fatenet received
    a higher epoch and parameter budget.}
    \label{fig:plot_unique}
\end{figure}
Since \fatenet, \fetanet and \ac{SDA} were able to achieve close to
\SI{100}{\percent} accuracy on the MNIST-Unique problem,
we performed an additional experiment where we generated instances
completely synthetically. Each number $i\in \{0, \dots, 9\}$ we represent by the 
corresponding standard unit vector $\vec{e}_i$, which is $1$ in the $i$-th
position and is $0$ everywhere else.
Apart from that, the task remains the same.
We calibrate each network to have roughly the same number of parameters
(\num{2870} for \fatenet, \num{2849} for \fetanet and \num{2850} for \ac{SDA})
and the remaining hyperparameters were equal for all networks.
We then trained them on a stream of newly generated batches with 1024 instances,
each of which with 10 objects until convergence.
The resulting convergence behavior is shown in \cref{fig:plot_unique}.
Both \fetanet and \ac{SDA} are able to converge to
\SI{100}{\percent} out-of-sample categorical accuracy within 100 epochs, while \fatenet
only achieves slightly over \SI{60}{\percent} and
more epochs alone were not able to let it learn the
target function without error.
We therefore repeated the experiment for
\fatenet with a higher epoch and parameter budget.
With \num{5985} parameters, \fatenet is now able
to perfectly learn the fully synthetic unique
problem within 400 epochs.
On the one hand, this shows that from a
representational perspective, all three models
are able to learn this particular target choice
function perfectly.
\fatenet appears to be less parameter- and data-efficient though, which could indicate that evaluating the
utilities in the context of the set embedding is not well suited to represent these kinds of problems.
The behavior of all three networks was consistent across repetitions of the experiment.

On the real-world datasets (i.\,e.
Sushi, Movielens Tag Genome, LETOR and
Expedia) the performance of \fatenet and \fetanet is closer to the ones
achieved by the remaining baselines.
Although they still obtain slightly higher accuracy on average, the margin is
not as pronounced.
Surprisingly, the \ac{SDA} achieved the worst accuracy on LETOR and Expedia.
We suspect that this results from the models being trained only on
a fixed choice task size in our experiments, while they are evaluated on choice tasks of varying size during test time.
Since \ac{SDA} learns a set-dependent aggregation function, it could be
that this does not generalize well to the larger choice tasks present
in the real-world datasets.

\subsubsection{Subset Choice}

\begin{figure}[t]  %
  \centering
  \includegraphics{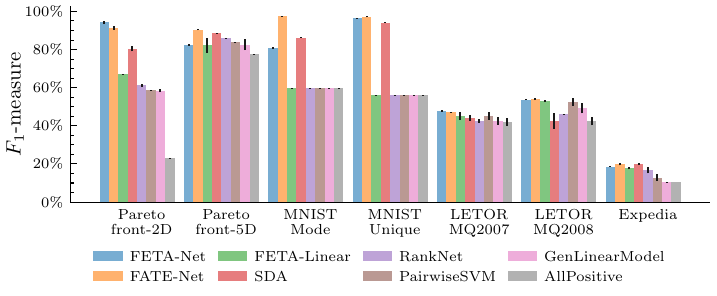}
  \caption{Average $F_1$-measure and standard deviation (vertical
    bars)
    of the subset-choice models on the tasks different choice tasks
    (measured across \num{5} outer cross-validation folds).}
  \label{fig:choicemodels}
\end{figure}

We evaluate the subset choice models in terms of their
$F_1$-measure \eqref{eq:f1x}
and report the results in \cref{fig:choicemodels}.
To see if the models are able to learn anything,
we also show the performance of the baseline that always predicts positive.
The general pattern is confirmed: \fatenet, \fetanet and \ac{SDA} surpass the other baselines on the datasets Pareto-front 2D, MNIST Mode, and MNIST Unique,
while being competitive for the real-world datasets LETOR and Expedia.
For the \acs{MNIST} tasks Unique and Mode, the first
observation is that all linear and/or context-independent baseline approaches fail to learn anything on these
datasets, since they all achieve the same $F_1$-measure
as the all-positive baseline.
Thus, it is clear that these tasks can only be solved by models that are
both context-dependent and non-linear.

For the Pareto problem, it can be observed that the context-dependent models \fetanet, \fatenet, and \ac{SDA} outperform all benchmark choice models on the 2D version.
On the 5D version of the dataset, however, the performance of all approaches reach a comparable level.
This indicates that solving the task of selecting the Pareto-front becomes
less context-dependent in higher dimensions, since the distance of a point
from the center becomes more and more informative.
At the same time, more points are on the Pareto-front overall, which
is apparent from the high $F_1$-measure of the AllPositive baseline.

As before, the results are more homogeneous on the real-world datasets Expedia
and \ac{LETOR}
\mq{2007}/\mq{2008}.
\fatenet and \fetanet are still outperforming all the benchmarks.
This suggests that the ability to model context-dependence in
the data is slightly more important for these datasets than learning a
non-linear utility function.
\ac{SDA} achieves the best result on the Expedia dataset, which
when compared to the bad performance on the \dc variant of the dataset
suggests that the thresholding of the utilities is robust to
the model output changing with varying choice task sizes.

Overall, the results demonstrate that \fatenet and \fetanet are able to
improve on the context-independent baselines by a large margin on tasks which are
strongly context-dependent and show competitive results when compared to
\ac{SDA}.
The improvement is due to both the task-sensitivity of these models and
the ability to model non-linear utility functions.
For the real-world datasets, the improvements are smaller, suggesting that
context-effects are either less pronounced or that the context-effects
in real-world data cannot fully be captured yet.

\subsubsection{Generalization Across Task Sizes}
\label{sec:otherresults}

\begin{figure*}[tb]
  \centering
  \includegraphics[width=0.9\linewidth]{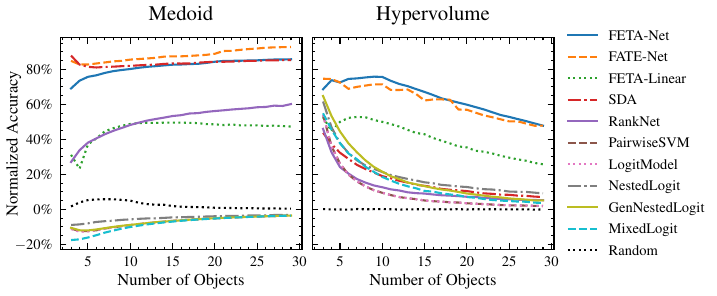}
  \caption{Normalized Accuracy of the \acp{SCM} trained on queries of
    size \num{10}, then predicting on queries of a varying size.}
  \label{fig:generalizationdc}
\end{figure*}

We conduct additional experiments to gauge the generalization
capability of the learned models to unseen task sizes (refer to~\cref{sub:experimentaldetails:genexp} for more details).
We show the results for the datasets Medoid and Hypervolume, because, as will be seen, they exhibit some interesting properties.
We specifically compare the performance on the \dc datasets
(\cref{fig:generalizationdc}).
We train the models on a fixed task size and then test them on sets
containing between \num{3} and
\num{21} objects.
Note that for singleton choice, the accuracy is not comparable across
differing task sizes.
We instead report the \emph{normalized accuracy} (see \cref{asub:metrics}),
which fixes this issue and guarantees that random guessing achieves
exactly~0.

Overall, the models manage to generalize quite well to
task sizes for which they were not trained.
The exact generalization behavior depends on the dataset, though.
Considering the Medoid dataset, we can observe that the models
\fetanet, \fetalinear and \ranknet even improve in performance with
larger task sizes.
This is plausible, since the more points fill the space, the more the
problem can be solved by a context-independent model, which assigns the
highest score to objects in the center.
For the singleton choice version of Hypervolume, on the other hand, the
performance of all models drops with an increasing numbers of objects,
suggesting it becomes much harder to identify the object that contributes the
most to the overall hypervolume.
This is especially visible for the baselines, which, even though they were
trained on 10
objects, achieve their best performance on 3 objects.
\fetanet, \fatenet, and \fetalinear stand out here, since their performance
decays much slower.
All in all, we conclude that our networks \fetanet and \fatenet are able to
generalize very
well to unseen task sizes, with \fetanet additionally benefiting if the task
becomes less context-dependent with larger task sizes.

\section{Conclusion and Future Work} %
\label{sec:conclusion}
In this paper, we tackle the problem of choice from a machine learning perspective. More specifically, we propose a framework for learning context-dependent choice functions, which, on the basis of choice behavior observed in the past, allow for predicting the choice of objects in new situations. This is essentially accomplished by learning generalized (latent) scoring (utility) functions, which are supposed to control the choice behavior.

Violations of context-independence are common in human choice behavior. Therefore, accounting
for the various context effects they can exhibit can be seen as an important problem.
Still, we consider the space of interesting non-trivial choice functions to
be vastly larger, and the goal is to have general purpose models that
can adapt to a wide variety of (yet unknown) context effects.

To this end, we propose two principled decompositions:
The \ac{FETA} decomposition is a first-order approximation to a
more general utility decomposition. It considers each object in \emph{local}
sub-contexts, the contributions of which are averaged.
The \ac{FATE} approach, on the other side, first transfers each object
into an embedding space and computes a representative of the choice task
by averaging these embedded points.
The utility of each object is then evaluated with the representative as
\emph{global} context.
Both approaches are complementary and have differing inductive
biases. In spite of this, both show promising predictive performance.

While the \ac{FETA} and \ac{FATE} decompositions are general and in a sense quite natural approaches to model context-dependent choice functions, a promising direction is the investigation of application-specific models with more focused inductive biases.
An example is the \ac{SDA} approach, which applies principles from behavioral
choice theory and also tries to take the risk-aversion of humans into account
\citep{rosenfeld2020}.

While the most influential context effects for human choices have been studied, 
gaining a deeper understanding of the rich mathematical structure of general
choice problems is an important future endeavor.

\section*{Acknowledgements}
The authors gratefully acknowledge the financial support provided by the European Regional Development Fund (ERDF) and the valuable feedback provided by the industry partners of the Smart-GM research project -- EFRE-0801915.

Funded by the Deutsche Forschungsgemeinschaft (DFG – German Research Foundation) -- 317046553.

This work is part of the Collaborative Research Center ``On-the-Fly Computing''
at Paderborn University, which is supported by the German Research Foundation
(DFG).
Experiments were performed on
resources provided by the Paderborn Center for Parallel Computing.

\printbibliography
\clearpage

\appendix
\renewcommand*{\thesection}{\Alph{section}}
\section{Notation}\label{asec:notation}

\begin{table}[!h]  %
  \begin{center}
    \caption{Notation used throughout the paper.}
    \begin{adjustbox}{max width=\textwidth}
    \begin{tabular}{ll}
      \hline
      Symbol         & Meaning                                  \\
      \hline
      $[n]$          & $\{1, 2, \dots, n\}$                     \\
      $\indic{A}$ & $1$ if $A$ is a true statement and $0$ otherwise\\
      $\vec{I}_{d}$ & the unit matrix of size $d\times d$ \\
      $\cX$          & set of reference objects                 \\
      $\vec{x}$      & object, choice alternative               \\
      $\cQ$          & choice task space                        \\
     $\IC$ & choice space                             \\
      $\cZ$          & embedding space                          \\
      $Q$            & task                                     \\
      $C$ & choice set, \thatis an element of $\cC$           \\
      $c$ & choice function $\cQ \fromto \cC$ \\
      $\pi$          & ranking                                  \\
      $\Prob$        & probability measure                      \\
      $\prob$        & probability distribution (mass function) \\
      $L$            & loss function on $\cC$                   \\
      $U$            & utility function                         \\
      $\Csingleton{U}{Q}$ & singleton choice from $Q$ according to $U$; \\
      & formally defined as 
      $\argmax_{\vec{x} \in Q} U(\vec{x})$ \\
      $\Csubset{t}{U}{Q}$ & subset choice from $Q$ according to $U$ and $t$; \\
      & formally defined as 
      $\{\vec{x} \in Q \sothat U(\vec{x},Q) \geq t\}$ \\ 
      $\DomainOfUk{k}$ & domain of the sub-utility function $U_{k}$ of FETA; formally defined as \\ 
      & $\{(\vec{x},A) \sothat \vec{x} \in \cX \text{ and } A\subseteq \cX \setminus \{\vec{x}\} \text{ with } |A| = k\}$\\
      $U_{\mathrm{FETA}}^{U_{0},U_{1}}$ & FETA utility function with sub-utility functions $U_{0}$ and $U_{1}$ \\
      $\phi$         & embedding function                       \\
      $U_{\mathrm{FATE}}^{U',\phi}$ & FATE utility function with sub-utility function $U'$ and transformation $\phi$ \\
      $\norm{\cdot}$ & standard euclidean norm in $\IR^{n}$, \thatis $\norm{\vec{x}} = \sqrt{x_{1}^{2}+\dots+x_{n}^{2}}$ \\
      $\theta$       & parameters of a model                    \\
      $\data$        & Dataset                                  \\
      \hline
    \end{tabular}
    \end{adjustbox}
  \end{center}
\end{table}

\section{Evaluation Measures} %
\label{asub:evaluation_metrics}

Besides the target losses introduced in \cref{sub:loss_functions},
we evaluate the trained models using additional evaluation measures.
These should give a more complete picture of the performance of the
different models.
The results including the additional measures can be found
in \cref{asub:results}.

\subsection{Singleton Choice}
\label{asub:dcfmetrics}
To define the evaluation measures in the singleton choice setting, suppose in the following a choice task space $\cQ \subset 2^{\cX}$, a utility function $U$ for $\cQ$ as well as $Q\in \cQ$ and $\vec{x} \in Q$ to be arbitrary but fixed.
\paragraph{\cattopk}
The \textlower{\cattopk} is defined as the fraction of times in which
the set of objects in the top $k$ positions,
according to the predicted scores, contains the ground-truth chosen
object~\citep{chollet2015keras,ben1985}. Formally, writing  $Q=\{\vec{y}_{1},\dots,\vec{y}_{|Q|}\}$
with $U(\vec{y}_{1},Q) \geq \dots \geq U(\vec{y}_{|Q|},Q)$, we have 
\begin{equation}
  \label{eq:topkaccuracy}
  \metric_{\text{top-$k$}}(U,Q,\{\vec{x}\}) \coloneqq \indic[\big]{ \vec{x} \in \{\vec{y}_{1},\dots,\vec{y}_{k}\}} \enspace .
\end{equation}

\paragraph{\catacc}
The \textlower{\catacc} is defined as the fraction of times in which
the object with the largest score is the same as that ground-truth \dc, \thatis 
\begin{equation}
  \label{eq:accuracy}
  \metric_{\text{CA}}(U,Q,\{\vec{x}\}) = \indic[\big]{\vec{x} \in \argmax\nolimits_{\vec{y} \in Q} U(\vec{y},Q)} \enspace .
\end{equation}
The \textlower{\catacc} is the most common measure used for the evaluation
of \acp{SCM} and commonly referred to as
\emph{hit-rate}~\citep{ben1985}.
It is evident that $\metric_{\text{CA}}(U,Q,\{\vec{x}\}) = \metric_{\text{top-$1$}}(U,Q,\{\vec{x}\})$ holds, provided
$\argmax_{\vec{y} \in Q} U(\vec{y},Q)$ is a singleton set.

\paragraph{Normalized Accuracy}
The measures defined above are not a reasonable estimate when observing the
performance of an \ac{SCM} on the \cproblems of different
sizes $\card{Q}$, since the task becomes harder as the
\cproblem size increases.
The hardness of the task should be adjusted with respect to the accuracy that
random guessing can achieve, which is defined as the probability of choosing
the correct \dc from the \cproblem $Q$.
Assuming each object to be chosen with the same probability, the probability for choosing a fixed object is $\frac{1}{|Q|}$.
These considerations motivate the definition of the \emph{normalized accuracy} as follows:
\begin{equation}
  \metric_{\text{CANorm}}(U,Q,\{\vec{x}\}) \coloneqq \frac{\metric_{\text{CA}}(U,Q,\{\vec{x}\}) - \frac{1}{|Q|}}{1 - \frac{1}{|Q|}} \enspace .
  \label{eq:normaccuracy}
\end{equation}
Note that this measure takes values in  $[-\frac{1}{\card{Q}-1}, 1]$.
The minimum value of $-\frac{1}{\card{Q}-1}$ is achieved when the algorithm performs with an accuracy of
\num{0}, \thatis it is worse than random guessing, and the maximum value of $1$ when the learner always predicts correctly.
A value of \num{0} indicates that the learner performs similar to random guessing.
This measure was derived using the “correction for guessing”
formulation~\citep{diamond1973correction}.

\subsection{Subset Choice}
\label{asub:metrics}
For the subset choice setting, we introduce accuracy measures in terms of a choice task $Q$ and two corresponding choices $C,\widehat{C} \subseteq Q$ for $Q$. Here, $C$ may be thought of as the ground-truth choice for $Q$ and $\widehat{C}$ as a prediction made by a learner.  In contrast to the \dc setting, these measures do not depend on a utility function. For the sake of convenience, we suppose $Q$, $C$ and $\widehat{C}$ to be arbitrary but fixed in the following.
To prepare some of the measures, let us formally define the quantities \emph{true positives} ($\widehat{TP}$), \emph{true negatives} ($\widehat{TN}$), \emph{false positives} ($\widehat{FP}$) and \emph{false negatives} ($\widehat{FN}$) via 
\begin{align*}
  \widehat{TP}(Q,C,\widehat{C}) & \coloneqq \frac{1}{\card{Q}} \sum\nolimits_{\vec{x} \in Q} \indic{\vec{x} \in C , \vec{x} \in \widehat{C}}, \\
  \widehat{TN}(Q,C,\widehat{C}) & \coloneqq \frac{1}{\card{Q}} \sum\nolimits_{\vec{x} \in Q} \indic{\vec{x} \not\in C , \vec{x} \not\in \widehat{C}}, \\
  \widehat{FP}(Q,C,\widehat{C}) & \coloneqq \frac{1}{\card{Q}} \sum\nolimits_{\vec{x} \in Q} \indic{\vec{x} \not\in C , \vec{x} \in \widehat{C}},\\
  \widehat{FN}(Q,C,\widehat{C}) & \coloneqq \frac{1}{\card{Q}} \sum\nolimits_{\vec{x} \in Q} \indic{\vec{x} \in C , \vec{x} \not\in \widehat{C}},
\end{align*}
respectively.
These quantities are similar to those used to define the confusion matrix in the case of binary classification \citep{koyejo2015}.

\paragraph{Subset $0/1$ Accuracy}
The \emph{Subset $0/1$ Accuracy} measures the number of times the
ground-truth choice set $C$ and the predicted choice
set $\widehat{C}$ are exactly the same.
This measure is used to measure how often the algorithms predictions match the
complete choice set.
Formally, it is defined as
\[
  \metric_{\text{SUBSET}}(Q,C,\widehat{C}) \coloneqq \indic{C =\widehat{C}}.
\]

\paragraph{Recall}
Recall is defined as the proportion of real positive cases that are correctly
predicted positive~\citep{powers2011evaluation}.
In the field of information retrieval, it is the fraction of the relevant
documents that are successfully retrieved.
For our choice setting this can be defined as the fraction of objects from the
ground-truth choice set $C$ which chosen successfully
or are present in the predicted choice set $\widehat{C}$, \thatis formally as
\[
  \metric_{\text{RE}}(Q,C,\widehat{C}) \coloneqq \frac{\widehat{TP}(Q,C,\widehat{C})}{\widehat{TP}(Q,C,\widehat{C}) + \widehat{FN}(Q,C,\widehat{C})}
\]

\paragraph{Precision}
\emph{Precision} denotes the proportion of predicted positive labels that are
correct~\citep{powers2011evaluation}.
For the choice setting, this can be defined as the fraction of objects from the
predicted choice set $\widehat{C}$ that are actually chosen by
the decision maker or that are present in the ground-truth choice set
$C$.
Formally, it is defined as:
\[
  \metric_{\text{PR}}(Q,C,\widehat{C}) \coloneqq \frac{\widehat{TP}(Q,C,\widehat{C})}{\widehat{TP}(Q,C,\widehat{C}) + \widehat{FP}(Q,C,\widehat{C})}
\]

\paragraph{$F_1$-Measure}
The \emph{$F_1$-measure} is defined as the harmonic
mean of precision and recall:
\[
  \metric_{F_1}(Q,C,\widehat{C}) \coloneqq \frac{2\metric_{\text{PR}}(Q,C,\widehat{C}) \metric_{\text{RE}}(Q,C,\widehat{C})}
  {\metric_{\text{PR}}(Q,C,\widehat{C}) + \metric_{\text{RE}}(Q,C,\widehat{C})}
\]
It can also be expressed in form of the confusion matrix quantities as
follows~\citep{koyejo2015}:
\[
  \metric_{F_1}(Q,C,\widehat{C}) = \frac{2 \widehat{TP}(Q,C,\widehat{C})}
  {2 \widehat{TP}(Q,C,\widehat{C}) + \widehat{FN}(Q,C,\widehat{C})
    +\widehat{FP}(Q,C,\widehat{C})}
\]

\paragraph{Informedness}
The \emph{informedness} is a measure proposed by \citet{powers2003,powers2011evaluation}, which is, in contrast to the $F_1$-measure, unbiased with respect to the
population prevalence of positives.
It specifies the probability that the learner makes an informed prediction if
compared to chance and is formally defined as
\begin{align*}
  \metric_{\text{Inf}}(Q,C,\widehat{C})
  \coloneqq \frac{\widehat{TP}(Q,C,\widehat{C})}
  {\widehat{TP}(Q,C,\widehat{C})+\widehat{FN}(Q,C,\widehat{C})}
  + \frac{\widehat{TN}(Q,C,\widehat{C})}{\widehat{TN}(Q,C,\widehat{C})
    + \widehat{FP}(Q,C,\widehat{C})} - 1
\end{align*}
A very desirable property of this measure is that it is exactly
\num{0} in case the learner is guessing or is constant.

\paragraph{AUC-ROC}
The \emph{AUC-ROC} is a performance measure, which estimates the capacity of a
classification  model to distinguish between two classes
\citep{fawcett2006roc, donna1989analyzing}.
It computes the probability that a classifier will rank a randomly chosen
positive instance higher than a randomly chosen negative
one~\cite{donna1989analyzing}.
It is estimated by computing the area under the ROC-curve, which is
created by plotting the true positive rate $\metric_{\text{TPR}}$ against the false positive
rate $\metric_{\text{FPR}}$, where 
\begin{align*}
  \metric_{\text{TPR}}(Q,C,\widehat{C}) &\coloneqq \frac{\widehat{TP}(Q,C,\widehat{C})}
  {\widehat{TP}(Q,C,\widehat{C}) + \widehat{FN}(Q,C,\widehat{C})}, \\
  \metric_{\text{FPR}}(Q,C,\widehat{C}) &\coloneqq \frac{\widehat{FP}(Q,C,\widehat{C})}
  {\widehat{TN}(Q,C,\widehat{C}) + \widehat{FP}(Q,C,\widehat{C})} \enspace .
\end{align*}
A very desirable property of this measure is that it exactly
\num{0.5} in
case the learner is guessing.

\section{Additional Experimental Details}
\label{sub:experimentaldetails}
In this section, we will now list all experimental details which were
excluded from the main paper for conciseness reasons.
First, we explain the process of nested cross-validation using the hyperparameter optimization in detail.
Then we explain different hyperparameters which were tuned for different models and which parameters were kept fixed. 
Lastly, we explain the design generalization experiment.

\paragraph{Empirical Comparison}
\begin{figure}[t]  %
  \centering
  \includegraphics[width=\linewidth]{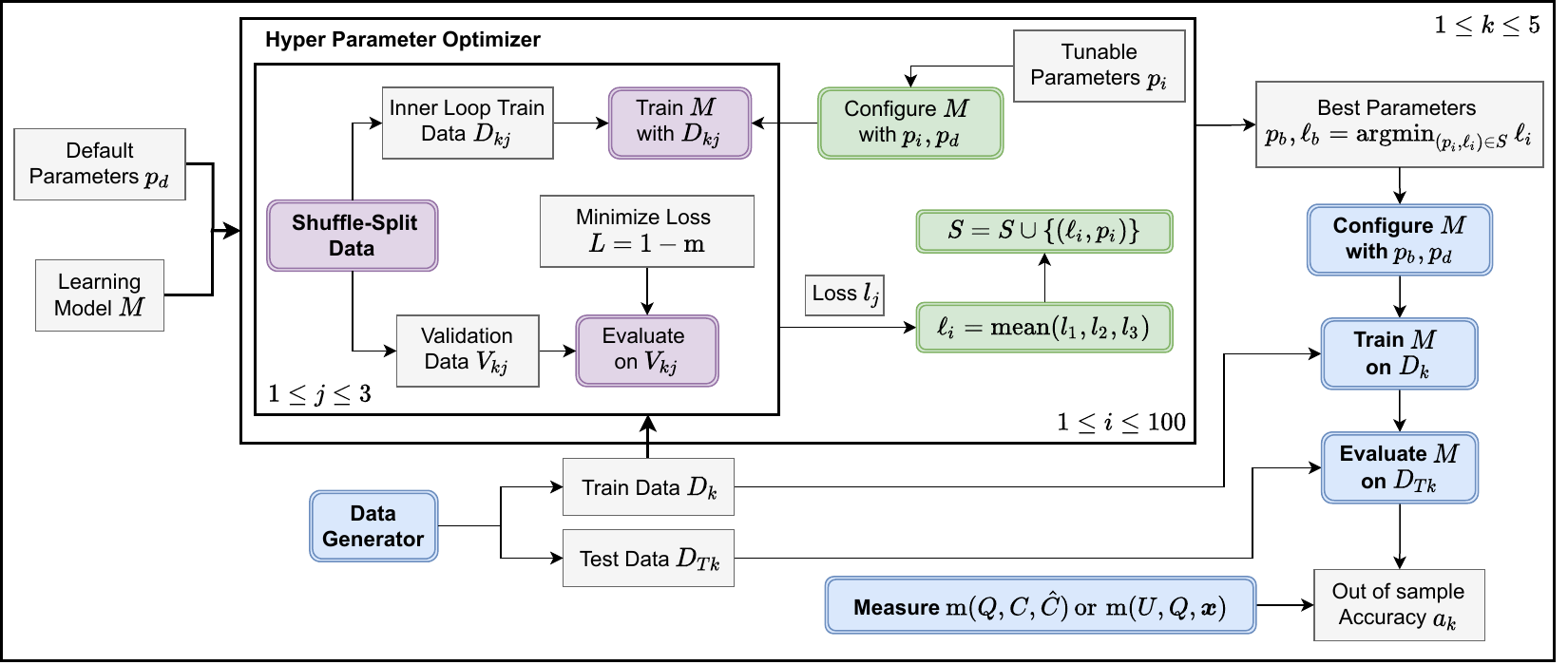}
  \caption{Overview of the complete evaluation pipeline.}
  \label{fig:expsetup}
\end{figure}
In order to compare all learners fairly, we do nested
cross-validation with
synchronized random streams for all the learning models, as shown in
\Cref{fig:expsetup}.
The hyperparameters of all models are tuned using extensive Bayesian
optimization.
We describe the complete procedure in two parts: first
the \emph{hyperparameter optimization} and second the \emph{out-of-sample evaluation}.
First, we configure the given the learner
$M$ with the default parameters
$p_d$
described in the next section.
Then we generate \num{5} sets of training
$\data_k$ and test dataset
$\data_{Tk}$ $\forall k \in [5]$ and the process which
is used to generate a train-test set
for $k$ is described in \cref{tab:datasets:overview}.

\paragraph{Hyperparameter Optimization}
The training set $\data_k$ is used to first identify the best
hyperparameters using
\num{3}-fold stratified cross-validation, and then to train
the final learner for out-of-sample
evaluation.
The hyperparameter optimizer picks hyperparameters from the ranges in
\Cref{tab:ranges} ($p_i$)
for the $i^{th}$ iteration.
In the inner loop $1 \leq j \leq 3$, we split the full training
dataset $\data_k$ into train set
($D_{kj}$ \SI{90}{\percent} of
$\data_k$) and validation dataset
($V_{kj}$ \SI{10}{\percent} of
$\data_k$) using the stratified shuffle split.
For the given hyperparameters $p_i$, we train the model
on the train set ($D_{kj}$) and evaluate
on the validation dataset $V_{kj}$ using the target loss
function.
We use the $1-F_1$-measure for general subset choice and the
1-\textlower{\catacc} for \dc as the target loss to
evaluate the hyperparameter configuration.
We calculate the mean loss $\ell_i = \text{mean}(l_1, l_2, l_3)$ for the given
hyperparameters $p_i$.
The optimization loop is run for \num{100} iterations to
validate \num{100} sets of
hyperparameters, in order to acquire the optimal parameters
$p_b$ for the given learning model.

\paragraph{Out-of-Sample Evaluation}
Finally, after optimization, we configure the learners
$M$ using the best found hyperparameters
$p_b$ and the remaining default parameters
$p_d$.
Then, we train the model $M$ on the complete training
dataset $\data_k$ and evaluate on the test
dataset $\data_{Tk}$ using different evaluation measures
$\metric$ defined in
\Cref{asub:evaluation_metrics}.
To obtain a good estimate of the mean performance and an estimate for the
standard deviation,
we repeat this procedure $5$ times using outer
cross-validation.
For each fold $k \in [K], K=5$, we get the evaluated value
$a_k$ and calculate the mean and
the standard deviation of the performance measure $\metric$.

\begin{table}[t]  %
  \centering
  \caption{Hyperparameter ranges used by the optimizer to select configurations for the
    learners.}
  \label{tab:ranges}
  \sisetup{
    table-figures-decimal = 0,
    table-figures-exponent = 0,
    table-number-alignment = center
  }
  \begin{adjustbox}{width=\linewidth,center}
    \begin{tabular}{
        l%
        S[table-figures-integer=2]
        S[table-figures-integer=2]
        S[table-figures-integer=2]
        S[table-figures-integer=2]
        S[table-figures-integer=2]
        S[table-figures-integer=2]
        S[table-figures-integer=2]
        S[table-figures-integer=2]
        S[table-figures-integer=2]
        S[table-figures-integer=2]
        S[table-figures-integer=2]
        S[table-figures-integer=2]
        S[table-figures-integer=2]
        S[table-figures-integer=2]}
      \toprule
      \multicolumn{1}{c}{ } & \multicolumn{4}{|c}{Architecture Parameters} & \multicolumn{3}{|c}{Other Parameters} & \multicolumn{2}{|c}{\lrscheduler} & \multicolumn{2}{|c}{Linear Model Parameters}                                                                                                                                                                                        \\
      \midrule
      {Learner}             & {Set Units}                                  & {Set Layers}                          & {Joint Units}                     & {Joint Layers}                     & {Regularizer Strength} & {Learning Rate}    & {Batch Size}  & {Epochs Drop $e_{drop}$} & {Drop $d_{r}$} & {tol}             & {C}     \\
      \midrule
      \fetanet              & NA                                           & NA                                           & ${[1,20]}$ & ${[4,1024]}$ & ${[10^{-10},0.1]}$     & ${[10^{-5},0.01]}$ & ${[32,4096]}$ & ${[50,250]}$             & ${[0.01,0.5]}$ & NA                & NA      \\
      \midrule
      \fatenet              & ${[4,1024]}$                                 & ${[1,20]}$                            & ${[4,1024]}$                      & ${[1,20]}$            & ${[10^{-10},0.1]}$     & ${[10^{-5},0.1]}$  & ${[32,4096]}$ & ${[50,250]}$             & ${[0.01,0.5]}$ & NA                & NA    \\
      \midrule
      \fetalinear           & NA                                           & NA                                          & NA         & NA           & ${[10^{-10},0.1]}$     & ${[10^{-5},0.1]}$  & ${[32,2048]}$ & ${[10,150]}$             & ${[0.01,0.5]}$ & NA                & NA    \\
      \midrule
      \sda         & ${r [4,64]}$ & ${r [1,4]}$ & ${w[4,64]}$                      & ${w[1,4]}$            & ${[10^{-10},0.1]}$     & ${[10^{-5},0.1]}$  & ${[8,1024]}$ & ${[50,250]}$             & ${[0.01,0.5]}$ & NA                & NA    \\
      \midrule
      \ranknet              & NA                                           & NA                      & ${[1,20]}$ & ${[4,1024]}$ & ${[10^{-10},0.1]}$     & ${[10^{-5},0.01]}$ & ${[64,8192]}$ & ${[50,250]}$             & ${[0.01,0.5]}$ & NA                & NA      \\
      \midrule
      \ranksvm              & NA                                           & NA                                          & NA         & NA           & NA                     & NA                 & NA            & NA                       & NA             & ${[10^{-4},0.5]}$ & ${[1,12]}$ \\

      \bottomrule
    \end{tabular}
  \end{adjustbox}
\end{table}
\paragraph{Hyperparameters \& Inference}
We will now describe the specific hyperparameters we optimize and which ranges
of values we
consider (see \cref{tab:ranges} for an overview).
For probabilistic models, we also describe how the inference is done.
For all neural network models, we make use of the following techniques:
\begin{itemize}
  \item We use either \ac{ReLU} non-linearities in conjunction with
        \ac{BN} \citep{Ioffe2015} or
        \ac{SELU} non-linearities \citep{selu}
        for each hidden layer.
  \item Regularization: $L_2$ penalties are applied and the
        corresponding regularization strength is tuned.
  \item Optimizer: \ac{SGD} with Nesterov
        momentum~\citep{nesterov1983}.
  \item A step-decay function is used for the learning rate annealing schedule.
        The decay factor is tuned~\citep{duchi2011adaptive}.
\end{itemize}
The step-decay function drops the learning rate by a factor after a certain
number
epochs~\citep{duchi2011adaptive}.
Formally, it is defined as:
\[\mathit{lr} = \mathit{lr}_0 \cdot
  d_r^{\floor{\frac{e}{e_{\text{drop}}}}}  \, ,\]
where $\mathit{lr}_0$ is the initial learning rate,
$0 < d_r < 1$ is the rate with which the learning rate should be
reduced, $e$ is the current epoch and
$e_{\text{drop}}$ is the number of epochs after which the learning
rate is decreased.
We set the maximum number of epochs the neural networks are trained for to \num{1000}.

The hyperparameters of each algorithm were tuned using the package
scikit-optimize \citep{skopt}.
Apart from the number of hidden layers and units, we also tune the learning
rate of the
stochastic gradient descent optimizer, regularization strength and batch size
(fraction of
training examples used for estimating the gradient in one iteration).
We also tune the drop-rate $d_r$ and epoch-drop
$e_{\text{drop}}$ for the step-decay function
used by the Stochastic gradient descent optimizer by the neural networks.
For \pairwisesvm, we tune the value of the penalty parameter
$C$ of the error term, and another
is \emph{tol} (\emph{tol} in scikit-learn) which is the tolerance for the stopping
criteria of
the optimization algorithm~\citep{scikit-learn}.
All of the different \ac{GEV} models are implemented in
PyMC3 a library for facilitating Markov Chain Monte Carlo estimation of the
posterior distribution~\citep{salvatier2016pymc3}.
An overview of all the hyperparameters and their admissible ranges is shown in
\Cref{tab:ranges}.

\paragraph{Threshold Tuning} %

In order to set the threshold for the subset choice models \eqref{eq:detsubset},
we tune the threshold for all models on a small validation set.
Obviously, an optimal value for $t$ will depend on the underlying target loss
function.
Our main target loss is the (micro-averaged) $F_1$-measure \eqref{eq:f1x},
which balances precision and recall of the predictions \parencite{lewis1995, ye2012, waegeman2014}.
\citet{koyejo2015} show that tuning a threshold on a validation set,
yields a consistent classifier, if the estimated marginal instance probabilities
(in our case the choice probabilities) converge in probability to the
population-level probabilities.
One important difference to the multi-label classification setting is the
absence of a fixed set of labels.
Instead, we have a dynamically changing set of objects.
Thus, it only makes sense to consider micro-averaged performance metrics.

\section{Design of the Generalization Experiment}
\label{sub:experimentaldetails:genexp}
\begin{figure}[t]  %
  \centering
  \includegraphics[width=\linewidth]{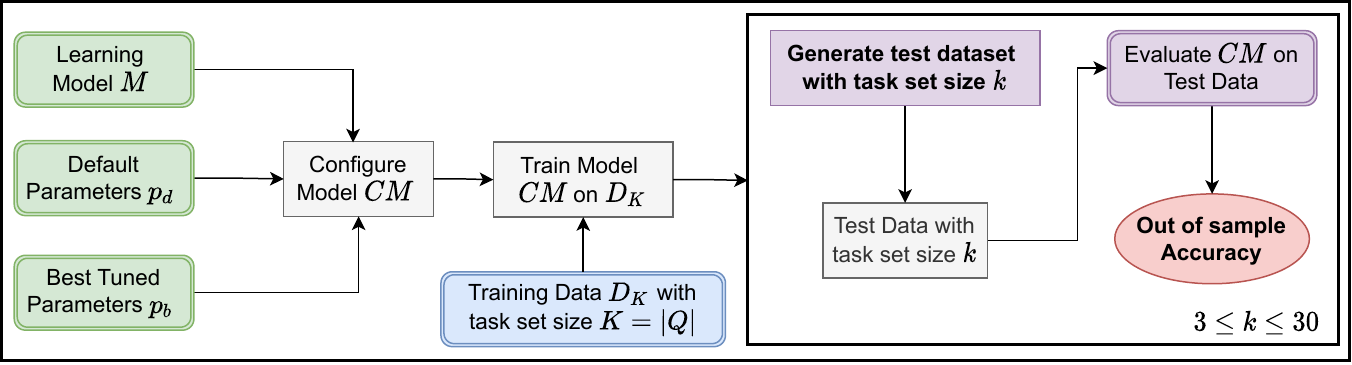}
  \caption{Design of the generalization experiments.}
  \label{fig:genexpsetup}
\end{figure}
The second experimental setup is designed to gauge the generalization
capability of the learning
models by measuring the accuracy obtained by a trained model on unseen \qset
sizes.
To this end, we vary the \qset sizes from \num{3} to
\num{30} as shown in \Cref{fig:genexpsetup}.

First, we configure the learning model with the best hyperparameters
$p_b$ obtained from the
empirical comparison experiment for the given dataset and the remaining default
parameters
$p_d$.
Then we generate the training dataset containing \qsets of size
$K = \card{Q}$ and train the configured
model on the training dataset $\data_K$.
\begin{table}[tb]  %
  \centering
  \caption{Dataset configurations for generalization experiments.
    Bracket notation is used to denote the range of values.
  }
  \label{tab:datasetconfiggen}
  \sisetup{
    table-figures-decimal = 0,
    table-figures-exponent = 0,
    table-number-alignment = center
  }
    \begin{tabular}{
        ll%
        S[table-figures-integer=2]
        S[table-figures-integer=2]
        S[table-figures-integer=2]
        S[table-figures-integer=2]
        S[table-figures-integer=2]}
      \toprule
      Problem & Dataset     & {\#\,Features} & {\#\,Train} & {\#\,Test}   & {\Qset sizes $S$} & {\Qset Size $\card{Q}$} \\
      \midrule
      \multirow{2}{*}{Singleton Choice}
              & Medoid      & \num{5}        & \num{10000} & \num{100000} & ${[3,30]}$        & \num{10}                \\
              & Hypervolume & \num{2}        & \num{10000} & \num{100000} & ${[3,30]}$        & \num{10}                \\
      \bottomrule
    \end{tabular}
\end{table}
Finally, we evaluate the trained model $CM$ on different
test datasets $\data_k$ containing the
\qsets of sizes in $S$ ($\card{Q} = k \in S$) as
described in \Cref{tab:datasetconfiggen}.

\section{Synthetic Datasets}
\label{sub:dgp}
In this section, we will formally describe the process of generating the
datasets for the experimental evaluation.
In the case of synthetic datasets, this entails the complete process by which
the objects and queries are generated.
\begin{figure}[t]  %
  \centering
  \begin{subfigure}[b]{0.30\linewidth}
    \includegraphics[width=\linewidth]{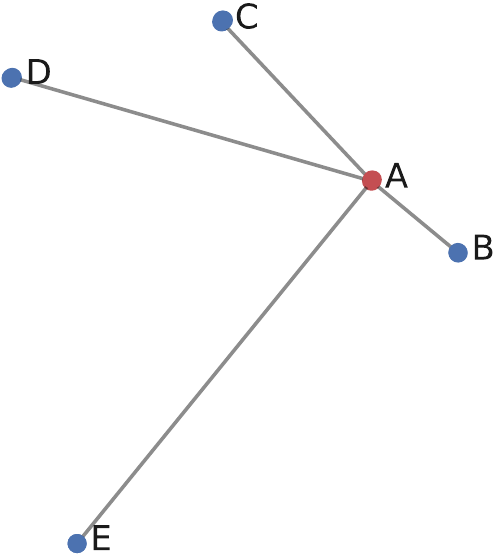}
    \subcaption{Random points in $[0,1]^2$ are ranked according to their
      distance to the medoid (here shown in orange}
    \label{fig:medoid}
  \end{subfigure}\hfill
  \begin{subfigure}[b]{0.30\linewidth}
    \includegraphics[width=\linewidth]{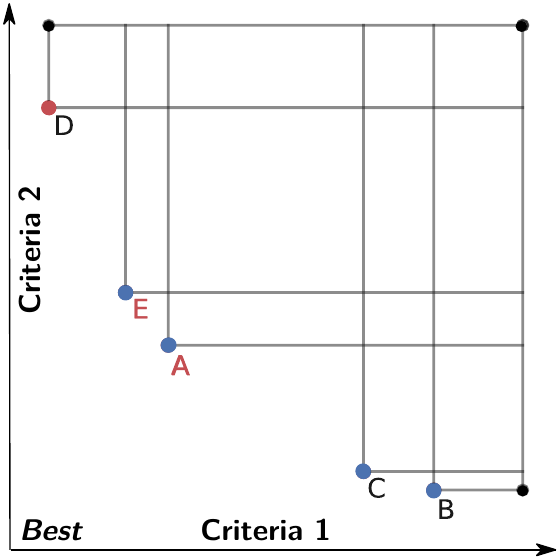}
    \subcaption{Random points in $[0,1]^2$, on Pareto-font ranked according to thier contributions}
    \label{fig:hypervolume}
  \end{subfigure}\hfill
  \begin{subfigure}[b]{0.30\linewidth}
    \includegraphics[width=\linewidth]{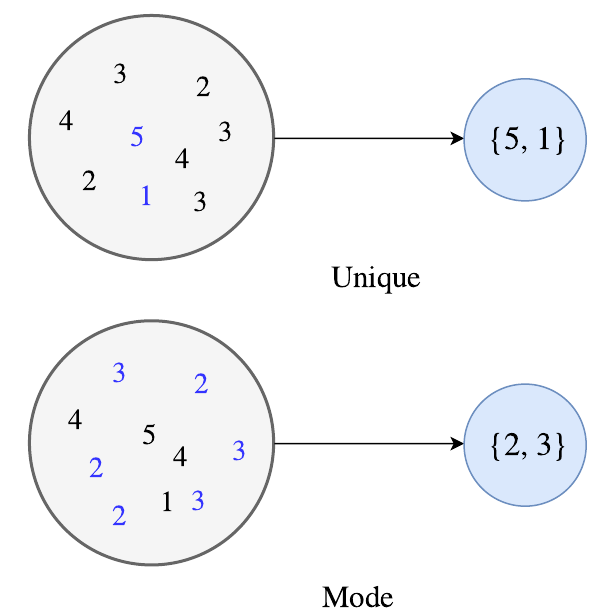}
    \subcaption{Example for unique and mode function on MNIST digits}
    \label{fig:mnist}
  \end{subfigure}
  \caption{Examples for synthetic datasets}
  \label{fig:syntheticdatasets}
\end{figure}
\subsection{The Medoid Problem}
\label{sub:dgp:medoid}
Recall that we have defined the \emph{medoid} of a set $Q \subset \R^{d}$ as $c_{\operatorname{medoid}}(Q)= \argmin\nolimits_{\vec{x} \in Q} \frac{1}{|Q|} \sum\nolimits_{\vec{y} \in Q} \norm{\vec{x}-\vec{y}}$, where $\norm{\cdot}$ is the standard euclidean norm in $\R^{d}$. Thus, the medoid of $Q$ may be thought of as the most centrally located object in $Q$, cf. the illustration of a choice set $Q$ of size $5$ and its medoid in \Cref{fig:medoid}. As it depends on its distance to any other point from $Q$, the medoid of $Q$ is sensitive to changes of any points in $Q$.

For our empirical study, we created a dataset $\data = \set{(\qi{1}, \ci{1}), \dotsc, (\qi{N}, \ci{N})}$ by drawing each $\qi{i}$ independently and uniformly at random
from the set
\begin{equation*}
    \left\{Q \subset [0,1]^{d} \sothat |Q|=n \text{ and } |c_{\operatorname{medoid}}(Q)|=1 \right\}
\end{equation*}
and then choose $C_{i} \coloneqq c_{\operatorname{medoid}}(Q_{i})$. Here, the sampling step can be performed via the acceptance-rejection method: One may repeatedly sample $\vec{x}_{1},\dots,\vec{x}_{n}$ uniformly at random from $[0,1]^{d}$ until $Q=\{\vec{x}_{1},\dots,\vec{x}_{n}\}$ has size $n$ and a unique medoid. Regarding that this condition is already fulfilled with probability $1$ after sampling $\vec{x}_{1},\dots,\vec{x}_{n}$ only once, this method is efficient.

\subsection{The Pareto Problem}
\label{sub:dgp:pareto}
Above, we introduced the \emph{Pareto set} $c_{\mathrm{Pareto}}(Q)$ of a set $Q\subset \R^{d}$ as the set of all elements $\vec{x} \in Q$ which are not dominated by any $\vec{y} \in Q \setminus \{\vec{x}\}$, wherein $\vec{x}$ was said to dominate $\vec{y}$ if $\forall i\in [d]\colon x_{i} \leq y_{i}$ and $\exists i\in [d]\colon x_i < y_i$.
\cref{fig:hypervolume} shows the Pareto set of a set $Q\subset \IR^{2}$.

With the help of Pareto sets we create a synthetic dataset $\data = \{(Q_{j},C_{j})\}_{j=1}^{N}$ for the subset choice task, where each sample $(Q,C) \in \data$ is generated independently of the others in the following way:
\begin{enumerate}
    \item Sample $\vec{\mu}_{1}, \dots, \vec{\mu}_{n}$ i.i.d. uniformly at random from $\{\vec{x} \in \R^{d} \sothat \norm{x} \leq 1\}$
    \item Draw i.i.d. samples  $\vec{\xi}_{1},\dots,\vec{\xi}_{n}$ from $N(\vec{0},\vec{I}_{d})$, the standard Gaussian distribution on $\R^{d}$, and define $\vec{x}_{i} \coloneqq \vec{\mu}_{i} + \vec{\xi}_{i}$ for each $i\in [n]$.
    \item Choose $Q\coloneqq \{\vec{x}_{1},\dots,\vec{x}_{n}\}$ and $C\coloneqq c_{\mathrm{Pareto}}(Q)$.
\end{enumerate}

\paragraph{Hypervolume}
In \cref{subsec_Hypervol} we have introduced for $Q\subset \R^{d}$ the choice set $c_{\mathrm{HypVol}}(Q)$ as the set  of all $\vec{x} \in Q$, which contribute the least among all elements in $Q$ to the \emph{hypervolume} of $Q$, cf. \cref{subsec_Hypervol} for the precise definitions and also for the connection of the hypervolume of $Q$ to the Pareto front of $Q$. As this contribution of each point depends on the position of other
points in $Q$, $c_{\mathrm{HypVol}}$ is context-dependent. This is illustrated in  \cref{fig:hypervolume}, where all five elements of $Q = \{A,B,C,D,E\}$ lie on the Pareto front of $Q$. There, the
contribution of point $A$ is largest in $Q$, but if we remove the point $D$ from the choice set, it increases the contribution of the point $E$ for the set.
So, the \dc changes from $A$ to $E$, after removing $D$ from $Q$.

Based on $c_{\mathrm{HypVol}}$ we construct a singleton choice dataset $\data = \{(Q_{i},C_{i})\}_{i=1}^{N}$ by sampling each $Q_{i}$ uniformly at random from the set of all  $Q\subseteq \R^{d}$, which fulfill
 \begin{equation*}
      \card{Q}=n, \  \forall \vec{x} \in Q\colon (\norm{\vec{x}} = 1 \text{ and } \forall i\in [d]\colon x_{i} \leq 0) \text{ and } \card{c_{\mathrm{HypVol}}(Q)} = 1,
 \end{equation*}
 and then defining $C_{i} \coloneqq c_{\mathrm{HypVol}}(Q_{i})$ afterwards.
 Similarly, as in the construction of the Medoid data set, sampling can be done via the acception-rejection method.

\subsection{MNIST Number Problems}
\label{sub:dgp:mnist}
\begin{figure*}[tb]
  \centering
  \includegraphics[width=0.9\linewidth]{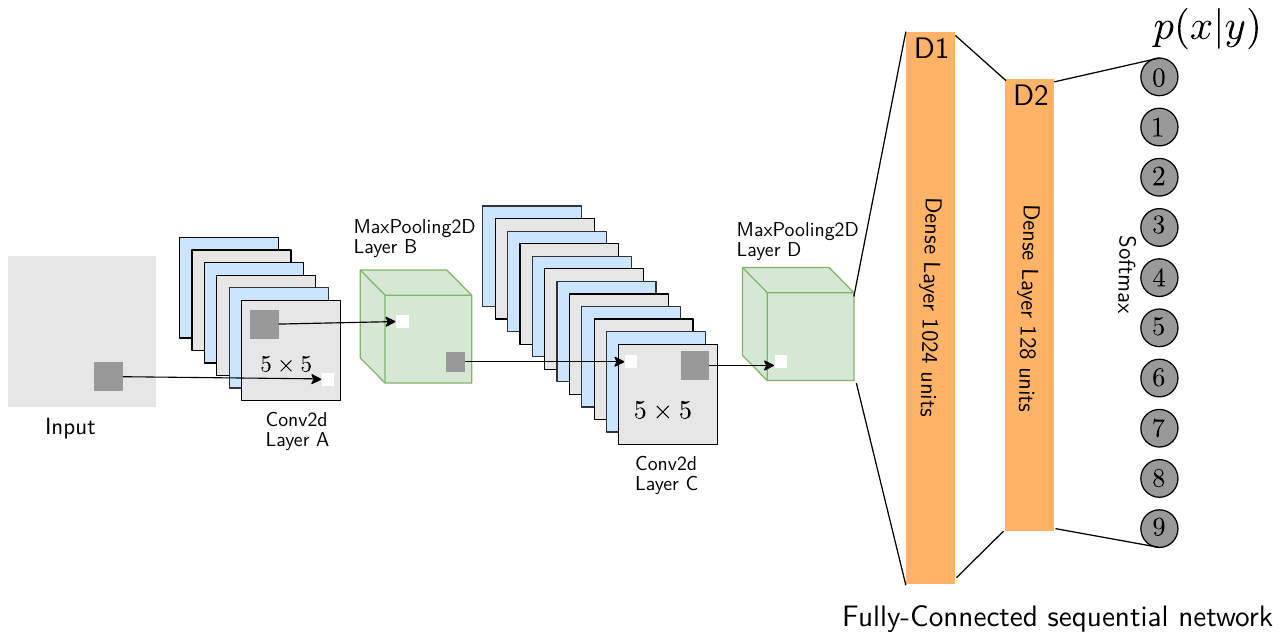}
  \caption{CNN-For converting MNIST images to high-level features}
  \label{fig:cnn}
\end{figure*}

In this section, we will describe the process of generating different
semisynthetic datasets using the \ac{MNIST}
dataset~\citep{mnisthandwrittendigit}.
\paragraph{Feature Extraction}
\label{sec:dgp:mnist:fe}
Since the dataset consists of \num{2}-D image maps, we
first train an off-the-shelf \ac{CNN} to solve the digit
multi-class classification task to level the playing field and abstract away from the
computer vision context.
This architecture of the \ac{CNN} consists of
\num{2}-D Convolutional, \num{2}-D
Max-Pooling, and fully-connected dense layers and applied batch normalization
to increase the stability of the network, by subtracting the batch mean and
dividing by the batch standard deviation as shown in
\cref{fig:cnn}~\citep{goodfellow2016deep, Ioffe2015}.
The \num{2}-D convolutional layer is of kernel-size
$5 \times 5$ using \acf{ReLU} non-linear
activation function and
\emph{l}-\num{2} regularization and
\num{2}-D max-pooling layer, with filter of size
$2 \times 2$ applied with a stride of
\num{2}, which down-samples the input by
\num{2} along the width and height, discarding
$\SI{50}{\percent}$ of the activations by applying max operation over
\num{4} numbers in $2 \times 2$
region~\citep{goodfellow2016deep}.
The output of these layers is provided as input to a fully-connected sequential
network with \num{10} outputs, where each output predicts
the probability of the input image belonging to a particular
class using the softmax~\citep{goodfellow2016deep}.
We train this network on \num{10000} instances, then we
transform the remaining \num{60000} digits to a high-level
feature representation by passing them through the trained
\ac{CNN} and recording the \num{128}
outputs of the last hidden layer~(D2).

The transformed \ac{MNIST} dataset
$\data_{M} = \set{(\vec{x}_1, l_1), \ldots ,(\vec{x}_N, l_N)}$, is represented as a set of tuples
$(\vec{x}_i, l_i)$, where $\vec{x}_i$ is the feature
vector and $l_i$ represents the corresponding label, such
that $ \card{\data_{M}} = N = 60000$, $\vec{x}_i \in \IR^{128}$, 
$l_i \in \cL = \set{0,1,2,3,4,5,6,7,8,9}$ and $\data_{M}(\vec{x}_i) = l_i$ holds for all $i
  \in [N]$.
For constructing the choice datasets, we sample instances
$(\vec{x}_i, l_i) \in \data_{M}$ from the transformed dataset uniformly at random,
to construct a \qset $Q = \cset$.
Based on $Q$ and $\vec{l} = (\data_{M}(\vec{x}_1), \ldots , \data_{M}(\vec{x}_n))$, we then select as choice set $C=g(Q,\vec{l})$, where is an appropriately predefined function $g$. We consider two variants for $g$, namely $g_{\mathrm{unique}}$ and $g_{\mathrm{mode}}$.

The function $g_{\mathrm{unique}}$ outputs the instances corresponding to the numbers which occur only once in the label vector. For example
\\$g_{\mathrm{unique}}(Q, (4,3,2,3,3,1,8,8,7,7)) = \set{\vec{x}_1, \vec{x}_3, \vec{x}_6}$, corresponding to the numbers $4$, $2$ and $1$.
For \dc choice, we sample only the \qsets, whose corresponding label vector $\vec{l}$ contains a single unique number, to make it identifiable, \thatis for example $g_{\mathrm{unique}}(Q, (4,3,3,2,2,1,1,1,5,5,5)) = \set{\vec{x}_1}$.
The section function is $g_{\mathrm{mode}}$, which outputs the instances corresponding to the number which occur most frequently in the label vector. For example
$g_{\mathrm{mode}}(Q, (4,3,2,3,3,8,8,7,7,7)) = \set{\vec{x}_8, \vec{x}_9, \vec{x}_{10}}$, corresponding to the mode $7$
For \dc choice, we choose the instances corresponding to the mode, which are at the least angle from a predefined weight vector $\vec{w}$.

Both functions used to generate choices depend on all other objects in the given \qset $Q$, thus making
the datasets highly context-dependent.
\paragraph{Unique}
In this subsection, we explain the data generation process for the
\emph{Unique} choice dataset using the $g_{\mathrm{unique}}$ function defined above.
For generating the dataset, we select a set of instances from
$\data_{M}$ uniformly at random to construct the \qset
$Q$ and the label vector $\vec{l}$.
Then we choose the objects from $Q$ which corresponds to
the unique digit in the label vector $\vec{l}$ (an example is
shown in \cref{fig:mnist}).
Let us assume we want to generate a dataset 
$\data = \set{(\qi{i}, \ci{i})}_{i=1}^N$ 
with $N$ instances.
\begin{enumerate}
    \item Sample $n$ data points $(\vec{x}_{i,1},l_{1}), \dotsc ,(\vec{x}_{i,n},l_{n})$ from $\data_{M}$, let $\vec{l}_{i} \coloneqq (l_{1},\dots,l_{n})$ and $Q_{i} \coloneqq \{\vec{x}_{i,1},\dots,\vec{x}_{i,n}\}$
    \item For each $l\in \mathcal{L}$ let $k_{l}$ be the number of times the label $l$ appears in the label vector $\vec{l}_{i}$ for $Q_{i}$, define $\vec{k} \coloneqq \{k_{0},\dots,k_{9}\}$ and write for convenience $\vec{k}(l) \coloneqq k_{l}$ in the following. For example for $\vec{l} = (1,2,4,4,4,5,5)$ we have
        $\vec{k} = (0,1,1,0,3,2,0,0,0,0)$.
  \item We create $\ci{i}$ by selecting the objects whose values
        occur only once in the label vector $\vec{l}$:
        \[\ci{i} \coloneqq  \set{\vec{x}_{i,j} \in Q_i \sothat \vec{k}(l_j)=1}\]
  \item In order to create the corresponding \dc or top-$1$
        version of this dataset, we discard $Q_{i}$ in case $\card{\ci{i}}>1$ and  repeat steps 1--4.  If $|C_{i}| =  1$ instead, we keep the sample $(Q_{i},C_{i})$.
\end{enumerate}

\paragraph{Mode}
In this subsection, we explain the data generation process for the 
\emph{Mode} choice dataset using the $g_{\mathrm{mode}}$ function defined above.
For generating the dataset, we select a set of instances from
$\data_{M}$ uniformly at random to construct the \qset
$Q$ and the label vector $\vec{l}$
(an example is shown in \cref{fig:mnist}).
Then we choose the objects from $Q$ which corresponds
to the mode value of the label vector $\vec{l}$ to construct
the ground-truth set of chosen objects.
For creating the corresponding \dc or top-$1$ dataset, we choose the object
corresponding to the mode value of the label vector, which is at the least
angle to the predefined weight vector $\vec{w}$.
Let us assume we want to generate a dataset $\data = \set{(\qi{i}, \ci{i})}_{i=1}^N$ with $N$ instances.
First, we sample the weight vector $\vec{w} \in \IR^{128} \overset{\text{iid}}{\sim} N(\vec{0},\vec{I}_{128})$.
\begin{enumerate}
    \item Sample $n$ data points $(\vec{x}_{i,1},l_{1}), \dots ,(\vec{x}_{i,n},l_{n})$ uniformly at random from $\data_{M}$, abbreviate $\vec{l}_{i} \coloneqq \{l_{1},\dots,l_{n})$ and let $Q_{i} \coloneqq \{\vec{x}_{i,1},\dots, \vec{x}_{i,n}\}$.
    \item As for the Unique dataset, write $k_{l}$ for the number of times the label appears $l$ in the label vector $\vec{l}_{i}$ for $Q_{i}$, define $\vec{k} \coloneqq \{k_{0},\dots,k_{9}\}$ and write again $\vec{k}(l) \coloneqq k_{l}$.
    \item For the case of subset choice define
    \[\ci{i} = \set{\vec{x}_{i,j} \in Q_i \sothat
            \vec{k}(l_i)=\max\nolimits_{l\in \mathcal{L}}(\vec{k}(l)) },\]
    and in case of singleton choice, select $\ci{i}$ to be that set, which contains only the object with the least angle to vector $\vec{w}$, i.\,e., 
    \[\ci{i} \coloneqq \set{\argmax\nolimits_{\vec{x} \in \ci{i}} \cos^{-1}
            \frac{\vec{x} \cdot \vec{w}}{\norm{\vec{x}} \norm{\vec{w}}}}\]
    
\end{enumerate}

\subsection{Tag Genome Dataset}
\label{sub:dgp:taggenome}
The GroupLens Research group released many datasets collected from the
MovieLens website\footnote{\url{https://movielens.org/}} for research in the
field of recommender systems~\cite{harper2016movielens}.
As of August 2017, the full dataset collected from this website consists of
\num{26000000} ratings and \num{750000} tags
applied to \num{45000} movies by \num{270000}
users~\cite{harper2016movielens}.
One of the datasets is the Tag Genome dataset\footnote{This dataset is
  available on \url{https://grouplens.org/datasets/movielens/}},
which provides real-valued features to characterize the movies~\cite{vig2011navigating}.

Tags are meta-data in the form of keywords, which help to describe an object
(such as movie, music, books).
In recent years tagging has gained popularity due to the growth of social
networking websites and web search engines~\cite{smith2007}.
On the MovieLens website, users create tags to describe a movie.
Other users can then use them to filter movies more effectively.
Users can also gain more information about a movie with the help of tags
applied by other users.

\begin{figure}[htb]
  \centering
  \includegraphics[width=0.35\linewidth]{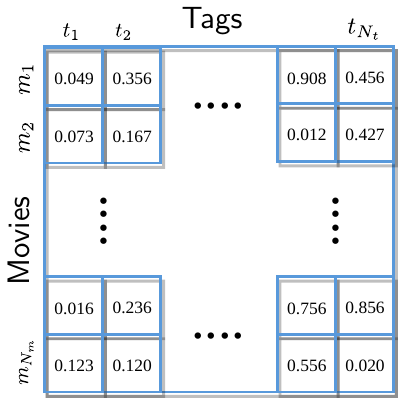}
  \caption{Structure of the Tag Genome dataset.}
  \label{fig:taggenome}
\end{figure}

The Tag Genome dataset was generated by applying machine learning algorithms
on the information provided by users for a movie in the form of tags,
reviews, and ratings~\cite{vig2012tag}.
It consists of movies and a set of tags applied to each of them, and a score
between $0$ and $1$ quantifying
the \emph{relevance} of each tag to the
particular movie (as shown in \Cref{fig:taggenome}).
Currently, this dataset consists of around 12 million relevance scores
across \num{1128} tags applied on
\num{10993} movies.

\paragraph{Framework}
According to \citet{vig2011navigating} the Tag Genome dataset consists of:
\begin{enumerate}
  \item $\mset$: The set of movies $\set{m_1, \ldots ,m_{\nmovies}}$,
        where $\card{\mset} = \nmovies = 10993$.
  \item $\tagset$: The set of tags $\tagset =\set{t_1, \ldots ,t_{\ntags}}$, where
        $\card{\tagset} = \ntags = 1128$.
  \item $\relmat: \mset \times \tagset \fromto [0,1]$: Relation such that $\relmat(m_i,t_j)$
        denotes the degree to which extent the tag $t_j \in \tagset$ applies to the movie
        $m_i \in \mset$ on a scale of $0$ to
        $1$; here $0$ indicates no relevance and $1$ indicates strong relevance to the movie (as shown in \Cref{fig:taggenome}).
  \item $\mmap: \mset \fromto \ui^{\ntags}$: Relation mapping each movie to its feature vector
        in tag-space (vector of tag relevance values across all tags), such that
        $\mmap(m_i) = \vec{x}_i \coloneqq (\relmat(m_{i},t_{1}),\dots,\relmat(m_{i},t_{N_{t}}))$.
        
  \item $\tp: \tagset \fromto \IN $: Function representing the popularity of
        a tag, measured as the number of users who applied the tag
        $t_j \in \tagset$.
  \item $\df: \tagset \fromto \IN $: Function representing the movie frequency of tag
        $t_j \in \tagset$, \thatis $\df(t_j) \coloneqq \sum_{m_i \in \mset}
          \indic{\relmat(m_i,t_j) > 0.5}$ denotes the number of movies for which the
        relevance of tag $t_j$ is greater than
        \num{0.5}.
  \item $P$: The set of top $20$ most
        popular-tags $P \subset \tagset$ based
        on the popularity $\tp$.
\end{enumerate}

The \wcs is a similarity measure defined
in~\citep{vig2011navigating} to measure the similarity between two movies.
The weight vector $\vec{w}$ is defined in such a way that
more weight is
assigned to both the popular tags because this implies that more users care
about these tags and also to more specific tags because they can uniquely
identify the similarity.
For example, if two movies have the \emph{harry potter} tag in common,
they
are more likely to be similar than the ones that have the tag
\emph{fantasy}
in common~\citep{vig2011navigating}.
A $\log$-transform is applied to both values to bring them
closer to the
normal distribution.
The weighted cosine similarity between two movies his defined as:
\begin{equation}
  \cosine(\vec{x}_i, \vec{x}_j, \vec{w}) = \frac{\sum_{k=1}^{\ntags} w_k  x_{ik} 
    x_{jk}}{\sqrt{(\sum_{k=1}^{\ntags} w_k  x_{ik}^2)} \cdot \sqrt{(\sum_{k=1}^{\ntags} w_k  x_{jk}^2)}} \enspace ,
  \label{eq:cosinesim}
\end{equation}
where $\vec{x}_i = \mmap(m_i), \enspace
  \vec{x}_j = \mmap(m_j)$ and $w_k =
  \frac{\log(\tp(t_k))}{\log(\df(t_k))}$ for any $t_k \in \tagset$.

To construct the \dc semisynthetic dataset, we sample uniformly at random 
$n$ movie items from $\mset$ to create a \qset
$Q$, and we choose the
medoid $\vec{r}$ of $Q$ as the reference movie.

We define two tasks based on the reference movie $\vec{r}$ of
the sampled
\qset $Q$.
The first task is to choose the \emph{most similar movie} to the reference
movie in \qset $Q$.
The second task is to choose the \emph{most dissimilar movie} with respect
to the reference movie $\vec{r}$ for a given \qset
$Q$.
This problem is similar to finding the outliers for a given set of objects
which can be used to solve the problem of anomaly
detection~\parencite{aggarwal2001outlier, chandola2009anomaly}.
Both tasks used to generate semisynthetic datasets depend on the similarity
between all objects in the given \qset $Q$, thus making
the datasets highly context-dependent.

\paragraph{Data Generation Process}
We explain the data generation process for the \emph{\msm} and
\emph{\mdm} datasets.
Let us assume we want to generate a \dc dataset
$\data = \set{(\qi{i}, \ci{i})}_{i=1}^N$ with
$N$ instances.
Each task set $Q_{i}$ and its corresponding \dcs $\ci{i}$ is constructed in the following way:
\begin{enumerate}
  \item 
  Sample i.i.d. and uniformly at random $m_{1},\dots,m_{n}$ from $M$, let $\vec{x}_{i,n} \coloneqq \mmap(m_{j})$ for each $j\in [n]$ and $Q_{i} \coloneqq \{\vec{x}_{i,1},\dots, \vec{x}_{i,n}\}$.
  \item Compute the reference object (movie) for $\qi{i}$ (medoid):
        \begin{equation*}
          \vec{r} \coloneqq \argmax\nolimits_{\vec{x} \in \qi{i}}
          \frac{1}{n} \sum\nolimits_{j=1}^n
          \cosine(\vec{x}, \vec{x}_{i,j}, \vec{w})
        \end{equation*}
  \item Now we define the corresponding \dcs $\ci{1}, \dotsc, \ci{N}$ for
        \emph{\msm} and \emph{\mdm} dataset.
        \begin{enumerate}
          \item 
                The \dc set $C_{i}$ for $Q_{i}$ for \emph{\mdm} is the set consisting of only that element of $Q_{i}$, which is most dissimilar to $\vec{r}$, \thatis formally
                \[\ci{i} \coloneqq  \set{\argmin\nolimits_{\vec{x}_{i,j} \in Q \setminus \{ \vec{r} \}} \,  \cosine(\vec{r}, \vec{x}_{i,j}, \vec{w})}\]
          \item For the \emph{\msm} dataset, we select for the task $Q_{i}$ the \dc set
           \[\ci{i} \coloneqq \set{\argmax\nolimits_{\vec{x}_{i,j} \in Q \setminus \{ \vec{r}\} } \,  \cosine(\vec{r}, \vec{x}_{i,j}, \vec{w}) },\]
           which consists of the one element from $Q_{i}$, that is most similar to $\vec{r}$.
        \end{enumerate}
\end{enumerate}

\section{Real-World Datasets}
\label{sec:realdatasets}

Some widely used benchmark-datasets available for solving this task are
\ac{LETOR} and SUSHI~\parencite{letor2013, kami_as10}.
In the following sections, we briefly describe these datasets and the process
we use to generate \emph{singleton} and \emph{subset choice} datasets.

\subsection{LETOR Datasets}
\label{asub:letor}
\ac{LETOR}\footnote{Version 4.0} is a package of
benchmark datasets released by Microsoft Research Asia, which are used to
compare and evaluate different learning algorithms in the field of preference
learning \citep{letor2013}.
We use the datasets \mq{2007} and \mq{2008} released for learning the task of
partial ranking to create the \emph{subset choice} dataset.
There are other datasets \mql{2007} and
\mql{2008} released for learning the task of complete
ranking\footnote{These datasets are available on \url{https://www.microsoft.com/en-us/research/project/letor-learning-rank-informatio
    n-retrieval/}} to create the
\emph{\dc} dataset.

\begin{figure}[htb]
  \centering
  \begin{subfigure}[b]{\textwidth}
    \includegraphics[width=\linewidth]{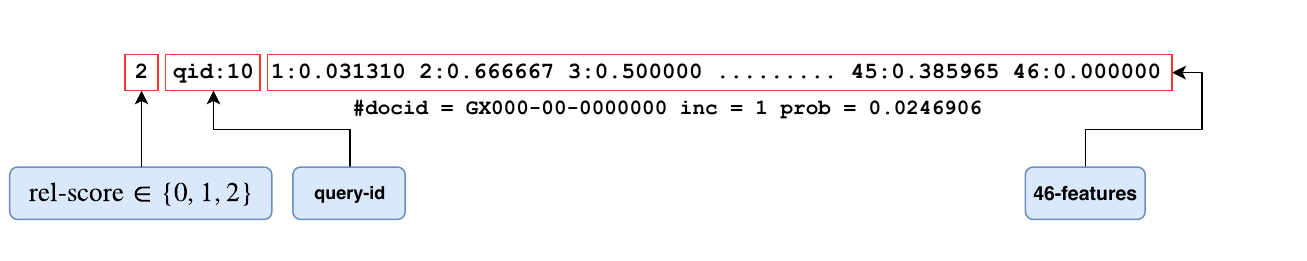}
    \caption{\mq{2007}/\mq{2008} format}
    \label{fig:letorsup}
  \end{subfigure}

  \begin{subfigure}[b]{\textwidth}
    \includegraphics[width=\linewidth]{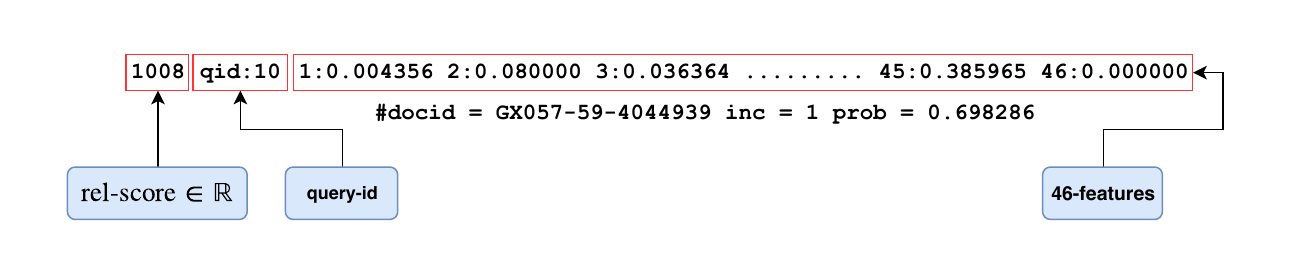}
    \caption{\mql{2007}/\mql{2008} format}
    \label{fig:letorlist}
  \end{subfigure}
  \caption{\ac{LETOR} datasets formats~\cite{letor2013}}
  \label{fig:letor}
\end{figure}

\paragraph{LETOR Supervised Datasets}
The datasets (\mq{2007} and \mq{2008})
consist of the queries and retrieved documents, with individual preferences in
the form of a relevance for each document with respect to the corresponding
query \citep{letor2013}.
The format of both datasets (\mq{2007} and
\mq{2008}) is the same, and there are about
\num{1500} queries in \mq{2007} and about
\num{500} in \mq{2008} with labelled
documents.
These datasets consist of \num{46} features extracted from
a query and document constructing an object called \emph{query-document}
and each pair is labelled with a relevance score in
$\set{0,1,2}$, indicating how relevant the document is to the
respective query as shown in \Cref{fig:letorsup}.
A relevance score of \num{0} means that the document is not
relevant, \num{1} means relevant and
\num{2} means very relevant to the query.
For this dataset, the goal of the choice problem is to choose all the relevant
documents for the given task.

\paragraph{Structure}
The dataset consists of a universal set of objects $\vec{x} \in \cX$.
Each instance of these datasets $\data_{S} = \set{(\tqi{1}, \lvq{1}), \ldots ,(\tqi{N}, \lvq{N})}$, is represented as
set of tuples $(\tqi{i}, \lvq{i})$, where $\tqi{i} = \cset$ is
the \qset ($\vec{x}_i$ features extracted from \emph{query-document}) and $\lvq{i} = (l_1, \ldots, l_n)$ represents
vector of relevance label for the given set of objects, such that
$\vec{x}_j \in \IR^{46}, \enspace l_j
  \in \set{0,1,2}$ for all $j\in [n]$  and  $ 5 \leq \card{\tqi{i}} \leq 147$ for every $i\in [N]$.

The size of the universal set of objects in the \mq{2007}
dataset is \num{59570}, \thatis $\card{\cX} = 59570$ and the \mq{2008} dataset is
\num{564}, \thatis $\card{\cX} = 12102$.
These datasets have been partitioned into \num{5} parts by
\citet{letor2013}, such that $\data_{S} = \data_{S1} \cup
  \data_{S2} \cup \data_{S3} \cup
  \data_{S4} \cup \data_{S5}$.
This partition is used to conduct \num{5}-fold
cross-validation, and for each fold, we use four parts for training and the
remaining part for testing as described in \Cref{tab:letordataset}.
\begin{table}[htb]
  \centering
  \caption{\num{5}-folds of the \ac{LETOR} dataset
    and the sub-sampled training \qsets of size \num{5}.}
  \label{tab:letordataset}
  \sisetup{
    mode=text
  }
  \begin{adjustbox}{width=\textwidth,center}
    \begin{tabular}{
        S[table-format=1]
        cc
        S[table-format=4]
        S[table-format=3]
        S[table-format=5]
        S[table-format=3]
        S[table-format=3]
        S[table-format=4]}
      \toprule
      \multicolumn{3}{c}{Dataset} & \multicolumn{3}{c}{\mq{2007}}      & \multicolumn{3}{c}{\mq{2008}}                                                                                                                                                        \\
      \midrule
      {Fold}                      & Test                               & Train                            & {\#\,Train}              & {\#\,Test}                         & {\#\,Sampled Train} & {\#\,Train}              & {\#\,Test} & {\#\,Sampled Train} \\
      \hline
      \num{1}                     & $\data_{S1}$                       & $\data_{S} \setminus \data_{S1}$ & \num{1172}               & \num{283}                          & \num{7111}          & \num{459}                & \num{105}  & \num{1187}          \\
      \num{2}                     & $\data_{S2}$                       & $\data_{S} \setminus \data_{S2}$ & \num{1160}               & \num{295}                          & \num{7012}          & \num{452}                & \num{112}  & \num{1083}          \\
      \num{3}                     & $\data_{S3}$                       & $\data_{S} \setminus \data_{S3}$ & \num{1163}               & \num{292}                          & \num{7069}          & \num{442}                & \num{122}  & \num{1122}          \\
      \num{4}                     & $\data_{S4}$                       & $\data_{S} \setminus \data_{S4}$ & \num{1160}               & \num{295}                          & \num{7047}          & \num{444}                & \num{120}  & \num{1203}          \\
      \num{5}                     & $\data_{S5}$                       & $\data_{S} \setminus \data_{S5}$ & \num{1165}               & \num{290}                          & \num{7077}          & \num{459}                & \num{105}  & \num{1201}          \\
      \midrule
      \multicolumn{3}{c}{}        & {\#\,Instances $\card{\data_{S}}$} & {\#\,Features}                   & {\#\,Objects $\card{Q}$} & {\#\,Instances $\card{\data_{S}}$} & {\#\,Features}      & {\#\,Objects $\card{Q}$}                                    \\
                                  &                                    &                                  & \num{1455}               & \num{46}                           & ${[6, 147]}$        & \num{564}                & \num{46}   & ${[5, 121]}$        \\
      \bottomrule
    \end{tabular}
  \end{adjustbox}
\end{table}

\paragraph{Choice Data Conversion}
The corresponding choice dataset is created by considering the documents in
$\tqi{i}$ as the task sets $\qi{i}$ and
the set of relevant documents 
$\ci{i} \coloneqq \set{ \vec{x}_j \in \tqi{i} \sothat l_j \in
    \set{1,2}}$ 
as the corresponding choice set for each instance $(\tqi{j}, \lvq{j}) \in
  \data_{S} \setminus \data_{Si}$.
For training the choice model, we sub-sample \num{10}
objects from each query instance $\tqi{i}$ to construct the
task sets.
Note, that we still evaluate the models on the corresponding test
choice dataset, which consists of all original queries for each fold  as described
in \Cref{tab:letordataset}.

\paragraph{LETOR Listwise Datasets}
The format of both listwise datasets is the same as the supervised one.
There are about \num{1700} queries in
\mql{2007} and about \num{800} queries in
\mql{2008} with each \emph{query-document} pair
consisting of \num{46} features.
In this dataset, all the documents for each query are labelled with a
real-valued relevance score instead of the multiple level relevance judgments
as shown in \Cref{fig:letorlist}.
The documents on top positions in the ground truth permutation have larger
value of the relevance degree.

\paragraph{Structure}
The dataset consists of a universal set of objects $\vec{x} \in \cX$.
Each instance of these datasets $\data_{L} = \set{(\tqi{1}, \lvq{1}), \ldots ,(\tqi{N}, \lvq{N})}$, is represented as
a set of tuples $(\tqi{i}, \lvq{i})$, where $\tqi{i} = \cset$ is
the \qset ($\vec{x}_i$ features extracted from \emph{query-document}) and $\lvq{i} = (l_1, \ldots, l_n)$ represents
a vector of relevance score for the given set of objects, such that
$\vec{x}_j \in \IR^{46}, \enspace l_j
  \in \IR$ for all $j \in [n]$ and $ 204 \leq \card{\tqi{i}} \leq 1831$ for every $i
  \in [N]$.

\begin{table}[htb]
  \centering
  \caption{\num{5}-folds of the \ac{LETOR}
    \mql{2007} and \mql{2008} dataset and the
    sub-sampled training \qsets of size \num{5}.}
  \label{tab:letordatasetlist}
  \sisetup{
    mode=text
  }
  \begin{adjustbox}{width=\textwidth,center}
    \begin{tabular}{
        S[table-format=1]
        cc
        S[table-format=4]
        S[table-format=3]
        S[table-format=5]
        S[table-format=3]
        S[table-format=3]
        S[table-format=4]}
      \toprule
      \multicolumn{3}{c}{Dataset} & \multicolumn{3}{c}{\mql{2007}} & \multicolumn{3}{c}{\mql{2008}}                                                                                                                                    \\
      \midrule
      {Fold}                      & Test                           & Train                            & {\#\,Train}              & {\#\,Test}      & {\#\,Sampled Train} & {\#\,Train}              & {\#\,Test} & {\#\,Sampled Train} \\
      \hline
      \num{1}                     & $\data_{L1}$                   & $\data_{L} \setminus \data_{L1}$ & \num{1353}               & \num{339}       & \num{97557}         & \num{627}                & \num{157}  & \num{71600}         \\
      \num{2}                     & $\data_{L2}$                   & $\data_{L} \setminus \data_{L2}$ & \num{1353}               & \num{339}       & \num{98055}         & \num{627}                & \num{157}  & \num{71908}         \\
      \num{3}                     & $\data_{L3}$                   & $\data_{L} \setminus \data_{L3}$ & \num{1353}               & \num{339}       & \num{97580}         & \num{627}                & \num{157}  & \num{72233}         \\
      \num{4}                     & $\data_{L4}$                   & $\data_{L} \setminus \data_{L4}$ & \num{1353}               & \num{339}       & \num{98000}         & \num{627}                & \num{157}  & \num{71868}         \\
      \num{5}                     & $\data_{L5}$                   & $\data_{L} \setminus \data_{L5}$ & \num{1356}               & \num{336}       & \num{98304}         & \num{628}                & \num{156}  & \num{71847}         \\
      \midrule
      \multicolumn{3}{c}{}        & {\#\,Instances}                & {\#\,Features}                   & {\#\,Objects $\card{Q}$} & {\#\,Instances} & {\#\,Features}      & {\#\,Objects $\card{Q}$}                                    \\
      Total                       &                                &                                  & \num{1692}               & \num{46}        & ${[257, 1346]}$     & \num{784}                & \num{46}   & ${[204, 1831]}$     \\
      \bottomrule
    \end{tabular}
  \end{adjustbox}
\end{table}

\paragraph{Singleton Choice Data Conversion}
The corresponding \dc datasets are created by considering the documents in
$\tqi{i}$ as the task sets $\qi{i}$ and
the most relevant document $\ci{i} = \set{\argmax_{\vec{x}_j \in \tqi{i}} l_j}$ as the corresponding \dc
set for each instance $(\tqi{j}, \lvq{j}) \in
  \data_{L} \setminus \data_{Li}$.
For training the \ac{SCM} we sub-sample
\num{10} objects from each query instance
$\tqi{i}$ to construct the task sets.
Note that we still evaluate the models on the corresponding \dc test dataset, which consists of all original queries for each fold  as described
in \Cref{tab:letordataset}.

\subsection{Expedia Hotel Dataset}
\label{asub:expedia}
Expedia released a dataset on the Kaggle website as a competition and for
research purposes\footnote{These datasets are available on \url{https://www.kaggle.com/c/expedia-personalized-sort/data}}.
The dataset includes browsing and booking data as well as information on price
competitiveness.
The data are organized around a set of search result impressions, the ordered
list of hotels that the user sees after they search for a hotel on the Expedia
website.
In addition to impressions from the existing algorithm, the dataset contains
impressions where the hotels were randomly sorted, to avoid the position bias
of the existing algorithm.
The user response is provided as a click on a hotel and/or a purchase of a
hotel room.
This dataset consists of \num{399344} search queries and
\num{45} features extracted from the search query and the hotel
constructing an object.
Each hotel is labelled with a relevance score of 
$0$, $1$ or $2$, indicating how relevant the hotel is to the respective query or the user.
A relevance score of \num{0} means that the hotel is not clicked, \num{1} means it was clicked and
\num{2} means the hotel was booked by the user.
This dataset is very similar to the \ac{LETOR} dataset as
shown in~\Cref{fig:letor}.
For this dataset, we define the learning target to be the set of relevant
hotels (clicked and/or booked).
Since for each query, the number of hotels displayed is different, this dataset
consists of different task sizes.

\paragraph{Structure}
The dataset consists of a universal set of objects $\vec{x} \in \cX$.
Each instance of the datasets $\data_{E} = \set{(\tqi{1}, \lvq{1}), \ldots ,(\tqi{N}, \lvq{N})}$, is represented as
a set of tuples $(\tqi{i}, \lvq{i})$, where $\tqi{i} = \cset$ is
the \qset ($\vec{x}_i$ features extracted from \emph{hotel}) and $\lvq{i} = (l_1, \ldots, l_n)$ represents
the vector of relevance label for the given set of objects, such that
$\vec{x}_j \in [-1, \infty]^{45}, \enspace l_j \in
  \set{0,1,2}$ for each  $j \in [n]$ and $ 5 \leq \card{\tqi{i}} \leq 38$ for all $i
  \in [N]$.

The number of instances $N$ in this dataset is
\num{399344}, \thatis $\card{\data_{E}} = 399344$ and the size of the universal set of objects
(hotels) is \num{136886}, \thatis $\card{\cX} = 136886$.
There are \num{31} features which have missing values, and
we removed the features which consist of more than \SI{50}{\percent}
missing values.
For the remaining \num{3} features which have of missing
values, we impute them with a negative value less than
\num{-1}.
The models are trained on the resulting dataset with
\num{17} features.

\begin{table}[tbp]
  \centering
  \caption{Properties of the Expedia dataset and the sub-sampled training queries of size
    \num{10}.}
  \label{tab:expediadataset}
  \begin{adjustbox}{width=\textwidth,center}
    \begin{tabular}{
        S[table-format=3]
        S[table-format=2]
        S[table-format=2]
        S[table-format=2]
        S[table-format=2]
        S[table-format=4]
        S[table-format=4]
        S[table-format=4]
        S[table-format=4]
        S[table-format=4]
      }
      \toprule
                             &
                             & \multicolumn{3}{c}{\#\,Features Missing Values}
                             & \multicolumn{4}{c}{\#\,Instances}
                             & {\#\,Objects}                                                                                                                                             \\
      \midrule
      {Learning Problem}
                             & {\#\,Features}
                             & {\#\,All}
                             & {$>$\,\SI{90}{\percent}}
                             & {$>$\,\SI{50}{\percent}}
                             & {\#\,Total}
                             & {\#\,Train}
                             & {\#\,Test}
                             & {\#\,Sampled Train}
                             & {$\card{Q}$}                                                                                                                                              \\
      \midrule
      \text{Choice}          & \num{45}                                        & \num{31} & \num{17} & \num{28} & \num{399344} & \num{79855} & \num{319489} & \num{238744} & ${[38, 5]}$ \\
      \text{Singleton choice} & \num{45}                                        & \num{31} & \num{17} & \num{28} & \num{390270} & \num{78041} & \num{312229} & \num{166940} & ${[38, 5]}$ \\
      \bottomrule
    \end{tabular}
  \end{adjustbox}
\end{table}
\paragraph{Data Conversion Process}
We create \num{5} folds by shuffle-splitting the dataset
randomly into \SI{80}{\percent} test and
\SI{20}{\percent} train instances.
The choice dataset is created by considering the hotels in
$\tqi{i}$ as the \qset $\qi{i}$ and the
set of relevant hotels 
$\ci{i} \coloneqq \{ \vec{x}_j \in \tqi{i} \sothat l_j \in
    \set{1,2} \}$
    as the corresponding choice
set for each instance $(\tqi{i}, \lvq{i}) \in  \data_{E}$.
The models are trained on the sampled training dataset and corresponding test
dataset using \num{5}-fold stratified cross-validation as
described in \Cref{tab:expediadataset}.
\paragraph{Singleton Choice}
In order to create the \dc dataset, we just consider the samples where the user
booked the hotel, which is the \dc for the given query.
The \dc dataset is created by considering the hotels in
$\tqi{i}$ as the \qset $\qi{i}$ and the
set of booked hotels 
$\ci{i} = \bigl\{ \vec{x}_j \in \tqi{i} \sothat l_j =
    2 \bigr\}$ 
    as the corresponding choice
set for each instance $(\tqi{i}, \lvq{i}) \in  \data_{E}$.
  
The models are trained on the sampled training dataset and corresponding test
dataset using \num{5}-fold stratified cross-validation as
described in \Cref{tab:expediadataset}.
Note, the instances where the hotel was not booked at all were discarded and
only the instances where there was booking were considered.
\subsection{SUSHI Dataset}
\label{asub:sushi}
SUSHI\footnote{This dataset can be downloaded from \url{http://www.kamishima.net/sushi/}} was another dataset released for solving the
task of \emph{object ranking}.
This dataset was collected by surveying \num{5000}
individuals, such that each person was provided with two item sets
$A$ and $B$.
Set $A$ consist of \num{10} most
famous sushi and $B$ consists of top
\num{100} sushi famous in Japan.
Individuals were asked to provide the preferences in form total order for items
in set $A$, and a real numbered score between
$0$ and $5$ for sushi in set
$B$.
There were missing rating values for many items in set
$B$, so they extracted the total order for the top
\num{10} preferred items by each user.

The SUSHI dataset consists of universal set of objects
$\vec{x} \in \cX$, with size \num{100}, \thatis $\card{\cX} = 100$, with \num{10000} set of
object $Q$ of size \num{10} and
each sushi consists of \num{7} features, \thatis
$\vec{x} \in \IR^{7}$.
The instances of the dataset $\data_{S} = \set{(\qi{1}, \pi_1), \ldots ,(\qi{N}, \pi_{N})}$, are represented as a set
of tuples $(\qi{i},\pi_i)$, where $\qi{i} = \cset$ is the
set of objects and $\pi_i$ represents the underlying
orderings for the given set of objects $\qi{i}$, such that
$N = \card{\data_{M}} = 10000$, $\vec{x}_i \in \IR^{7}$ and 
$\card{\qi{i}} = 10$ holds for all $i\in [N]$.

\begin{table}[tb]
  \centering
  \caption{Major Group feature description}
  \label{tab:mg}
    \begin{tabular}{
        S[table-format=3]
        S[table-format=4]
        S[table-format=4]
        S[table-format=4]
        S[table-format=4]
        S[table-format=4]
      }
      \toprule
      \multicolumn{6}{c}{Major Group}                                                                                                                 \\
      \midrule
      \text{Value} & \text{Species}                             & \text{Value} & \text{Species}          & \text{Value} & \text{Species}              \\
      \midrule
      \num{0}      & \text{Aomono (blue-skinned fish)}          & \num{4}      & \text{Clam or shell}    & \num{8}      & \text{ Other seafood}       \\
      \num{1}      & \text{Akami (red meat fish)}               & \num{5}      & \text{Squid or octopus} & \num{9}      & \text{Egg}                  \\
      \num{2}      & \text{Shiromi (white-meat fish)}           & \num{6}      & \text{Shrimp or crab}   & \num{10}     & \text{Meat other than fish} \\
      \num{3}      & \text{Tare (something like baste for eel)} & \num{7}      & \text{Roe}              & \num{11}     & \text{Vegetables}           \\
      \bottomrule
    \end{tabular}
\end{table}
The dataset contains the following features:
\begin{enumerate}
  \item Style: This is a binary feature, which describes whether the sushi is a Maki or
        other, where $0$ means Maki sushi and
        $1$ means others.
  \item Major Group: This is a binary feature, which describes whether it is listed as a seafood ($0$) or not ($1$).
  \item Minor group: Described the species group used to prepare the suchi.
  The group is denoted by the categorical value between $0$ and $11$, i.e. it lies in the set $\set{0,1,2,3,4,5,6,7,8,9,10,11}$.
  Refer to ~\cref{tab:mg} for description of each group.
  \item Oiliness/Heaviness: The amount of oil or fat present in the sushi, expressed as
        a real number between \num{0} and
        \num{4}, where \num{0} indicates 
        heavy/oil and \num{4} oil-free.
  \item Demand: The frequency with which the user demands the sushi, expressed as a
        real number between \num{0} and
        \num{3}, where  \num{3} means most
        frequently and \num{0} not at all.
  \item Normalized Price: The price of sushi normalized over the given
        \num{100} sushis.
  \item Supply: The frequency of selling a sushi in the shop, expressed as a real
        number between \num{0} and \num{1},
        where \num{0} indicates not at all and
        \num{1} frequently.
\end{enumerate}
\paragraph{Singleton Choice Data Conversion}
For using the SUSHI dataset for \dc setting, we re-utilize the set of object
$Q$ in $\data_{S}$ and choose the most
preferred object as the \dc.
We created the \dc dataset $\data_{SDC} = \set{(\qi{1}, \ci{1}),  \dotsc, (\qi{N}, \ci{N}}$ with
$N = \card{\data_{S}}$ instances, such that $\card{\qi{k}} = 10$ and $\ci{k} \coloneqq \set{\vec{x}_{\pi_i(1)}}$ for all $k\in [N]$. 
The \dc models are evaluated using \num{5}-folds by train-test shuffle-split with 80 \% train  and 20 \% test instances.

\section{Detailed Experimental Results}
\label{asub:results}
The following \cref{tab:choicemodels,tab:singletonchoice,tab:singletonchoice2}
contain all
experimental results as discussed in \cref{sub:results} in numeric
form for additional evaluation measures.
\begin{table}[htbp]  %
  \centering
  \caption{Results for the general subset choice models (mean and standard
    deviation of different measures, measured across \num{5} outer
    cross-validation folds).
    Best entry for each measure marked in bold.
  }
  \sisetup{
    table-align-uncertainty=true,
    separate-uncertainty=true,
    table-format=1.3(3),
    detect-weight,
    mode=text
  }
  \begin{adjustbox}{width=0.92\textwidth,center}
    \renewrobustcmd{\bfseries}{\fontseries{b}\selectfont}
    \renewrobustcmd{\boldmath}{}
    \begin{tabular}{
        l
        l
        S
        S
        S
        S}
      \toprule
      Dataset                              & {Choice Model} & {$F_1$-measure}    & {Subset $0/1$ Accuracy} & {Informedness}      & {AUC-ROC}           \\
      \midrule
      \multirow{8}{*}{Pareto-front-\num{2}D}  & \fetanet & \bfseries 0.942(8) & \bfseries 0.680(28) & \bfseries 0.956(12) & \bfseries 0.999(0)\\
											  & \fatenet & 0.912(9) & 0.506(37) & 0.911(6) & 0.996(1)\\
											  & \fetalinear & 0.673(1) & 0.064(7) & 0.694(15) & 0.955(0)\\
											  & \sda & 0.805(14) & 0.223(31) & 0.806(14) & 0.984(2)\\
											  & \ranknet & 0.612(7) & 0.060(10) & 0.672(14) & 0.971(6)\\
											  & \pairwisesvm & 0.588(1) & 0.044(3) & 0.646(7) & 0.956(0)\\
											  & \glm & 0.585(8) & 0.044(5) & 0.633(13) & 0.952(7)\\
											  & \allpositive & 0.232(0) & 0.000(0) & 0.000(0) & 0.500(0)\\
      \midrule
      \multirow{8}{*}{Pareto-front-\num{5}D}  & \fetanet & 0.826(5) & 0.001(0) & 0.406(32) & 0.854(14)\\
										      & \fatenet & \bfseries 0.904(4) & \bfseries 0.115(209) & \bfseries 0.743(94) & \bfseries 0.958(21)\\
										      & \fetalinear & 0.823(39) & 0.002(2) & 0.379(257) & 0.808(140)\\
										      & \sda & 0.887(2) & 0.013(1) & 0.656(12) & 0.935(2)\\
										      & \ranknet & 0.859(0) & 0.006(0) & 0.581(6) & 0.923(0)\\
										      & \pairwisesvm & 0.839(0) & 0.002(0) & 0.491(21) & 0.895(0)\\
										      & \glm & 0.826(29) & 0.002(1) & 0.402(225) & 0.738(351)\\
										      & \allpositive & 0.775(0) & 0.000(0) & 0.000(0) & 0.500(0)\\
	  \midrule
      \multirow{8}{*}{MNIST-Unique}           & \fetanet & 0.963(3) & 0.814(20) & 0.945(5) & 0.992(1)\\
										      & \fatenet & \bfseries 0.973(4) & \bfseries 0.848(21) & \bfseries 0.960(6) & \bfseries 0.995(1)\\
										      & \fetalinear & 0.562(1) & 0.000(1) & 0.000(1) & 0.517(1)\\
										      & \sda & 0.942(1) & 0.702(6) & 0.915(2) & 0.984(0)\\
										      & \ranknet & 0.562(0) & 0.000(0) & 0.000(0) & 0.504(1)\\
										      & \pairwisesvm & 0.562(0) & 0.000(0) & 0.000(0) & 0.511(6)\\
										      & \glm & 0.562(0) & 0.000(0) & 0.000(0) & 0.508(4)\\
										      & \allpositive & 0.562(0) & 0.000(0) & 0.000(0) & 0.500(0)\\
      \midrule

      \multirow{8}{*}{MNIST-Mode}             & \fetanet & 0.809(5) & 0.311(32) & 0.695(9) & 0.981(6)\\
										      & \fatenet & \bfseries 0.976(1) & \bfseries 0.883(10) & \bfseries 0.961(2) & \bfseries 0.992(1)\\
										      & \fetalinear & 0.597(1) & 0.003(0) & 0.003(2) & 0.516(1)\\
										      & \sda & 0.863(2) & 0.357(9) & 0.807(2) & 0.973(1)\\
										      & \ranknet & 0.597(0) & 0.003(0) & 0.000(0) & 0.503(2)\\
										      & \pairwisesvm & 0.597(0) & 0.003(0) & 0.000(0) & 0.509(6)\\
										      & \glm & 0.597(0) & 0.003(0) & 0.000(0) & 0.497(4)\\
										      & \allpositive & 0.597(0) & 0.003(0) & 0.000(0) & 0.500(0)\\
      \midrule

	  \multirow{8}{*}{\ac{LETOR}\mq{2007}}    & \fetanet & \bfseries 0.477(4) & \bfseries 0.007(2) & \bfseries 0.235(9) & \bfseries 0.729(11)\\
											  & \fatenet & 0.470(2) & 0.000(0) & 0.232(2) & 0.704(2)\\
											  & \fetalinear & 0.452(22) & 0.001(2) & 0.231(35) & 0.694(6)\\
											  & \sda & 0.441(15) & 0.001(2) & 0.195(22) & 0.666(3)\\
											  & \ranknet & 0.427(10) & 0.001(12) & 0.029(7) & 0.610(15)\\
											  & \pairwisesvm & 0.453(21) & 0.000(0) & 0.220(26) & 0.696(7)\\
											  & \glm & 0.427(21) & 0.001(2) & 0.058(29) & 0.614(9)\\
											  & \allpositive & 0.421(21) & 0.001(2) & 0.000(0) & 0.500(0)\\
      \midrule

      \multirow{8}{*}{\ac{LETOR}\mq{2008}}    & \fetanet & 0.537(1) & \bfseries 0.044(1) & \bfseries 0.440(3) & \bfseries 0.842(4)\\
										      & \fatenet & \bfseries 0.540(5) & 0.041(2) & 0.431(2) & 0.837(6)\\
										      & \fetalinear & 0.529(6) & 0.026(9) & 0.421(12) & 0.803(9)\\
										      & \sda & 0.425(41) & 0.018(11) & 0.287(38) & 0.727(23)\\
										      & \ranknet & 0.461(2) & 0.017(4) & 0.323(2) & 0.758(4)\\
										      & \pairwisesvm & 0.526(22) & 0.042(22) & 0.428(16) & 0.786(18)\\
										      & \glm & 0.493(28) & 0.014(10) & 0.311(61) & 0.739(19)\\
										      & \allpositive & 0.424(21) & 0.000(0) & 0.000(0) & 0.500(0)\\
      \midrule

	  \multirow{8}{*}{Expedia}                & \fetanet & 0.186(1) & 0.009(2) & 0.322(3) & 0.688(1)\\
										      & \fatenet & 0.198(6) & 0.018(2) & 0.346(10) & 0.707(7)\\
										      & \fetalinear & 0.179(7) & \bfseries 0.020(2) & 0.324(6) & 0.696(7)\\
										      & \sda & \bfseries 0.201(5) & 0.013(3) & \bfseries 0.352(12) & 0.708(8)\\
										      & \ranknet & 0.167(17) & 0.003(1) & 0.278(34) & \bfseries 0.716(6)\\
										      & \pairwisesvm & 0.129(17) & 0.004(2) & 0.165(97) & 0.680(50)\\
										      & \glm & 0.107(1) & 0.000(0) & 0.004(7) & 0.503(102)\\
										      & \allpositive & 0.106(0) & 0.000(0) & 0.000(0) & 0.500(0)\\
      \bottomrule
    \end{tabular}
  \end{adjustbox}
  \label{tab:choicemodels}
\end{table}

\begin{table}[tbp]  %
  \centering
  \caption{Mean and standard deviation of the accuracies on the \dc data
    (measured across \num{5} outer cross-validation folds).
    Best entry for each measure marked in bold.
  }
  \sisetup{
    table-align-uncertainty=true,
    separate-uncertainty=true,
    table-format=1.3(3),
    detect-weight,
    mode=text
  }
    \renewrobustcmd{\bfseries}{\fontseries{b}\selectfont}
    \renewrobustcmd{\boldmath}{}
    \begin{tabular}{
        l
        l
        S
        S
        S}
      \toprule
      Dataset                        & \ac{SCM}     & {\acc}              & {\topk{3}}         & {\topk{5}}         \\
      \midrule
      \multirow{10}{*}{Medoid}       & \fetanet     & 0.846(10)           & 0.994(1)           & \bfseries 1.000(0) \\
                                     & \fatenet     & \bfseries 0.881(7)  & \bfseries 0.996(1) & \bfseries 1.000(0) \\
                                     & \fetalinear  & 0.356(26)           & 0.715(7)           & 0.883(11)          \\
                                     & \sda 		& 0.839(4) 			  & 0.987(1) 		   & 0.998(0)\\
                                     & \ranknet     & 0.531(8)            & 0.873(6)           & 0.970(4)           \\
                                     & \pairwisesvm & 0.021(1)            & 0.194(9)           & 0.501(2)           \\
                                     & \ac{MNL}         & 0.020(1)            & 0.191(5)           & 0.500(1)           \\
                                     & \ac{NL}         & 0.049(14)           & 0.216(6)           & 0.463(27)          \\
                                     & \ac{GNL}         & 0.020(0)            & 0.195(4)           & 0.500(1)           \\
                                     & \ac{ML}         & 0.003(0)            & 0.055(12)          & 0.249(32)          \\
      \midrule

      \multirow{10}{*}{Hypervolume}  & \fetanet     & \bfseries 0.769(22) & \bfseries 0.933(7) & \bfseries 0.980(1) \\
                                     & \fatenet     & 0.730(18)           & 0.920(13)          & 0.968(6)           \\
                                     & \fetalinear  & 0.236(42)           & 0.404(42)          & 0.560(28)          \\
                                     &\sda           & 0.233(19)           & 0.417(29)          & 0.589(36)\\
                                     & \ranknet     & 0.203(4)            & 0.369(6)           & 0.562(4)           \\
                                     & \pairwisesvm & 0.186(1)            & 0.340(2)           & 0.550(2)           \\
                                     & \ac{MNL}         & 0.201(8)            & 0.360(10)          & 0.559(4)           \\
                                     & \ac{NL}         & 0.291(3)            & 0.511(7)           & 0.651(6)           \\
                                     & \ac{GNL}         & 0.293(18)           & 0.471(21)          & 0.663(14)          \\
                                     & \ac{ML}         & 0.189(14)           & 0.451(19)          & 0.621(14)          \\
      \midrule

      \multirow{10}{*}{MNIST-Unique} & \fetanet     & \bfseries 0.972(2)  & \bfseries 0.995(1) & \bfseries 0.998(0) \\
                                     & \fatenet     & 0.954(9)            & 0.993(1)           & \bfseries 0.998(1) \\
                                     & \fetalinear  & 0.127(6)            & 0.320(3)           & 0.505(10)          \\
                                     & \sda 		& 0.858(29) 		  & 0.935(26)		   & 0.955(18)\\
                                     
                                     & \ranknet     & 0.134(8)            & 0.307(2)           & 0.495(2)           \\
                                     & \pairwisesvm & 0.124(10)           & 0.319(8)           & 0.502(7)           \\
                                     & \ac{MNL}         & 0.170(6)            & 0.325(9)           & 0.495(2)           \\
                                     & \ac{NL}         & 0.207(16)           & 0.354(4)           & 0.502(6)           \\
                                     & \ac{GNL}         & 0.651(6)            & 0.763(3)           & 0.841(1)           \\
                                     & \ac{ML}         & 0.490(3)            & 0.718(5)           & 0.784(2)           \\
      \midrule

      \multirow{10}{*}{MNIST-Mode}   & \fetanet     & \bfseries 0.908(4)  & \bfseries 0.961(3) & \bfseries 0.978(4) \\
                                     & \fatenet     & 0.669(5)            & 0.907(4)           & 0.943(3)           \\
                                     & \fetalinear  & 0.290(6)            & 0.674(10)          & 0.877(7)           \\
                                     &\sda 			& 0.513(47) 		  & 0.806(41) 			& 0.901(61)\\
                                     & \ranknet     & 0.284(2)            & 0.668(3)           & 0.876(3)           \\
                                     & \pairwisesvm & 0.289(7)            & 0.675(11)          & 0.881(7)           \\
                                     & \ac{MNL}         & 0.285(6)            & 0.652(11)          & 0.853(10)          \\
                                     & \ac{NL}         & 0.282(7)            & 0.646(12)          & 0.848(10)          \\
                                     & \ac{GNL}         & 0.274(3)            & 0.641(8)           & 0.849(6)           \\
                                     & \ac{ML}         & 0.216(10)           & 0.536(20)          & 0.765(22)          \\
      \midrule

      \multirow{10}{*}{\msm}         & \fetanet     & 0.184(1)            & 0.481(2)           & 0.699(2)           \\
                                     & \fatenet     & \bfseries 0.185(3)  & \bfseries 0.482(6) & 0.699(4)           \\
                                     & \fetalinear  & 0.138(9)            & 0.391(23)          & 0.613(30)          \\
                                     &\sda 			& 0.099(22) 		  & 0.306(50) 			& 0.511(58)\\         
                                     & \ranknet     & 0.174(3)            & 0.477(2)           & \bfseries 0.708(3) \\
                                     & \pairwisesvm & 0.145(11)           & 0.405(19)          & 0.626(18)          \\
                                     & \ac{MNL}         & 0.179(2)            & 0.472(3)           & 0.694(4)           \\
                                     & \ac{NL}         & 0.178(4)            & 0.467(6)           & 0.689(7)           \\
                                     & \ac{GNL}         & 0.179(2)            & 0.472(3)           & 0.694(3)           \\
                                     & \ac{ML}         & 0.117(1)            & 0.353(9)           & 0.575(13)          \\
      \bottomrule
    \end{tabular}
  \label{tab:singletonchoice}
\end{table}

\begin{table}[tbp]  %
  \centering
  \caption{Mean and standard deviation of the accuracies on the \dc data
    (measured across \num{5} outer cross-validation folds).
    Best entry for each measure marked in bold.
  }
  \sisetup{
    table-align-uncertainty=true,
    separate-uncertainty=true,
    table-format=1.3(3),
    detect-weight,
    mode=text
  }
    \renewrobustcmd{\bfseries}{\fontseries{b}\selectfont}
    \renewrobustcmd{\boldmath}{}
    \begin{tabular}{
        l
        l
        S
        S
        S}
      \toprule
      Dataset                                    & \ac{SCM}     & {\acc}              & {\topk{3}}          & {\topk{5}}          \\

      \midrule
      \multirow{10}{*}{\mdm}                     & \fetanet     & \bfseries 0.512(4)  & \bfseries 0.835(4)  & \bfseries 0.942(2)  \\
                                                 & \fatenet     & 0.510(1)            & 0.830(2)            & 0.938(2)            \\
                                                 & \fetalinear  & 0.440(2)            & 0.759(2)            & 0.889(1)            \\
                                                 &\sda 			& 0.451(47) 		  & 0.694(72) 			& 0.789(54)\\   
                                                 & \ranknet     & 0.435(2)            & 0.779(1)            & 0.914(1)            \\
                                                 & \pairwisesvm & 0.369(16)           & 0.712(12)           & 0.871(8)            \\
                                                 & \ac{MNL}         & 0.447(2)            & 0.692(5)            & 0.795(5)            \\
                                                 & \ac{NL}         & 0.438(6)            & 0.671(15)           & 0.775(18)           \\
                                                 & \ac{GNL}         & 0.443(4)            & 0.681(10)           & 0.784(11)           \\
                                                 & \ac{ML}         & 0.417(3)            & 0.763(1)            & 0.895(5)            \\
      \midrule

      \multirow{10}{*}{\ac{LETOR}\mq{2007}-list} & \fetanet     & \bfseries 0.334(7)  & \bfseries 0.577(12) & \bfseries 0.705(6)  \\
                                                 & \fatenet     & 0.288(2)            & 0.508(6)            & 0.639(4)            \\
                                                 & \fetalinear  & 0.293(18)           & 0.551(7)            & 0.697(7)            \\
                                                 &\sda & 0.047(13) & 0.137(7) & 0.211(14)\\ 
                                                 & \ranknet     & 0.287(33)           & 0.513(50)           & 0.627(37)           \\
                                                 & \pairwisesvm & 0.302(8)            & 0.541(31)           & 0.654(39)           \\
                                                 & \ac{MNL}         & 0.282(6)            & 0.503(29)           & 0.622(38)           \\
                                                 & \ac{NL}         & 0.285(18)           & 0.499(30)           & 0.608(43)           \\
                                                 & \ac{GNL}         & 0.287(20)           & 0.509(29)           & 0.625(37)           \\
                                                 & \ac{ML}         & 0.282(5)            & 0.503(38)           & 0.628(37)           \\
      \midrule

      \multirow{10}{*}{\ac{LETOR}\mq{2008}-list} & \fetanet     & 0.266(15)           & 0.396(19)           & 0.504(17)           \\
                                                 & \fatenet     & \bfseries 0.281(12) & 0.369(15)           & \bfseries 0.544(12) \\
                                                 & \fetalinear  & 0.197(7)            & 0.392(27)           & 0.506(32)           \\
                                                 &\sda & 0.028(7) & 0.078(32) & 0.124(34)\\
                                                 & \ranknet     & 0.225(26)           & \bfseries 0.399(20) & 0.501(23)           \\
                                                 & \pairwisesvm & 0.203(14)           & 0.376(32)           & 0.497(21)           \\
                                                 & \ac{MNL}         & 0.217(25)           & 0.362(20)           & 0.500(27)           \\
                                                 & \ac{NL}         & 0.212(24)           & 0.355(30)           & 0.472(30)           \\
                                                 & \ac{GNL}         & 0.222(20)           & 0.366(34)           & 0.494(26)           \\
                                                 & \ac{ML}         & 0.213(15)           & 0.367(19)           & 0.501(25)           \\
      \midrule

      \multirow{10}{*}{Expedia}                  & \fetanet     & \bfseries 0.215(6)  & \bfseries 0.451(16) & 0.587(8)            \\
                                                 & \fatenet     & 0.203(6)            & 0.434(3)            & 0.576(3)            \\
                                                 & \fetalinear  & 0.176(3)            & 0.394(2)            & 0.543(3)            \\
                                                 & \sda 		& 0.115(8) 			  & 0.288(14) 		    & 0.431(15)\\ 
                                                 & \ranknet     & 0.210(1)            & 0.445(1)            & \bfseries 0.590(1)  \\
                                                 & \pairwisesvm & 0.179(0)            & 0.405(1)            & 0.550(0)            \\
                                                 & \ac{MNL}         & 0.199(4)            & 0.423(5)            & 0.565(4)            \\
                                                 & \ac{NL}         & 0.171(6)            & 0.388(8)            & 0.534(8)            \\
                                                 & \ac{GNL}         & 0.168(6)            & 0.385(10)           & 0.531(9)            \\
                                                 & \ac{ML}         & 0.181(10)           & 0.406(10)           & 0.551(7)            \\
      \midrule

      \multirow{10}{*}{SUSHI}                    & \fetanet     & 0.295(3)            & 0.552(3)            & 0.766(3)            \\
                                                 & \fatenet     & \bfseries 0.322(3)  & \bfseries 0.589(5)  & \bfseries 0.817(5)  \\
                                                 & \fetalinear  & 0.273(6)            & 0.500(14)           & 0.680(12)           \\
                                                 &\sda 			& 0.270(15) 			& 0.498(43) 	& 0.689(43)\\    
                                                 & \ranknet     & 0.272(7)            & 0.559(35)           & 0.721(16)           \\
                                                 & \pairwisesvm & 0.258(4)            & 0.480(22)           & 0.679(13)           \\
                                                 & \ac{MNL}         & 0.271(4)            & 0.502(3)            & 0.677(10)           \\
                                                 & \ac{NL}         & 0.253(6)            & 0.533(19)           & 0.730(25)           \\
                                                 & \ac{GNL}         & 0.259(7)            & 0.562(23)           & 0.735(16)           \\
                                                 & \ac{ML}         & 0.281(4)            & 0.575(13)           & 0.777(7)            \\
      \bottomrule
    \end{tabular}
  \label{tab:singletonchoice2}
\end{table}

\end{document}